\documentclass[fleqn,10pt]{wlscirep}
\usepackage[utf8]{inputenc}
\usepackage[T1]{fontenc}
\usepackage{import}
\usepackage{makeidx}
\usepackage{ulem}
\usepackage{symbols}
\usepackage{bm}
\usepackage{siunitx}
\usepackage{float}
\usepackage{stmaryrd}
\usepackage{textcomp}
\usepackage{collcell}
\usepackage{tabularx}
\usepackage{xcolor,colortbl}
\usepackage{booktabs}
\usepackage{multirow}
\usepackage{rotating}
\usepackage[resetlabels]{multibib}
\usepackage{caption}
\usepackage{hyperref}
\usepackage{algorithm}
\usepackage{algpseudocode}
\algnewcommand\algorithmicforeach{\textbf{for each}}
\algdef{S}[FOR]{ForEach}[1]{\algorithmicforeach\ #1\ \algorithmicdo}
\newcommand{\beginsupplement}{%
     \setcounter{section}{0}
        \renewcommand{\thesection}{Supplementary sec. \arabic{section}}%
        \setcounter{equation}{0}
        \renewcommand{\theequation}{Supplementary eq. \arabic{equation}}%
        \setcounter{table}{0}
        \renewcommand{\tablename}{Supplementary tab.}
        \setcounter{figure}{0}
        \renewcommand{\figurename}{Supplementary fig.}
     }

\usepackage{tikz}

\usetikzlibrary{positioning}
\usetikzlibrary{3d} 

\title{Progressive reduced order modeling: empowering data-driven modeling with selective knowledge transfer}

\author[1]{Teeratorn Kadeethum}
\author[2]{Daniel O'Malley}
\author[3]{Youngsoo Choi}
\author[2]{Hari S. Viswanathan}
\author[1,*]{Hongkyu Yoon}

\affil[1]{Sandia National Laboratories, Albuquerque, NM, 87185, USA}
\affil[2]{Los Alamos National Laboratory, Los Alamos, NM, 87545, USA}
\affil[3]{Lawrence Livermore National Laboratory, Livermore, CA, 94550, USA}

\affil[*]{hyoon@sandia.gov}

\begin{abstract}
Data-driven modeling can suffer from a constant demand for data, leading to reduced accuracy and impractical for engineering applications due to the high cost and scarcity of information. To address this challenge, we propose a progressive reduced order modeling framework that minimizes data cravings and enhances data-driven modeling's practicality. Our approach selectively transfers knowledge from previously trained models through gates, similar to how humans selectively use valuable knowledge while ignoring unuseful information. By filtering relevant information from previous models, we can create a surrogate model with minimal turnaround time and a smaller training set that can still achieve high accuracy. We have tested our framework in several cases, including transport in porous media, gravity-driven flow, and finite deformation in hyperelastic materials. Our results illustrate that retaining information from previous models and utilizing a valuable portion of that knowledge can significantly improve the accuracy of the current model. We have demonstrated the importance of progressive knowledge transfer and its impact on model accuracy with reduced training samples. For instance, our framework with four parent models outperforms the no-parent counterpart trained on data nine times larger. Our research unlocks data-driven modeling's potential for practical engineering applications by mitigating the data scarcity issue. Our proposed framework is a significant step toward more efficient and cost-effective data-driven modeling, fostering advancements across various fields.

\end{abstract}
\begin{document}

\flushbottom
\maketitle

\thispagestyle{empty}

\begin{NoHyper}
\section*{Introduction}

In our quest to sustain global economics and address population growth, we face the pressing need to produce energy sustainably, extract groundwater responsibly, and store undesirable substances underground \cite{masood2022cop27, falk2022urgent}. However, understanding the complex multiphysics phenomena that govern subsurface physics and characterize underground structures poses significant challenges. These structures are often anisotropic, heterogeneous, and exhibit discontinuous material properties that span multiple orders of magnitude \cite{hu2008multiple, hartmann2016putting}. Traditional approaches, such as lab experiments, are expensive, time-consuming, and limited in their ability to simulate complex initial and boundary conditions \cite{ginn2002processes}. On the other hand, field trials can be risky, compromising operational safety and potentially leading to irreversible damage to assets. Furthermore, a lack of understanding of these complex behaviors may prevent optimizing development plans \cite{mccarthy2004colloid}. High-fidelity numerical models like finite difference, finite volume, or finite element approaches present an alternative to overcome these limitations. These models can effectively simulate coupled processes across different scales, accommodating complex geometries, boundary conditions, and heterogeneous materials \cite{evans2012numerical}. However, their use often entails a high computational cost, mainly when dealing with highly nonlinear and high-resolution problems \cite{hesthaven2016certified}. For instance, in subsurface physics applications such as geothermal or geologic carbon storage, considerable uncertainties exist in parameter space, including rock properties and reservoir management \cite{chen2023capacity,wen2023real}. To comprehensively understand the impact of these uncertainties on operational risks like induced seismicity or groundwater contamination, conducting over 10,000 Monte Carlo simulations may be necessary \cite{lengler2010impact,cho2021estimation}. Each simulation can be time-consuming, taking hours or even days to complete, making this procedure impractical. \par


Machine learning has revolutionized the field of data-driven modeling, offering a faster and more accessible approach to tackle a wide array of complex physical problems \cite{choi2020sns, kapteyn2022data, silva2023data}. By leveraging this technology, we can now expedite crucial workflows like large-scale inverse modeling, optimizations, and uncertain quantification processes \cite{qin2021deep, xu2022physics, pachalieva2022physics}, significantly boosting our understanding and decision-making capabilities. However, data-driven modeling faces significant hurdles, including the constant hunger for data. Acquiring or generating high-quality data can be time-consuming, costly, and resource-intensive, mainly when dealing with high-dimensional datasets. Furthermore, as the number of parameters increases, the demand for training samples grows exponentially, leading to intractable problems. It's important to highlight another significant limitation of data-driven modeling: it does not consider physical principles, potentially leading to predictions that lack a physical basis. One viable approach to address this concern is introducing a residual evaluation mechanism based on a system of partial differential equations, as demonstrated in the concept of physics-informed neural networks \cite{raissi2019physics}. However, for the purpose of this study, our primary focus will be on tackling this specific challenge.

One can utilize transfer learning to overcome the data-hungry challenge and make data-driven machine learning modeling more practical and efficient. Transfer learning is a powerful approach that leverages previously trained models to enhance the current model's performance \cite{weiss2016survey}. There are various techniques for transfer learning, two of which are commonly used: fine-tuning and adding fresh layers. In fine-tuning, we take the entire set of trainable parameters from a pre-trained model and use them as an initialization for the current model. A relatively small learning rate is employed during the training process to allow the model to adjust and adapt to the new task while retaining the knowledge gained from the previous training. This approach is practical when the pre-trained model's features are applicable to the current task. \par

Alternatively, in adding fresh layers, we utilize the pre-trained trainable parameters and introduce new layers to the model. These fresh layers can be added to the pre-trained model or replaced with specific layers. The pre-trained layers are typically frozen, preventing their weights and biases from being updated during training, while only the newly introduced layers are trained. This method is handy when the pre-trained model captures generic features that can be used as a starting point for the current task. While transfer learning techniques offer significant benefits, it is essential to acknowledge their limitations. Two notable drawbacks are (1) current procedures only leverage knowledge from one model and apply it to the next (i.e., one-to-one transfer) \cite{zhuang2020comprehensive}. Imagine you have many (e.g., 10’s-100’s) trained models that might be useful but can only utilize one; that would be a considerable loss of opportunities and (2) the phenomenon known as catastrophic forgetting, where the newly trained model completely loses the knowledge of the previous task and cannot be applied without retraining \cite{rusu2016progressive}. This limitation highlights the need for progressive learning approaches that enable knowledge transfer while preserving previously acquired information. The survey by Zhuang et al. \cite{zhuang2020comprehensive} provides a valuable resource for a comprehensive understanding of transfer learning techniques, including their advantages and disadvantages. \par

Progressive neural networks (PNN) have been developed to address the challenge of many-to-one transfer and mitigate the issue of catastrophic forgetting, leveraging prior knowledge from previously trained models through their lateral connections \cite{rusu2016progressive}. PNN is often associated with lifelong or continual learning, where a model continues learning from new data and tasks without forgetting previously learned information. Imagine a scenario where we aim to build a data-driven model that emulates high-fidelity models for single-phase flow problems. Traditionally, we would create a training set, train our proxy model, optimize hyperparameters, and utilize it for our desired applications (e.g., \cite{kadeethum2021framework,kadeethum2022reduced, kadeethum2022uqbtrom}). However, this approach becomes inefficient when we tackle multi-phase problems the next day, as we repeat the entire process, disregarding the valuable knowledge gained from previously trained models.

From our perspective, the solution lies in progressive reduced order models (built upon the PNN concept), which aim to capitalize on prior knowledge and make the most of previously developed models, see Fig. \ref{fig:N-levelTransferLearning}. Compared with ongoing research on transfer learning, this framework progressively accumulates common features of physics of interest and expands them instead of fine-tuning (including adding fresh layers) \cite{weiss2016survey, zhuang2020comprehensive} or enforcing loss functions \cite{goswami2022deep} resulting in immunization to catastrophic forgetting. Besides, it can transfer learning from many trained models instead of one-to-one transfer. The main ingredient of this framework lies within \emph{gates} used to control information flow from previous generations (e.g., grandparents and parents). 

These gates are pivotal in controlling information flow while concurrently training a child model. The total trainable parameters consist of a combination of trainable parameters from the child model and trainable parameters from all the gates associated with the child model, as indicated by the red square in Fig. \ref{fig:N-levelTransferLearning}. Each gate corresponds to a connection between the child and a preceding model (grandparents or parents in this illustration). During the training process, if the information from a prior model proves beneficial (i.e., it reduces the evaluation of a loss function), the corresponding gate increases the flow of information and vice versa. In this context, we employ a linear layer as a gate, meaning that the trainable parameters of each gate are adjusted in response to evaluating the loss function. It's worth noting that one can enhance the complexity of these gates by employing techniques such as recurrent units \cite{huang2023gate}, attention mechanisms \cite{vaswani2017attention}, or adaptive algorithms \cite{alesiani2023gated}.

\begin{figure}[!ht]
  \centering
  \includegraphics[width=12.0cm,keepaspectratio]{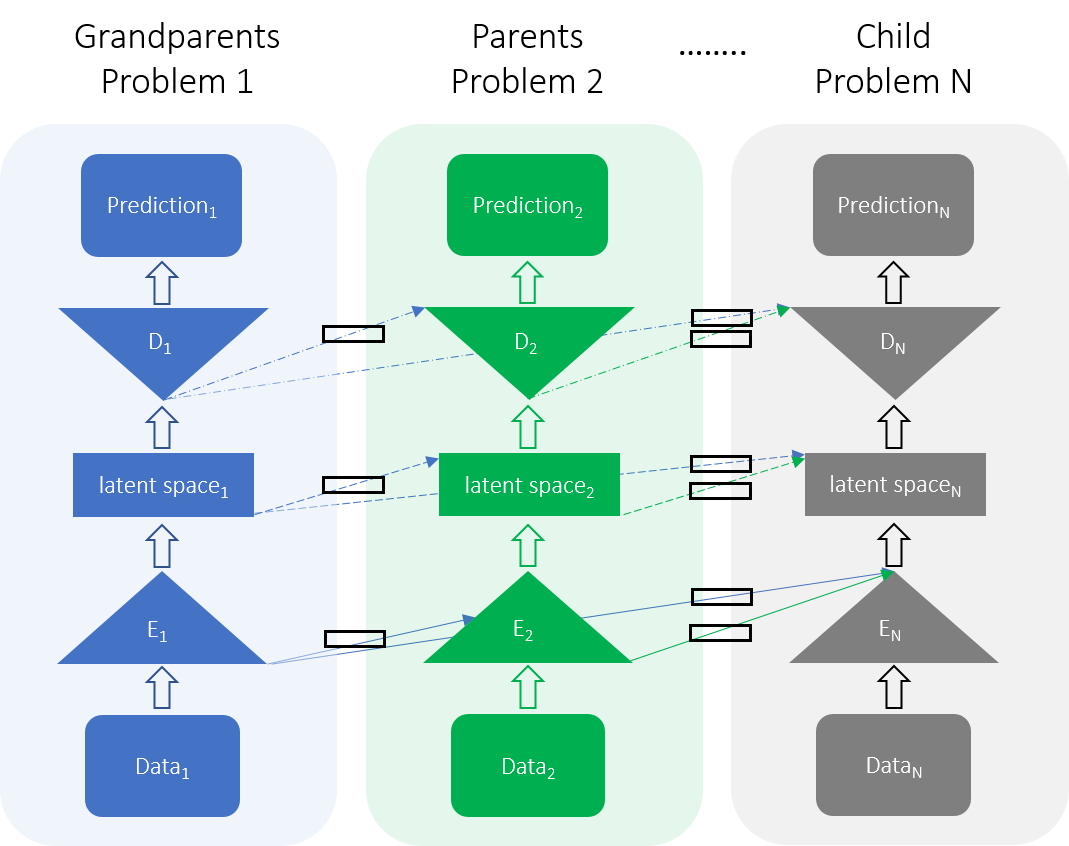}
  \caption{A schematic figure of multi-level transfer learning. $N$ pre-trained data-driven ML models for $N$ different problems can accelerate the training and improve the accuracy of a new data-driven ML model for the new problems through multi-level transfer learning techniques. E and D represent the Encoder and Decoder, respectively. The red square represents combined trainable parameters for the child model.}
  \label{fig:N-levelTransferLearning}
\end{figure}

Since this framework, Fig. \ref{fig:N-levelTransferLearning} is non-intrusive or data-driven in nature (i.e., it does not require a knowledge of underlying physics, governing equations, or modifications inside high-fidelity software), it enables our proposed platform to learn from physics simulators that can be treated as black-box. Throughout this study, we will use the Barlow Twins reduced order model (BT-ROM) \cite{kadeethum2022reduced} to showcase our progressive reduced order modeling framework (p-BT-ROM hereafter). We use a series of physics-based problems ranging from transport in porous media and gravity-driven flow to a finite deformation of a hyperelastic material since we want to emphasize that transferring knowledge among different physics (fluid to solid mechanics, in this case) is possible. Besides, we employ a range of topologies from 2- to 3-Dimensional domains to further emphasize the possibilities of transferring knowledge across different topologies. 

We aim to demonstrate that our framework effectively mitigates the data scarcity challenge, often associated with approaches like few-shot learning \cite{drori2022neural}, within data-driven methodologies. Instead of starting anew for each task, we systematically build upon existing models, harnessing the knowledge and insights acquired from prior models to inform and enhance subsequent iterations. This strategy conserves time and resources and significantly improves the overall efficiency of data-driven machine learning. This transformative approach expedites research and development in subsurface physics, facilitating safer and more cost-effective decision-making for energy production, groundwater extraction, and underground storage.

\par




\section*{Results}\label{sec:numer_results}


We introduce the proposed progressive learning framework that demonstrates enhanced accuracy compared to models lacking prior knowledge (without a parent) and achieves comparable or superior accuracy to models trained on larger datasets (more training samples but without a parent). In this section, we assess the performance of the progressive Barlow Twins reduced order model (p-BT-ROM) by evaluating its results on a range of physics problems discussed in Sections Data generation and \ref{sec:physics_problems}.

\subsection*{Data generation}\label{sec:data_gen}

We will be showcasing four physics problems, which are detailed in Table \ref{tab:sum_problems}. The first problem, referred to as Problem \#1, involves the study of transport in porous media with varying velocity fields. The parameter space for this problem encompasses fluid velocity, and our primary focus is on the concentration. Moving on to Problem \#2, it revolves around transport scenarios with varying diffusivity coefficients. In this case, the parameter space pertains to the diffusivity coefficient, and we continue to analyze the concentration as the quantity of interest. Problem \#3 explores gravity-driven flow in porous media, with the Rayleigh number ($\mathrm{Ra}$) as a parameter. In this instance, the Rayleigh number defines the parameter space, and our quantity of interest is the fluid temperature. This problem is adapted from the research conducted by Kadeethum et al. \cite{kadeethum2021nonTH}. Finally, Problem \#4 addresses the finite deformation of a hyperelastic material. The parameter space for this problem varies according to dimensionality. For the 2-Dimensional case, it involves the displacement at a boundary, while for the 3-Dimensional case, it includes the displacement within the source term. In both cases, our primary interest lies in the magnitude of solid displacement. This problem is adapted from the work of Kadeethum et al. \cite{kadeethum2022fomassistrom}. For comprehensive details regarding the problem settings, quantities of interest, parameter space, and the methodology for generating our training, validation, and testing datasets, please refer to Section \ref{sec:data_generation}.

\begin{table}[htbp]
  \centering
  \caption{Summary of physical problems. }
\begin{tabular}{|l|l|l|l|l|l|l|}
\hline
Problem              & Dimension          & Physics                                                                         & State variables               & Quantity of interest                                                                        & Parameters                                                                  & Section                                                           \\ \hline
\#1                  & 2                  & \begin{tabular}[c]{@{}l@{}}transport in \\ porous media\end{tabular}            & concentration                 & concentration                                                                               & \begin{tabular}[c]{@{}l@{}}fluid \\ velocity\end{tabular}                   & \ref{sec:physics_problems_tv}                  \\ \hline
\#2                  & 2                  & \begin{tabular}[c]{@{}l@{}}transport in \\ porous media\end{tabular}            & concentration                 & concentration                                                                               & \begin{tabular}[c]{@{}l@{}}diffusivity \\ coefficient\end{tabular}         & \ref{sec:physics_problems_td}                  \\ \hline
\multirow{3}{*}{\#3} & \multirow{3}{*}{2} & \multirow{3}{*}{\begin{tabular}[c]{@{}l@{}}gravity-driven \\ flow\end{tabular}} & pressure                      & \multirow{3}{*}{temperature}                                                                & \multirow{3}{*}{\begin{tabular}[c]{@{}l@{}}Rayleigh \\ number\end{tabular}} & \multirow{3}{*}{\ref{sec:physics_problems_el}} \\ \cline{4-4}
                     &                    &                                                                                 & velocity                      &                                                                                             &                                                                             &                                                                   \\ \cline{4-4}
                     &                    &                                                                                 & temperature                   &                                                                                             &                                                                             &                                                                   \\ \hline
\multirow{2}{*}{\#4} & 2                  & \multirow{2}{*}{solid deformation}                                              & \multirow{2}{*}{displacement} & \multirow{2}{*}{\begin{tabular}[c]{@{}l@{}}magnitude of \\ displacement\end{tabular}} & \begin{tabular}[c]{@{}l@{}}boundary \\ condition\end{tabular}              & \multirow{2}{*}{\ref{sec:physics_problems_hy}} \\ \cline{2-2} \cline{6-6}
                     & 3                  &                                                                                 &                               &                                                                                             & source term                                                                &                                                                   \\ \hline
\end{tabular}
\label{tab:sum_problems}
\end{table}

\subsection*{Investigation strategy}\label{sec:invest_strategy}

\textbf{Throughout this Result section in the main text}, we have designed the numerical experiments to address two primary questions. The following is an outline of the experimental setup and structure:
\begin{enumerate}

\item First Question: Knowledge Transfer between Physics with Fixed Topology

\begin{itemize}

\item Objective: Determine if the progressive learning framework can transfer knowledge between different physics problems while maintaining a fixed topology (i.e., same mesh and connectivity).
\item Parent Problems (Fluid): Problems \#1, \#2, and \#3
\item Child Problem (Solid): Problem \#4
\item Implication: Successful knowledge transfer implies the framework's ability to share knowledge across different physics domains.

\end{itemize}

\item Second Question: Knowledge Sharing between Different Topologies

\begin{itemize}

\item Objective: Investigate whether the progressive learning framework can share knowledge between different topologies.
\item Parent Problems (2-Dimensional Topology): Problems \#1, \#2, \#3, and \#4
\item Child Problem (3-Dimensional Topology): Problem \#4
\item Implication: Successful knowledge sharing indicates the framework's potential to transfer knowledge across different dimensions and topologies.

\end{itemize}
\end{enumerate}

\noindent
To ensure a systematic approach, we divide the investigation into two stages:

\begin{itemize}

\item  Stage 1: Similar Topology (Section Similar Topology and \ref{sec:sup_same_topo})

Concentration on problems with different physics but the same topology.
Focus on evaluating the effects of the number of training samples and the number of parents on the performance of the Barlow Twins reduced order model (BT-ROM) \cite{kadeethum2022reduced} and p-BT-ROM.

\item  Stage 2: Different Topologies (Section Different Topologies)

An investigation involving problems with different physics and different topologies.
Emphasis on analyzing the impact of the number of training samples and the number of parents on the performance of BT-ROM and p-BT-ROM.
Detailed descriptions of each topology are provided in \ref{sec:geo}.
By following this systematic structure, we aim to provide clear insights into the progressive learning framework's capabilities regarding knowledge transfer and sharing across various physics domains, topologies, and dimensions.

\end{itemize}

In addition to our research findings in the main text, we offer comprehensive numerical examples in supplementary sections to enhance the understanding of our work. Specifically, Section \ref{sec:init_results} demonstrates the conventional approach of transfer learning, where pre-trained models' weights and biases are utilized to initialize a new model (referred to as "init.") \cite{weiss2016survey, zhuang2020comprehensive}. This approach serves as a benchmark for comparison with our progressive learning framework, known as p-BT-ROM, presented in Section \ref{sec:pbt_results}. Furthermore, we explore the potential benefits of combining initialization with p-BT-ROM, denoted as "p-BT-ROM with init." Through our analysis in Section \ref{sec:si_discuss_results}, we conclude that the preferred approach going forward is to employ p-BT-ROM with init. To supplement our findings in Section Similar topology, we provide additional results of p-BT-ROM with init. for Problems \#1, \#2, \#3, and \#4 with fixed topology in Section \ref{sec:sup_same_topo}. This supplementary section sheds light on the performance of our framework when applied to different problem instances. Additionally, we have thoroughly investigated the impact of varying the number of parents within our framework to assess its effectiveness, yielding valuable insights. \par

\subsection*{Similar topology}\label{sec:same_topo}

We will illustrate that progressive learning enables us to transfer knowledge to problems with different underlying physics (e.g., fluid to solid mechanics) and parameter spaces (e.g., boundary conditions or material properties). We present a sample of a reconstruction loss \eqref{eq:loss_ae} result of the validation set (as a function of the number of epochs and parents) for Problem \#4 in a 2-Dimensional setting (the detail of this geometry is shown in Supplementary fig. \ref{fig:sum_geo}a) in Fig. \ref{fig:prelim_results}. We note that more comprehensive results can be found in \ref{sec:pbt_init_results}, but here, we want to introduce our framework's benefit briefly for the training phase of the (p-)BT-ROM (see Supplementary fig. \ref{fig:bbt_explain} - third step). To reiterate, our approach involves a combination of p-BT-ROM, utilizing the weights and biases of the trained models as an initialization. We provide in-depth results and rationale for adopting this approach in Sections \ref{sec:init_results}, \ref{sec:pbt_results}, \ref{sec:pbt_init_results}, and \ref{sec:si_discuss_results}. From this figure, the more parents we provide the p-BT-ROM model, the more accurate it becomes. We note that with 0 parents, the p-BT-ROM is the original BT-ROM \cite{kadeethum2022reduced}. These results imply that the p-BT-ROM could transfer knowledge from fluid problems (Problems \#1 to \#3) to a solid problem (Problem \#4). We note that as we add more parents to the framework, our trainable parameters inevitably become larger (see \ref{sec:num_para}). These additional parameters lead to a heavier computational time and memory load, but the cost remains far below running the physical model.  \par

\begin{figure}[!ht]
  \centering
  \includegraphics[width=15.0cm,keepaspectratio]{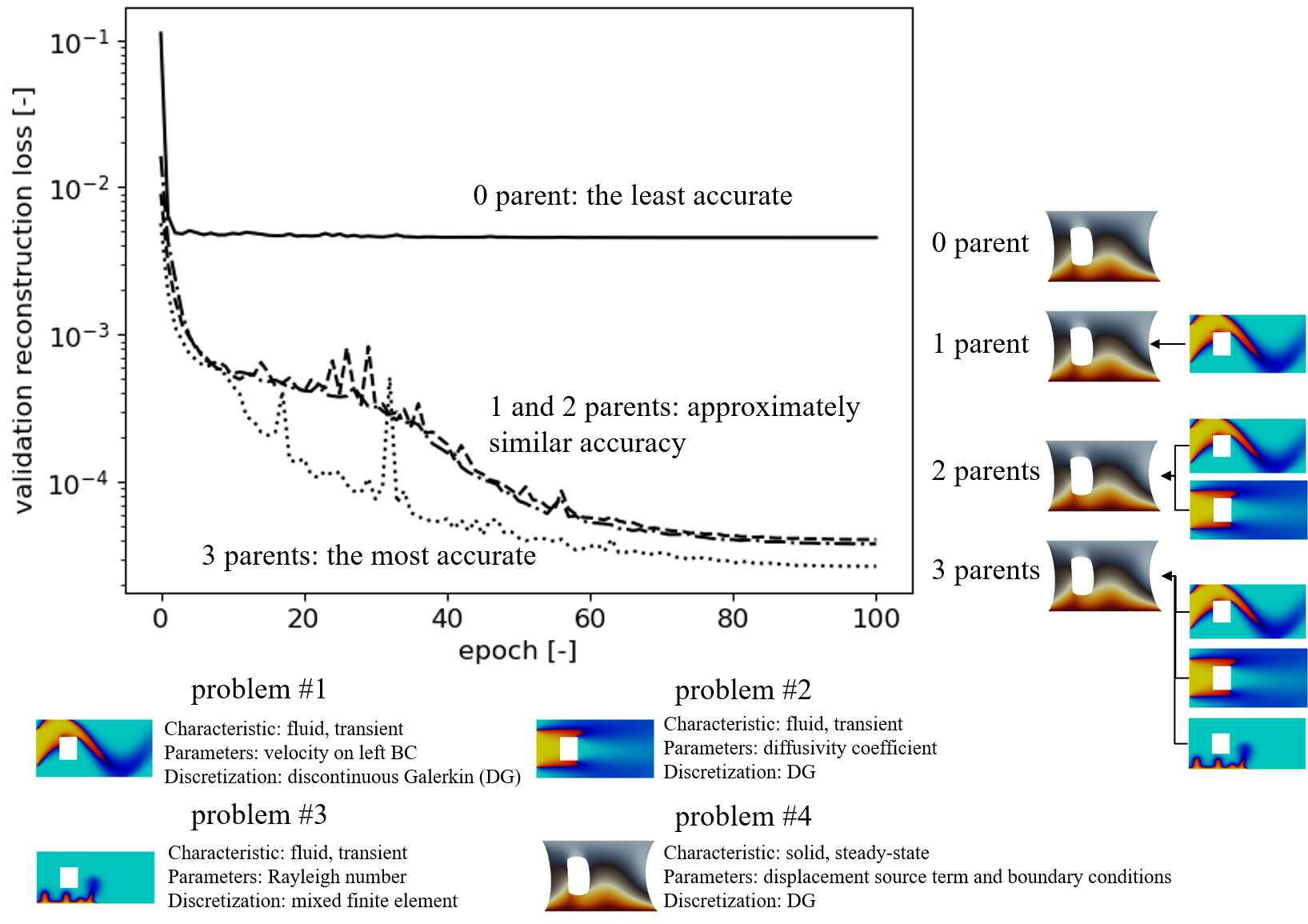}
  \caption{Illustration that our framework can transfer knowledge from fluid mechanics (Problems \#1-\#3) to solid mechanics (Problem \#4) problems. Reconstruction loss (\eqref{eq:loss_ae}) of the validation set for Problem \#4 (\ref{sec:physics_problems_hy}) with 2-Dimensional domain (Supplementary fig. \ref{fig:sum_geo}a). This loss reflects the training phase of the (p-)BT-ROM (see Supplementary fig. \ref{fig:bbt_explain} - third step). We note that for Problem \#4, we show the results in the deformed configuration.}
  \label{fig:prelim_results}
\end{figure}

To summarize our testing process, we train our model with a specific number of epochs (100 in this case, however, one should always perform trial and error for any new physics-based problems or new neural network architecture.) with our \emph{training set} (see Supplementary fig. \ref{fig:bbt_explain} - third step). Subsequently, we pick a set of weights and biases that deliver the lowest loss for a \emph{validation set} (still Supplementary fig. \ref{fig:bbt_explain} - third step). Finally, we use a decoder of p-BT-ROM with the set of weights and biases in combination with radial basis function (RBF) interpolator for the evaluation using a \emph{test set} (see Supplementary fig. \ref{fig:bbt_explain} - fourth and fifth steps). We present the test sets' results in \ref{sec:sup_same_topo}. In a nutshell, the results of the test samples are similar to the behavior we observed in the validation cases (Fig. \ref{fig:prelim_results}). To elaborate, as we increase the number of parents, our p-BT-ROM gains more accuracy. As expected, the BT-ROM (0 parent) has the worst accuracy. However, even the model with many parents still could not achieve the same level of accuracy as the model that trained with a very large training set with 0 parent (BT-ROM). This behavior implies that there are benefits of adding training samples that progressive neural networks might not be able to compensate for or substitute. On the other hand, they can produce more accurate results for fixed-sized datasets (supplemented with data from the simpler parent problems).\par

\subsection*{Different topologies}\label{sec:diff_topo}

In this study, we demonstrate the effectiveness of the progressive approach in transferring knowledge to problems with significantly different topologies. Specifically, we investigate the transfer from 2-dimensional to 3-dimensional domains or between 2-dimensional domains with varying numbers of holes (1 to 3 holes). We assess the validation reconstruction loss (\eqref{eq:loss_ae}) as a function of the number of epochs and parents for Problem \#4 in a 3-dimensional context (see Supplementary fig. \ref{fig:sum_geo}d), shown in Fig. \ref{fig:prelim_results_diff_topo}. Consistent with our earlier observations, we find that increasing the number of parents provided to the p-BT-ROM model improves its accuracy. Notably, there is a minimal discrepancy between the model's performance with 3 parents and that with 4 parents. These results indicate the feasibility of transferring knowledge across different topologies and dimensions, ultimately enhancing the precision of our data-driven framework. Additionally, as discussed in the previous section on similar topology, incorporating more parents into the framework expands the trainable parameter space, increasing training costs. For a comprehensive understanding of the number of parameters involved, please refer to \ref{sec:num_para}. \par


\begin{figure}[!ht]
  \centering
  \includegraphics[width=15.0cm,keepaspectratio]{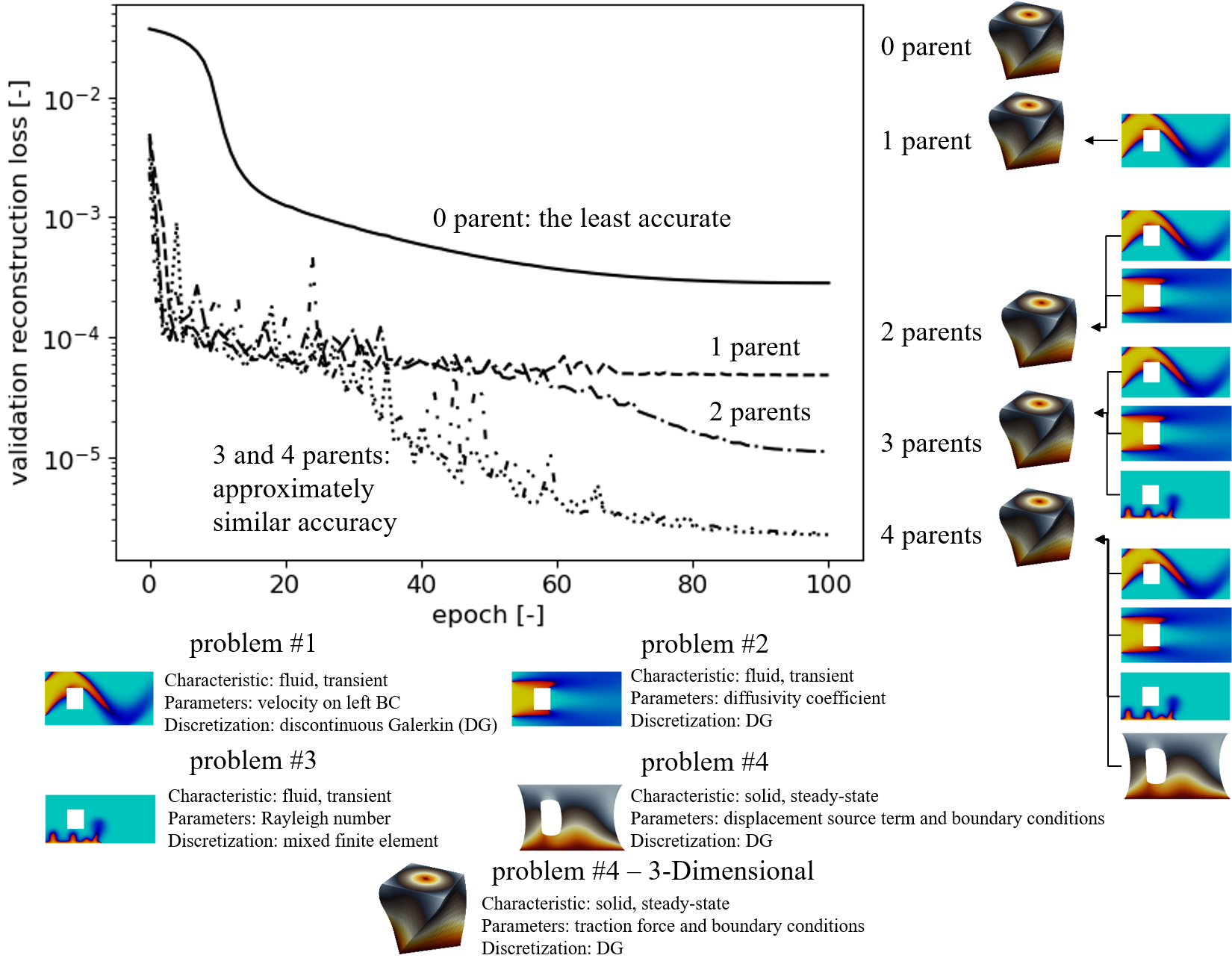}
  \caption{Illustration that our framework can transfer knowledge from 2-Dimensional to 3-Dimensional problems. Reconstruction loss (\eqref{eq:loss_ae}) of the validation set for Problem \#4 (\ref{sec:physics_problems_hy}) with 3-Dimensional domain (Supplementary fig. \ref{fig:sum_geo}d). This loss reflects the training phase of the (p-)BT-ROM (see Supplementary fig. \ref{fig:bbt_explain} - third step). We note that for Problem \#4, we show the results in the deformed configuration.}
  \label{fig:prelim_results_diff_topo}
\end{figure}

From Section Similar topology, we have illustrated that our p-BT-ROM can improve the accuracy of BT-ROM by adding more parents to the model. However, our investigation is limited to one mesh; see Supplementary fig. \ref{fig:sum_geo}a, for two physics problems, Problems \#3 to \#4. Here, we will investigate the p-BT-ROM's performance using different meshes and topologies, see Supplementary figs. \ref{fig:sum_geo}b-c for three physics problems, Problems \#2 to \#4 (test model performance on different physics and topologies). The schematic of p-BT-ROM specifies child$-$parent(s) relationship used throughout this section, and its subsections are shown in Fig. \ref{fig:diff_topo_scheme}. In short, we use the p-BT-ROM with the small dataset from \ref{sec:same_topo_problem4}. To elaborate, for our 4 parents, we use Problem \#1 (\ref{sec:physics_problems_tv}) with a training set of $\mathrm{M}N_t = 2405$ ($\mathrm{M} = 5$), Problem \#2 (\ref{sec:physics_problems_td}) with a training set of $\mathrm{M}N_t = 2405$ ($\mathrm{M} = 5$), Problem \#3 (\ref{sec:physics_problems_el}) with a training set of $\mathrm{M}N_t = 3623$ ($\mathrm{M} = 5$), and Problem \#4 (\ref{sec:physics_problems_hy}) with a training set of $\mathrm{M} = 100$. This is representative of many realistic scenarios where the model for which we need a ROM can only produce a small training set with $M=100$ samples, but larger datasets are available for simpler problems. We want to emphasize that we can combine different types of physics problems (i.e., fluid and solid mechanics problems) and different characteristics (i.e., transient and steady-state problems) into our pool of parents. \par

\begin{figure}[!ht]
  \centering
  \includegraphics[width=14.0cm,keepaspectratio]{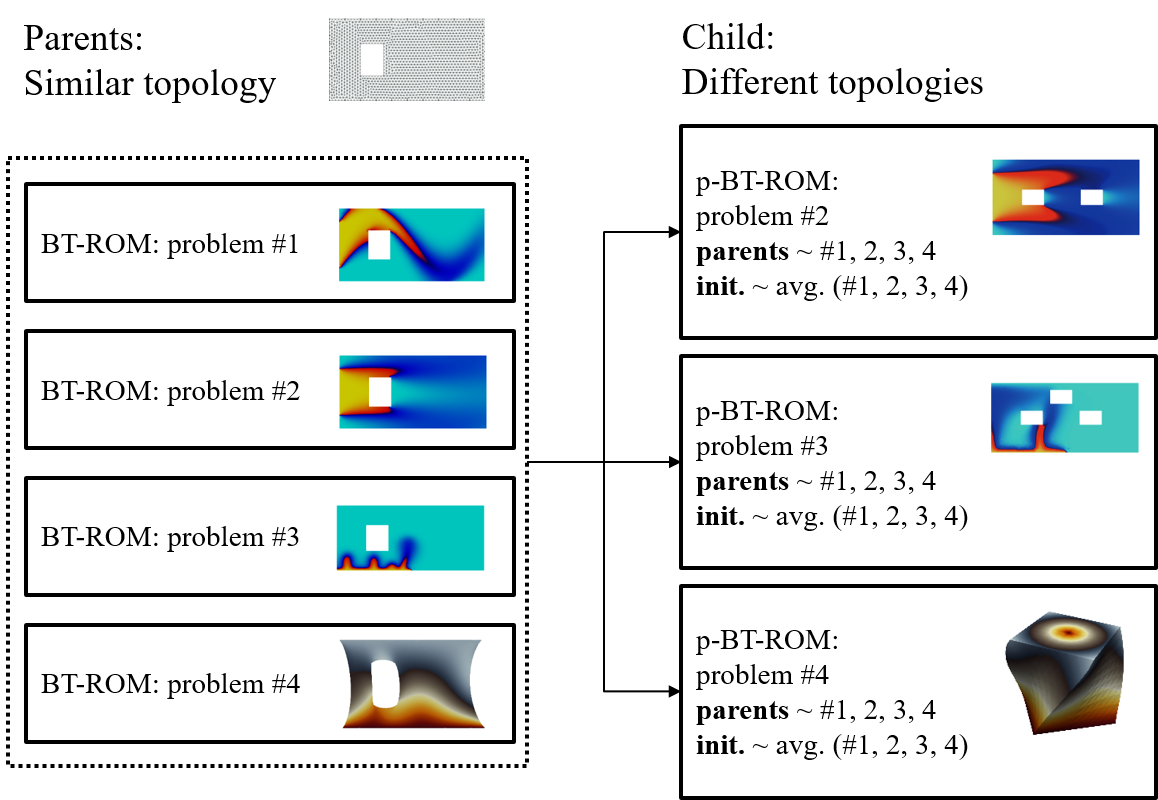}
  \caption{Different topologies - schematic of p-BT-ROM specifies child$-$parent(s) relationship. See Supplementary fig. \ref{fig:sum_geo} for more information on the geometries.}
  \label{fig:diff_topo_scheme}
\end{figure}

\subsubsection*{Transport in porous media - Problem \#2}\label{sec:diff_topo_problem2}

Here, we focus on a physics problem of transport in porous media - Problem \#2 (\ref{sec:physics_problems_td}) with mesh shown in Supplementary fig. \ref{fig:sum_geo}b. A sample of snapshots is presented in Supplementary fig. \ref{fig:diff_topo_samples}a. We have one primary variable, saturation. We use a small training set of $\mathrm{M}N_t = 1443$ ($\mathrm{M} = 3$), large training set of $\mathrm{M}N_t = 9620$ ($\mathrm{M} = 20$), and a test set of $\mathrm{M_{test}}N_t = 4810$ ($\mathrm{M_{test}} = 10$). We present the mean squared error (MSE), \ref{eq:mse}, and mean absolute error (MAE), \ref{eq:mae}, results in Figs. \ref{fig:mse_mae_diff_topo}a-b. From these plots, the p-BT-ROM with 4 parents with the small training set performs much better than the BT-ROM (0 parent) with the small training set. Besides, the p-BT-ROM with 4 parents' performance is very similar to that of the BT-ROM (0 parent) with the large training set while using eight times fewer training samples. We present the MAE results as a function of parameter $\bm{\mu}$ in Figs. \ref{fig:mae_as_mu_diff_topo}a-b. The pattern of these three figures is similar (i.e., the shape), but the magnitude of the MAE is different. This observation implies that the p-BT-ROM only alters the magnitude of the error, not its pattern. \par

\begin{figure}[!ht]
  \centering
  \includegraphics[width=14.0cm,keepaspectratio]{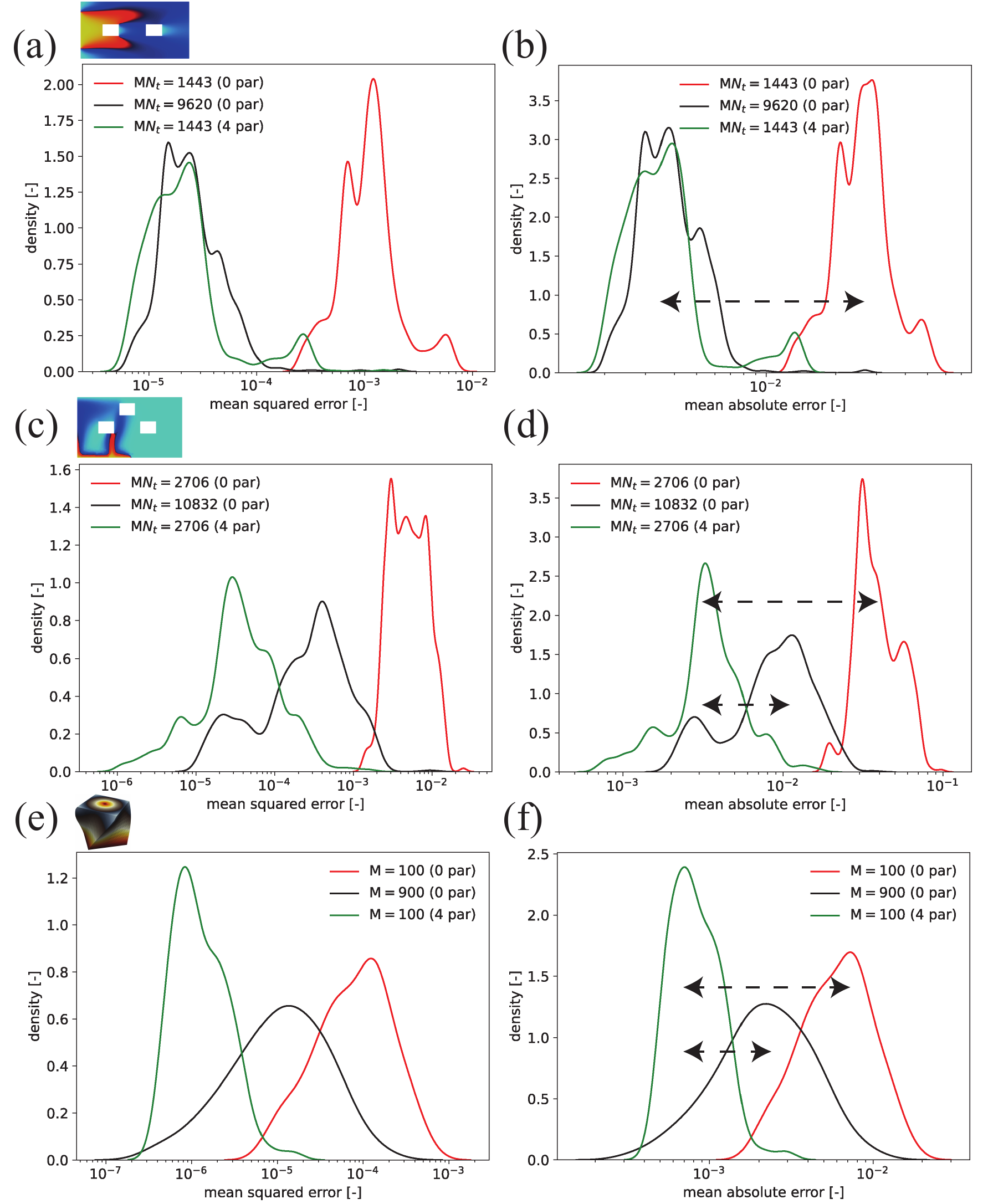}
  \caption{Different topologies - (a) mean squared error (MSE) and (b) mean absolute error (MAE) results for Problem \#2 (\ref{sec:physics_problems_td}) using the topology shown in Supplementary fig. \ref{fig:sum_geo}b (for $\mathrm{M}N_t = 1443$ ($\mathrm{M} = 3$) and $\mathrm{M}N_t = 9620$ ($\mathrm{M} = 20$)), (c) mean squared error (MSE) and (d) mean absolute error (MAE) results for Problem \#3 (\ref{sec:physics_problems_el}) using the topology shown in Supplementary fig. \ref{fig:sum_geo}c (for $\mathrm{M}N_t = 2706$ ($\mathrm{M} = 5$) and $\mathrm{M}N_t = 10832$ ($\mathrm{M} = 20$)), and (e) mean squared error (MSE) and (f) mean absolute error (MAE) results for Problem \#4 (\ref{sec:physics_problems_hy}) using the topology shown in Supplementary fig. \ref{fig:sum_geo}d. We present the small training set in red for 0 parent and in green for 4 parents. Solid black lines represent the large training set. Dashed black lines show accuracy gain by including knowledge from parents: p-BT-ROM.}
  \label{fig:mse_mae_diff_topo}
\end{figure}


\subsubsection*{gravity-driven in porous media - Problem \#3}\label{sec:diff_topo_problem3}

Next, we focus on a physics problem of gravity-driven in porous media - Problem \#3 (\ref{sec:physics_problems_el}) with mesh shown in Supplementary fig. \ref{fig:sum_geo}c. A sample snapshot is illustrated in Supplementary fig. \ref{fig:diff_topo_samples}b. Here, we have three primary variables, fluid pressure, velocity, and temperature. We only focus on the fluid temperature in this case. We use a small training set of $\mathrm{M}N_t = 2706$ ($\mathrm{M} = 5$), a large training set of $\mathrm{M}N_t = 10832$ ($\mathrm{M} = 20$), and a test set of $\mathrm{M_{test}}N_t = 5346$ ($\mathrm{M_{test}} = 10$). We present the MSE and MAE results in Figs. \ref{fig:mse_mae_diff_topo}c-d. Again, the p-BT-ROM with 4 parents trained with the small training set delivers much better results than those of the BT-ROM (0 parent) trained with the small training set. Moreover, the p-BT-ROM with 4 parents trained with the small training set outperforms the BT-ROM (0 parent) trained with the large training set (approximately five times larger). Similar to the previous section, the pattern of MAE results as a function of parameter $\bm{\mu}$ are also similar between the p-BT-ROM (4 parents) and BT-ROM (0 parent), see Supplementary figs. \ref{fig:mae_as_mu_diff_topo}c-d. \par


\subsubsection*{Hyperelasticity problem - Problem \#4}\label{sec:diff_topo_problem4}

Lastly, we move on to a physics problem of finite deformation of hyperelastic material - Problem \#4 (\ref{sec:physics_problems_hy}) with mesh shown in Supplementary fig. \ref{fig:sum_geo}d, one sample snapshot is shown in Supplementary fig. \ref{fig:diff_topo_samples}c. We want to emphasize that not only do we test a setting where all parents are fluid mechanics problems but the child is solid mechanics problems, but we also want to illustrate that we can use 2-Dimensional meshes for the parents but 3-Dimensional mesh for the child. We have only one primary variable, displacement, and we focus on its magnitude of it. We use a small training set of $\mathrm{M} = 100$, a large training set of $\mathrm{M} = 1600$, and a test set of $\mathrm{M_{test}} = 100$. We present the MSE and MAE results in Figs. \ref{fig:mse_mae_diff_topo}e-f and the pattern of MAE results as a function of parameter $\bm{\mu}$ in Supplementary figs. \ref{fig:mae_as_mu_diff_topo}e-f. Again, the p-BT-ROM with 4 parents trained with the small training set delivers much better results than those of the BT-ROM (0 parent) trained with the small and large training sets. The pattern  of MAE results as a function of parameter $\bm{\mu}$ are also not much different between the p-BT-ROM (4 parents) and BT-ROM (0 parent) - see Supplementary figs. \ref{fig:mae_as_mu_diff_topo}e-f (i.e., only the shape of the distribution but the magnitude of the MAE is drastically different.). \par


\section*{Discussion} \label{sec:discussion}

By conducting a series of numerical experiments, we have successfully demonstrated the effectiveness of a progressive reduced order model. The core element of this framework is centered around gates used to regulate the flow of information from previously trained models. These gates are crucial in managing information flow while simultaneously training the current model. The total trainable parameters encompass a combination of the trainable parameters from the current model and those from all the gates associated with it. Each gate corresponds to a connection between the current model and a preceding model, with a dedicated gate for each connection. During the training process, if information from a prior model proves advantageous, meaning it reduces the loss function, the corresponding gate increases the flow of information. Conversely, if the information is detrimental, the gate reduces the flow. In this context, we employ a linear layer as a gate, which means that the $\mathbf{W}$ and $\mathbf{b}$ parameters of each gate are adjusted in response to evaluating the loss function. It's important to note that the complexity of these gates can be enhanced by utilizing techniques such as recurrent units \cite{huang2023gate}, attention mechanisms \cite{vaswani2017attention}, or adaptive algorithms \cite{alesiani2023gated}.

The outcomes of our experiments are succinctly presented in Tab. \ref{tab:summary_results}. These results provide compelling evidence that leveraging the knowledge gained from previously trained models offers a promising solution to address the limited availability of training data. Notably, our framework, which incorporates 4 parent models, surpasses its counterpart that lacks such parental guidance, even when the latter is trained with 9 times more data. This evidence highlights the remarkable potential of utilizing prior knowledge from trained models to enhance performance. Moreover, our research showcases the feasibility of transferring knowledge across diverse physics domains, such as fluid mechanics, to solid mechanics problems. Additionally, we demonstrate the applicability of our approach to different domain topologies, including transitions from 2-Dimensional to 3-Dimensional domains and variations in the number of holes inside a 2-Dimensional domain. Our findings underscore the considerable benefits of harnessing previously acquired knowledge within a progressive reduced order model, enabling improved performance and knowledge transfer across various domains and topologies.



\begin{table}[htbp]
  \centering
  \caption{Summary of results for test sets. }
    \begin{tabular}{|c|c|c|c|c|c|l|}
    \toprule
    \multicolumn{2}{|c|}{setting} & num. parent(s) & \multicolumn{1}{c|}{training samples} & $\operatorname{avg.}{\mathrm{MAE}}$   & $\operatorname{std.}{\mathrm{MAE}}$   & \multicolumn{1}{c|}{remark} \\
    \midrule
    \multirow{9}[18]{*}{\begin{sideways}similar topology\end{sideways}} & \multirow{4}[8]{*}{Problem \#3} & 0     &  $\mathrm{M} = 40$ ($\mathrm{M}N_t = 28966$)      & 2.41E-03 & 1.27E-03 & large training set \\
\cmidrule{3-7}          &       & 0     &  $\mathrm{M} = 5$ ($\mathrm{M}N_t = 3623$)     & 1.61E-02 & 5.83E-03 & small training set \\
\cmidrule{3-7}          &       & 1     &   $\mathrm{M} = 5$ ($\mathrm{M}N_t = 3623$)     & 5.87E-03 & 2.75E-03 & \multirow{2}[4]{*}{from transport to gravity-driven problems} \\
\cmidrule{3-6}          &       & 2      &    $\mathrm{M} = 5$ ($\mathrm{M}N_t = 3623$)   & 4.82E-03 & 2.36E-03 &  \\
\cmidrule{2-7}          & \multirow{5}[10]{*}{Problem \#4} & 0     &    $\mathrm{M} = 1600$     & 1.08E-04 & 4.84E-05 & large training set \\
\cmidrule{3-7}          &       & 0     &  $\mathrm{M} = 100$     & 3.69E-03 & 1.93E-03 & small training set \\
\cmidrule{3-7}          &       & 1      &  $\mathrm{M} = 100$     & 3.79E-04 & 7.19E-05 & \multirow{3}[6]{*}{from fluid to solid problems} \\
\cmidrule{3-6}          &       & 2     &  $\mathrm{M} = 100$     & 3.70E-04 & 6.52E-05 &  \\
\cmidrule{3-6}          &       & 3     &   $\mathrm{M} = 100$    & 3.20E-04 & 4.40E-05 &  \\
    \midrule
    \multirow{9}[18]{*}{\begin{sideways}different topologies\end{sideways}} & \multirow{3}[6]{*}{Problem \#2} & 0     &  $\mathrm{M} = 20$ ($\mathrm{M}N_t = 9620$)     & 3.99E-03 & 1.64E-03 & large training set \\
\cmidrule{3-7}          &       & 0     &  $\mathrm{M} = 3$ ( $\mathrm{M}N_t = 1443$)     & 2.68E-02 & 7.77E-03 & small training set \\
\cmidrule{3-7}          &       & 4     &  $\mathrm{M} = 3$ ( $\mathrm{M}N_t = 1443$)       & 4.01E-03 & 2.48E-03 & from 1 hole to 2 holes topologies \\
\cmidrule{2-7}          & \multirow{3}[6]{*}{Problem \#3} & 0     & $\mathrm{M} = 20$ ($\mathrm{M}N_t = 10832$)      & 9.77E-03 & 5.25E-03 & large training set \\
\cmidrule{3-7}          &       & 0     &  $\mathrm{M} = 5$ ($\mathrm{M}N_t = 2706$)     & 4.06E-02 & 1.32E-02 & small training set \\
\cmidrule{3-7}          &       & 4     &   $\mathrm{M} = 5$ ($\mathrm{M}N_t = 2706$)    & 3.84E-03 & 2.16E-03 & from 1 hole to 3 holes topologies \\
\cmidrule{2-7}          & \multirow{3}[6]{*}{Problem \#4} & 0     &  $\mathrm{M} = 1600$      & 2.64E-03 & 1.73E-03 & large training set \\
\cmidrule{3-7}          &        & 0     &  $\mathrm{M} = 100$     & 6.66E-03 & 3.17E-03 & small training set \\
\cmidrule{3-7}          &       & 4     &  $\mathrm{M} = 100$     & 8.79E-04 & 3.61E-04 & from 2- to 3-Dimensional topologies \\
    \bottomrule
    \end{tabular}%
    \flushleft
child$\sim$parent(s) relationship can be found in Supplementary figs. \ref{fig:pnn_elder_s_topo_sep} and \ref{fig:pnn_hyper_s_topo_sep} for the similar topology cases, and Fig. \ref{fig:diff_topo_scheme} for different topologies cases.
  \label{tab:summary_results}%
\end{table}%

However, there is an additional computational cost associated with this benefit. As the number of parents increases, the number of trainable parameters also grows, which requires more computational resources in terms of time and space complexities (\ref{sec:num_para}). It's important to note that this additional cost primarily affects the training process (the prediction or inferring phase remains speedy evaluation). For the specific case illustrated in Fig. \ref{fig:prelim_results} and Tab. \ref{tab:count_params_s_topo}, when training the model with 0 and 3 parents using NVIDIA Quadro RTX 8000, there is an approximate increase of 10\% in the wall time required. However, the prediction time remains very similar, with both cases taking less than a second. In contrast, running a single high-fidelity model for Problem \#3 would typically take at least 30 minutes. \par

It is worth mentioning that the framework we have developed, known as progressive reduced order model concept, can be applied to various machine-learning architectures and is not limited to the Barlow Twins reduced order model (p-BT-ROM) alone. Moreover, our framework is not constrained to any specific application domain, such as fluid or solid mechanics, and can be utilized for any physical problem. Additionally, it is adaptable to different sources of data, including high-fidelity numerical simulations or field measurements. This flexibility is highly advantageous as the framework can accumulate knowledge from diverse sources, enhancing its overall performance. \par

Even though this study illustrates the potential of growing our data-driven model via the concept of progressive neural networks \cite{rusu2016progressive}, six limitations need to be acknowledged and discussed. The first one is more complex geometries and topologies should be used to test this framework. Our framework does not have direct information on topologies at this stage. Still, one could use an additional network such as (graph-)trunk nets \cite{lu2021learning} to embed the topologies of each parent and child into the framework. Secondly, as mentioned in \ref{sec:data_generation} we uniformly sample our parameter space $\bm{\mu}$; subsequently, one might argue that if we perform a smart sampling or adaptive sampling \cite{choi2020gradient}, we might also end up using less training samples similarly to p-BT-ROM. Hence, one also wants to combine our progressive framework with an adaptive sampling technique to reduce training samples further (or maintain the number of training samples but improve the models' accuracy). \par

Thirdly, it is worth noting that all the physics problems discussed here, labeled as Problems \#1 to \#4, operate on the same physical scale in terms of computational size. However, future investigations should consider incorporating information from different scales, such as integrating pore-scale and field-scale data. To control the flow of information across these scales, adapting an attention mechanism \cite{vaswani2017attention} could prove beneficial. 

Fourthly, it is essential to acknowledge that increasing the number of parents in our framework leads to a higher number of trainable parameters, as explained in \ref{sec:num_para}. This results in a more intricate and computationally expensive training process. To address this challenge, a potential solution could involve employing a reinforcement learning technique to choose parents that are relevant for the current generation selectively. This approach would help optimize the selection of parents, ensuring that only the most useful models contribute to training the reduced order model (ROM) for a specific application, such as $\mathrm{CO_2}$ storage.

For instance, in Fig. \ref{fig:future_plan}, we have multiple trained models available in our inventory. In the case of developing a ROM for $\mathrm{CO_2}$ storage, a straightforward approach would be to utilize all the trained models (i.e., six parents) as parents for the $\mathrm{CO_2}$ storage ROM. However, considering the increased number of trainable parameters with more parents, a reinforcement learning (or meta-learning) framework can be employed to determine which parents are truly influential for the $\mathrm{CO_2}$ storage ROM. By selectively choosing and utilizing only the relevant ROMs as parents, such as excluding the ROM for the contact problem if it does not contribute significantly to the performance of the $\mathrm{CO_2}$ storage ROM, we can optimize the training process and streamline the computational resources.


\begin{figure}[!ht]
  \centering
  \includegraphics[width=12.0cm,keepaspectratio]{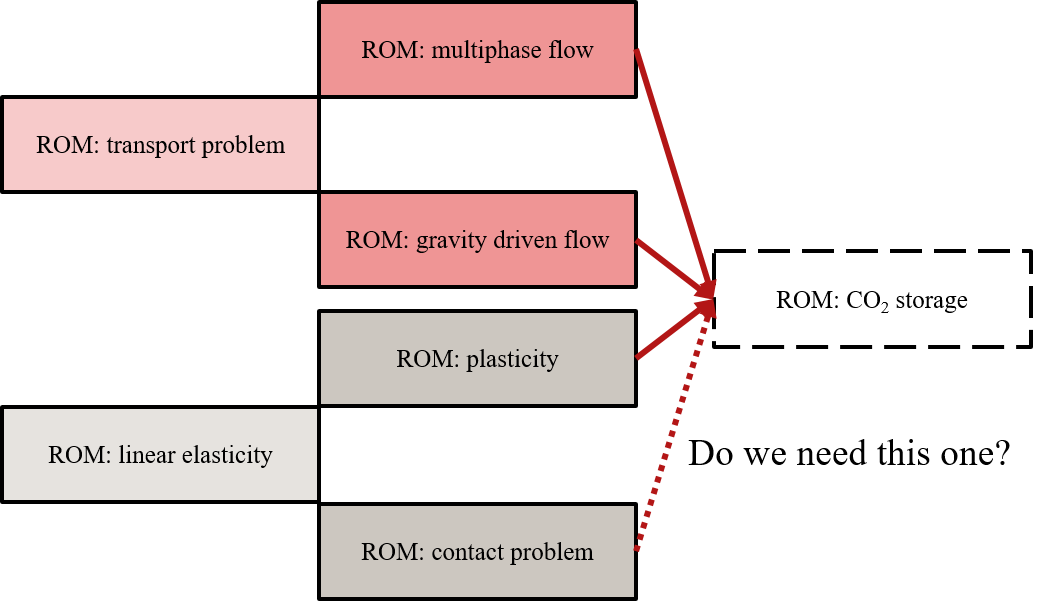}
  \caption{A hypothetical situation where we want to develop a ROM of $\mathrm{CO_2}$ storage and have trained ROMs of linear elasticity, contact problem, plasticity, transport problem, gravity-driven flow, and multiphase flow in our inventory that we can pick as (grand)parent(s) of $\mathrm{CO_2}$ storage ROM. We use red to represent fluid problems and grey to show solid problems. Lighter shades are used to represent grandparents, while darker ones are for parents.}
  \label{fig:future_plan}
\end{figure}


The fourth challenge also involves reaching knowledge saturation. Based on our observations, simply increasing the number of parents does not consistently lead to enhanced model performance. To illustrate, as shown in Fig. \ref{fig:prelim_results_diff_topo}, having three or four parents yields roughly equivalent results. A potential solution to this issue could involve employing a meta-learning framework (as well as reinforcement learning) to select relevant parents.

The fifth challenge revolves around integrating physical information, either by including an additional loss function or by modifying the network architecture itself. This entails finding ways to embed fundamental physical principles into the learning process. Given that our framework strictly relies on data, it might not yield a genuinely physical solution and could generate predictions lacking a physical basis. To overcome this issue for each individual component (each child or parent), one can employ the concept of physics-informed neural networks \cite{raissi2019physics}. However, when dealing with a combination of parents and children, each influenced by different sets of physical phenomena, this framework may not be applicable. In such cases, it might be beneficial to explore transfer learning techniques using physics-guided models \cite{oommen2022learning}.

Lastly, it is important to highlight that our progressive learning framework follows a strict hierarchical structure. To elaborate, consider Fig. \ref{fig:future_plan}. If we select ROM: multiphase flow as a parent for the ROM: $\mathrm{CO_2}$ storage, it automatically implies that ROM: transport problem becomes a grandparent. In such a scenario, if we decide to retrain ROM: transport problem, it necessitates the retraining of ROM: multiphase flow, followed by ROM: $\mathrm{CO_2}$ storage. This strict hierarchical growth can introduce complexities and dependencies in the training process.

To address this challenge, one possible approach is to utilize a directed acyclic graph (DAG) that allows for information exchange among different sources without assuming a strict hierarchical connection. The use of a DAG can provide more flexibility in incorporating information from multiple models or sources, eliminating the need for a fixed hierarchical structure. This approach has been explored in the context of machine learning models, as demonstrated in the work of Gorodetsky et al. \cite{gorodetsky2021mfnets}. By adopting a DAG-based approach, we can mitigate the problem of strict hierarchical dependencies and enhance the adaptability and scalability of our framework.

\section*{Methodology}\label{sec:method}

We propose the use of progressive learning applied in the context of progressive ROMs. We note our framework can be applied to any data-driven ROMs; however, here in our application, each machine learning model corresponds to the Barlow-Twins reduced order model (BT-ROM), as described in \ref{sec:si_btrom} and detailed in \cite{kadeethum2022reduced}. In Fig. \ref{fig:p_bt_scheme}, we present our progressive BT-ROM (p-BT-ROM) consisting of three columns. Each column represents a model, with two previously trained models referred to as parent 1 ($c_1$) and parent 2 ($c_2$), and the current model referred to as the child ($c_3$). It is important to note that in practice, the p-BT-ROM can be extended to a finite number of columns based on the available machine memory. The relationship between the parents and child in the p-BT-ROM is not limited to a hierarchical structure, where parent 2 is the child of parent 1. The parents can also be independent, forming standalone BT-ROMs without lateral connections. However, the child can still establish lateral connections with each parent individually, building separate relationships with them. \par

By employing progressive learning in the p-BT-ROM, we facilitate the transfer of knowledge and information from the previously trained models to enhance the performance and learning capability of the current model. This approach allows us to leverage prior knowledge effectively and overcome the limitations of one-to-one transfer and catastrophic forgetting. \par

\begin{figure}[!ht]
  \centering
  \includegraphics[width=18.0cm,keepaspectratio]{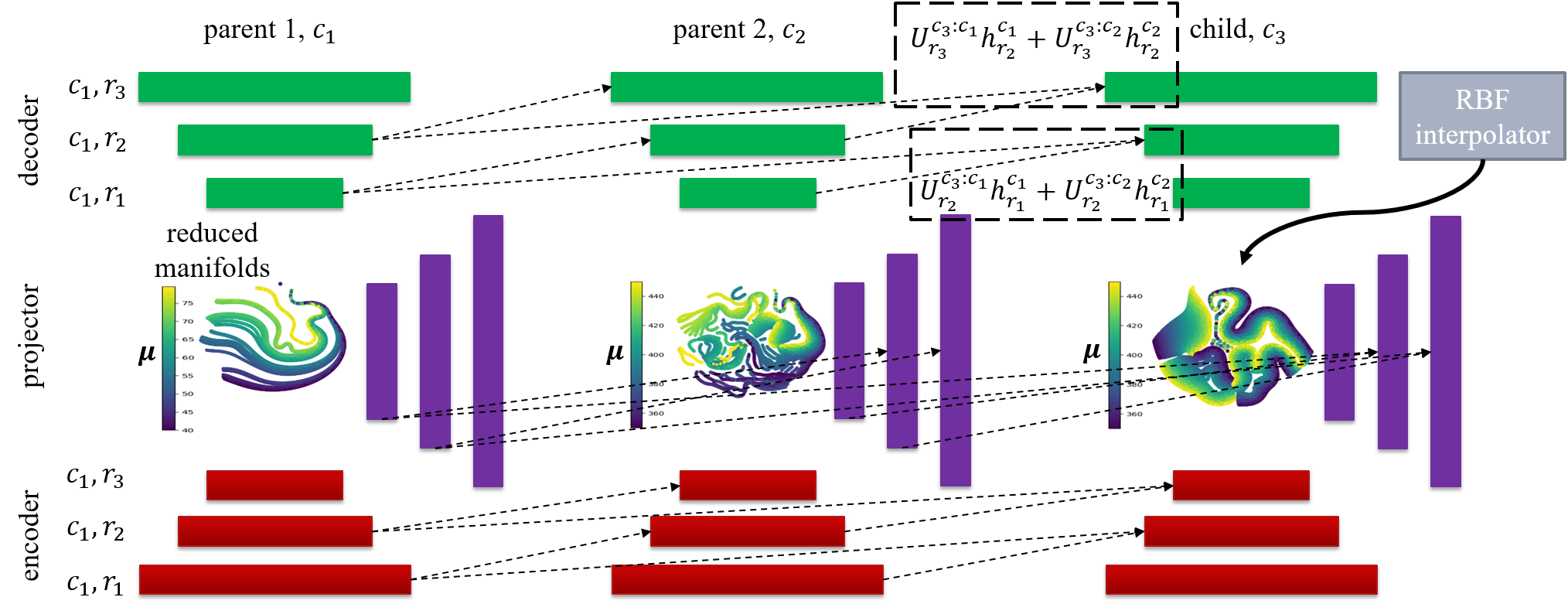}
  \caption{A schematic of p-BT-ROM. Assuming we have three columns, $c_1$, $c_2$, and $c_3$ representing parent 1, parent 2, and child, respectively. Each column represents a BT-ROM; see \ref{sec:si_btrom} for a detailed description. Each BT-ROM consists of one encoder, one decoder, and one projector. Assuming each one of them has only three layers, $r_1$, $r_2$, and $r_3$. According to Eq. \eqref{eq:progressive_main}, for child $c_3$, $r_3$, we have a contribution from parent 1 ($c_1$) and parent 2 ($c_2$) as $U_{r_3}^{c_3: c_1} h_{r_2}^{c_1}+U_{r_3}^{c_3: c_2} h_{r_2}^{c_2}$, and child $c_3$, $r_2$, we have a contribution from parent 1 ($c_1$) and parent 2 ($c_2$) as $U_{r_2}^{c_3: c_1} h_{r_1}^{c_1}+U_{r_2}^{c_3: c_2} h_{r_1}^{c_2}$. The reduced manifolds plots are constructed using t-Distributed Stochastic Neighbor Embedding (t-SNE). Different colors represent each value of $\bm{\mu}$ value. We calculate the t-SNE plots using the Scikit-Learn package using its default setting and perplexity of 15.}
  \label{fig:p_bt_scheme}
\end{figure}

Each BT-ROM consists of one encoder, one decoder, and one projector, as shown in Fig. \ref{fig:p_bt_scheme}. Each of them has different numbers of layers or rows (i.e., $r_1, r_2, r_3, \dots, r_n$). We use $\mathcal{N}\left( \cdot \right)$ to represent any nonlinear activation functions (e.g., tanh, Sigmoid, or ReLU). $\mathcal{L}\left( \cdot \right)$ is a linear layer within each column (i.e., take the output from the previous layer within the same column, $h_{r_i-1}^{c_k}$, and perform a linear transformation with its $\mathbf{W}$ and $\mathbf{b}$). ${U}\left( \cdot \right)$ is also a linear layer that acts as a gate for each lateral connection (i.e., take the output from the previous layer of the previous columns, $h_{r_i-1}^{c_j}$ where $j<k$ and $k$ is the current column, and perform a linear transformation with its $\mathbf{W}$ and $\mathbf{b}$). To this end, our output from the $k$ column is read as follows

\begin{equation}\label{eq:progressive_main}
h_{r_i}^{c_k}=\mathcal{N}\left(\mathcal{L}_{r_i}^{c_k} \left(h_{r_i-1}^{c_k}\right)+\sum_{j<k}\left({U}_{r_i}^{{c_k}: {c_j}} \left( h_{r_i-1}^{c_j} \right)  \right)\right),
\end{equation}

\noindent
and for the first column (i.e., $c_k=0$)

\begin{equation}\label{eq:progressive_if_k_0}
h_i^{c_k=0}=\mathcal{N}\left(\mathcal{L}_i^{c_k=0} \left(h_{i-1}^{c_k=0} \right) \right).
\end{equation}

From these equations, one can see that if information from the previously trained models is not useful, ${U}\left( \cdot \right)$ will be zero out its output during the training process. Besides, this framework can be used to tackle its previous tasks by simply focusing its output from the focus column (i.e., immune to catastrophic forgetting). To elaborate, assuming we want to predict a job that parent 2 is trained for, we only need to ignore the outputs from parent 1 and child and only consider the output from parent 2. We note that each component of p-BT-ROM acts independently (i.e., there are lateral connections within the encoder but no lateral connections among the encoder, decoder, and projector).  \par






\section*{Acknowledgments}

This work is supported by the US Department of Energy Office of Fossil Energy and Carbon Management project - Science-Informed Machine Learning to Accelerate Real-Time Decisions in Subsurface Applications (SMART) initiative. Sandia National Laboratories is a multi-mission laboratory managed and operated by National Technology \& Engineering Solutions of Sandia, LLC (NTESS), a wholly owned subsidiary of Honeywell International Inc., for the U.S. Department of Energy’s National Nuclear Security Administration (DOE/NNSA) under contract DE-NA0003525. This written work is authored by an employee of NTESS. The employee, not NTESS, owns the right, title, and interest in and to the written work and is responsible for its contents. Any subjective views or opinions that might be expressed in the written work do not necessarily represent the views of the U.S. Government. The publisher acknowledges that the U.S. Government retains a non-exclusive, paid-up, irrevocable, world-wide license to publish or reproduce the published form of this written work or allow others to do so, for U.S. Government purposes. The DOE will provide public access to results of federally sponsored research in accordance with the DOE Public Access Plan.

\section*{Author contributions statement}

\textbf{T. Kadeethum}: Conceptualization, Formal analysis, Software, Validation, Writing - original draft, Writing - review \& editing. \textbf{D. O'Malley}: Conceptualization, Formal analysis, Supervision, Validation, Writing - review \& editing. \textbf{Y. Choi}: Conceptualization, Formal analysis, Supervision, Validation, Writing - review \& editing. \textbf{H.S. Viswanathan}: Conceptualization, Supervision, Writing - review \& editing. \textbf{H. Yoon}: Conceptualization, Formal analysis, Funding acquisition, Supervision, Writing - review \& editing.

\section*{Competing interests}
The authors declare no competing interests




\newpage

\beginsupplement

\section*{Supplementary information}

\section{A collections of geometries used in this manuscript} \label{sec:geo}

Here, we summarize all meshes used throughout this manuscript in Supplementary fig. \ref{fig:sum_geo} and Supplementary tab. \ref{tab:sum_geo}. In short, Supplementary fig. \ref{fig:sum_geo}a is a 2-Dimensional domain with a size of $2 \times 1$ and has a hole inside the domain. It has 1503 nodes and is used for all physics problems ( \ref{sec:physics_problems_tv} to \ref{sec:physics_problems_hy}). Supplementary fig. \ref{fig:sum_geo}b is also a 2-Dimensional domain with a size of $2 \times 1$ and has two holes inside the domain. It has 1698 nodes and is used for Problem \#2 (\ref{sec:physics_problems_td}). Supplementary fig. \ref{fig:sum_geo}c is a 2-Dimensional domain with a size of $2 \times 1$ and has three holes inside the domain. It has 620 nodes and is used for Problem \#3 (\ref{sec:physics_problems_el}). Supplementary fig. \ref{fig:sum_geo}d is a 3-Dimensional domain with a size of $1 \times 1 \times 1$ and has no hole inside the domain. It has 1330 nodes and is used for Problem \#4 (\ref{sec:physics_problems_hy}). We note that the details of each physics problem can be found in \ref{sec:physics_problems}.

\begin{figure}[!ht]
  \centering
  \includegraphics[width=8.5cm,keepaspectratio]{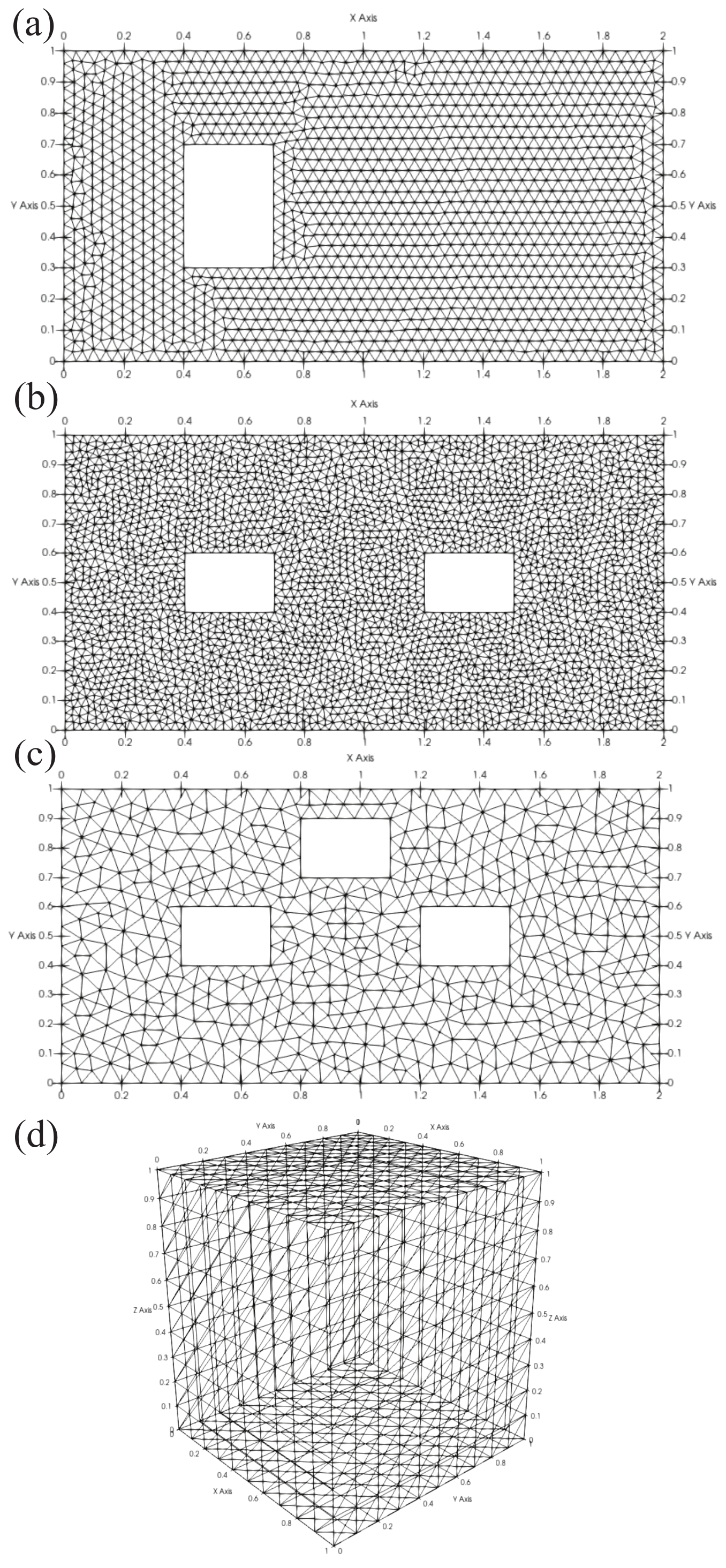}
  \caption{Illustrations of geometries and their corresponding mesh used in this manuscript. The details of each mesh can be found in Table \ref{tab:sum_geo}.}
  \label{fig:sum_geo}
\end{figure}

\begin{table}[!ht]
\centering
\caption{Summary of main information for each mesh.}
\begin{tabular}{|l|c|c|c|c|}
\hline
                                                                      & Dimension & Size      & \begin{tabular}[c]{@{}c@{}}Number of\\ nodes\end{tabular} & Used for              \\ \hline
\begin{tabular}[c]{@{}l@{}}A rectangular \\ with 1 hole\end{tabular}  & 2         & 2 x 1     & 1503                                                      & Problems \#1, 2, 3, 4 \\ \hline
\begin{tabular}[c]{@{}l@{}}A rectangular\\ with 2 holes\end{tabular}  & 2         & 2 x 1     & 1698                                                      & Problem \#2           \\ \hline
\begin{tabular}[c]{@{}l@{}}A rectangular \\ with 3 holes\end{tabular} & 2         & 2 x 1     & 620                                                       & Problem \#3           \\ \hline
A cube                                                                & 3         & 1 x 1 x 1 & 1330                                                      & Problem \#4           \\ \hline
\end{tabular}\label{tab:sum_geo}
\end{table}





\section{A collections of physical problems used in this manuscript} \label{sec:physics_problems}

For all problems, let $\Omega \subset \mathbb{R}^d$ ($d \in \{1,2,3\}$) denote the computational domain and $\partial \Omega$ denote the boundary. $\partial \Omega$ can be represented by the Dirichlet boundary condition ($\partial \Omega_D$) or the Neumann boundary condition ($\partial \Omega_N$). For Problems \# 1 to 3, the time domain is denoted by $\mathbb{T} = \left(0,\tau\right]$ with $\tau>0$ (i.e., $\tau$ is the final time). We note that Problem \# 4 is a steady-state problem.

\subsection{Problem \#1: transport problem with velocity as a parameter} \label{sec:physics_problems_tv}

In this problem, we focus on the transport problem, and we have a concentration $c : \Omega \times \mathbb{T} \rightarrow \mathbb{R}$ (\si{fraction}) as a primary variable. Our governing equations are shown below

\begin{equation} \label{eq:transport}
\begin{split}
\frac{\partial}{\partial t}(\phi c) + \nabla \cdot \left( \bm{q} c  \right)- \nabla \cdot \left( \bm{D} \nabla c \right) = q  &\mbox{ \: in \: } \Omega \times (0,\mathbb{T}], \\
\eta(\bm{q},c) \cdot \mathbf{n} = {c_{in}} \bm{q}\cdot \mathbf{n} &\mbox{ \: on \: } \partial \Omega_{{\rm in}} \times (0,\mathbb{T}],  \\
{\bm{D}} \nabla c \cdot \mathbf{n} = 0  &\mbox{ \: on \: } \partial \Omega_{{\rm out}} \times (0,\mathbb{T}], \\
c={c_{0}}  &\text{ \: in \: } \Omega \text { at } t = 0,
\end{split}
\end{equation}

\noindent
where $\phi$ is a porosity, $\bm{D}$ is a tensor of diffusivity coefficient, $q$ represents a source term, $c_{in}$ is the inflow concentration, $c_{0}$ is the initial concentration, $\mathbf{n}$ is a normal unit vector to each element surface, the mass flux $\eta(\bm{q},c)$ is defined as 

\begin{equation}
\eta(\bm{q},c) :=  \bm{q} c - {\bm{D}}(\phi) \nabla c.
\end{equation}

\noindent
$\bm{q}$ is the superficial velocity vector defined by $\bm{q} = (20.0, \mu_1 \cos (\pi x))$. To approximate $c_h$, we use the first-order discontinuous Galerkin approximation, and the detail of discretization and the FOM source codes can be found in \cite{kadeethum2021locally,kadeethum2021nonTH}. Here, we have one parameter $\bm{\mu} = (\mu_1 = [5.0, 25.0])$. The rest parameters are fixed as $\phi = 0.3$, $\bm{D} = (0.2, 0.0, 0.0, 0.2)$ (i.e., homogeneous and isotropic), and we enforce $c_{0} = 0.0$ and $c_{in} = 1.0$ on the left boundary. We use geometries and meshes shown in Supplementary fig. \ref{fig:sum_geo}. \par

\subsection{Problem \#2: Transport problem with diffusivity coefficient as a parameter} \label{sec:physics_problems_td}

Here, we also use the same governing equations shown in \eqref{eq:transport}. For fixed parameters, we use $\phi = 0.3$, $\bm{q} = (20.0, \cos (\pi x))$, and we enforce $c_{0} = 0.0$ and $c_{in} = 1.0$ on the left boundary. The parameter $\bm{\mu} = (\bm{D} = (\mu_1, 0.0, 0.0, \mu_1))$, again, homogeneous and isotropic, and $\mu_1= [0.1, 1.0]$. We use geometries and meshes shown in Figs. \ref{fig:sum_geo}a, b.

\subsection{Problem \#3: gravity-driven problem with Rayleigh number as a parameter} \label{sec:physics_problems_el}

For Problem \#3, we use a gravity-driven flow in porous media. We follow the FOM used in \cite{kadeethum2021nonTH, kadeethum2022reduced}, and here we briefly discuss the governing equations of this problem. Essentially, we deal with coupled PDEs of mass conservation and heat advection-diffusion equations. As presented in \cite{kadeethum2021nonTH, kadeethum2022reduced}, we write our system of equations in dimensionless form, and primary variables in this setting are $\bm{u} (\cdot , t) : \Omega \times  \mathbb{T} \to \mathbb{R}^d$, which is a vector-valued Darcy velocity (\si{dimensionless}), $p (\cdot ,  t) : \Omega \times  \mathbb{T} \to \mathbb{R}^d$, which is a scalar-valued fluid pressure (\si{dimensionless}), and $T (\cdot , t) : \Omega \times \mathbb{T} \to \mathbb{R}^d$, which is a scalar-valued fluid temperature (\si{dimensionless}). Time is denoted as $t$ (\si{dimensionless}). The system of our governing equations read

\begin{equation} \label{eq:mass_dimless}
\begin{split}
\bm{u}+\nabla p-\mathbf{y} \operatorname{Ra} T=0, &\text { \: in \: } \Omega \times \mathbb{T}, \\ 
\nabla \cdot \bm{u}=0,  &\text { \: in \: } \Omega \times \mathbb{T}, \\ 
p=p_{D} &\text { \: on \: } \partial \Omega_{p} \times \mathbb{T}, \\ 
\bm{u} \cdot \mathbf{n}=q_{D} &\text { \: on \:} \partial \Omega_{q} \times \mathbb{T}, \\ 
p=p_{0} &\text { \: in \: } \Omega \text { at } t = 0,
\end{split}
\end{equation}

\noindent
for the mass balance equation where $\partial \Omega_{p}$ and $\partial \Omega_{q}$ are the prescribed pressure and flux boundaries, respectively, and $\mathrm{Ra}$ is the Rayleigh number

\begin{equation} \label{eq:Ra}
\mathrm{Ra}:=\frac{g \alpha \kappa \Delta T^{*} H}{K}.
\end{equation}

\noindent
Subsequently, the heat advection-diffusion equation is described as

\begin{equation} \label{eq:temp_dimless}
\begin{split}
\frac{\partial T}{\partial t}+\bm{u} \cdot \nabla T-\nabla^{2} T - f_c=0,  &\mbox{ \: in \: } \Omega \times (0,\mathbb{T}], \\  
T = T_D  &\mbox{ \: on \: } \partial \Omega_{{T}} \times (0,\mathbb{T}], \\ 
(-\bm{u} T+\nabla T) \cdot \mathbf{n}={T_{\rm in}} \bm{u}\cdot \mathbf{n}  &\mbox{ \: on \: } \partial \Omega_{{\rm in}} \times (0,\mathbb{T}],  \\ 
\nabla T \cdot \mathbf{n} = 0  &\mbox{ \: on \: } \partial \Omega_{{\rm out}} \times (0,\mathbb{T}], \\ 
T={T_{0}}  &\text{ \: in \: } \Omega \text { at } t = 0,
\end{split}
\end{equation}

\noindent
where $\partial \Omega_{{T}}$ is prescribed temperature boundary, $\partial \Omega_{\rm in}$ and $\partial \Omega_{\rm out}$ denote inflow and outflow boundaries, respectively, defined as

\begin{equation}
\partial \Omega_{\rm in} := \{ \bm{X} \in \partial \Omega : \bm{u} \cdot \mathbf{n} < 0\} \quad \mbox{ and } \quad \partial \Omega_{\rm out} := \{ \bm{X} \in \partial \Omega : \bm{u} \cdot \mathbf{n} \geq 0\}.
\label{eq:in_and_out}
\end{equation}

The detail of discretization, similar to problems \#1 and 2 that we use discontinuous Galerkin approximation, as well as the FOM source codes, could be found in \cite{kadeethum2021locally,kadeethum2021nonTH}. We note that as our finite element solver utilizes an adaptive time-stepping \cite{kadeethum2021locally,kadeethum2021nonTH}, each snapshot may have a different number of time-steps $N^t$. We use geometries and meshes shown in Figs. \ref{fig:sum_geo}a, c. Here, we have one parameter $\bm{\mu} = (\mu_1 = \mathrm{Ra})$, and $\mathrm{Ra}= [350.0, 450.0]$.  \par

\subsection{Problem \#4: Hyperelasticity problem with external forces as a parameter} \label{sec:physics_problems_hy}

Problem \#4 focuses on the finite deformation of a hyperelastic material, which is applicable to many engineering applications, such as biomedical engineering and material science \cite{kumar2013contact, luo2016topology, nezamabadi2011solving}. The weak form of mechanical equilibrium equations in the reference configuration, where the domain before deformation is represented with $\Omega_0$ and after deformation with $\Omega_i$, where $i$ represents different states of deformation (i.e., $\bm{\delta} \mathbf{u} = \mathbf{u}_i - \mathbf{u}_0$), is

\begin{equation}\label{EqCase2HEPenaltyWF}
\int_{\Omega_0} \mathbf{P} : \nabla (\pmb{\delta} \mathbf{u} ) \, dV - \int_{\Omega_0} \mathbf{B} \cdot \pmb{\delta} \mathbf{u} \, dV s- \int_{\partial \Omega_N} \mathbf{T} \cdot \pmb{\delta} \mathbf{u} \, dS = 0, 
\end{equation}

\noindent
where $\mathbf{P}$ is the 1$^{\mathrm{st}}$ Piola Kirchhoff stress tensor, which is a function of Young's modulus ($\mathrm{E}$) and Poisson ratio ($\nu$). $\mathbf{B}$ and $\mathbf{T}$ are the body and traction forces, respectively, and $\mathbf{u}$ is a displacement vector (primary variable). To approximate $\mathbf{u}$ (i.e., $\mathbf{u}_h$), following \cite{kadeethum2022fomassistrom}, we use a discontinuous Galerkin approximation of the first order. We use PETSc SNES as a nonlinear solver, and MUMPS as a linear solver \cite{petsc-user-ref} with absolute and relative tolerances of $1 \times 10^{-6}$ and  $1 \times 10^{-16}$, respectively. We utilize a backtracking line search with slope descent parameter of $1 \times 10^{-4}$, initial step length of $1.0$, and quadratic order of the approximation. We utilize both 2- and 3-Dimensional domains for this problem.

\subsubsection{2-Dimensional domain}

First, we use geometry and mesh shown in Figs. \ref{fig:sum_geo}a-c. $\mathrm{E} = 100$ \si{MPa}, and $\nu = 0.46$. We set $\mathbf{B} = (0.0, 0.0)$ and $\mathbf{T} = (0.0, 0.0)$. We enforce $\mathbf{u} = (\mu_1, \mu_2)$ at the boundary condition on the face of where $y = 1.0$, top surface. Other faces have a roller boundary condition (i.e., no normal displacement). Our parameter space, here we have two parameters, $\bm{\mu} = (0.05\mu_1, 0.05\mu_2)$, and $\mu_1= [-1.0, 1.0]$ and $\mu_2 = [-1.0, 1.0]$.

\subsubsection{3-Dimensional domain}

Here, we use a geometry and mesh shown in Supplementary fig. \ref{fig:sum_geo}d, $\mathrm{E} = 10$ \si{Pa}, and $\nu = 0.3$. We set $\mathbf{B} = (0.0, -0.5, 0.0)$ and $\mathbf{T} = ({\mu_1},  0.0, 0.0)$. We enforce $\mathbf{u} = (0.0,  0.0, 0.0)$ at the boundary condition on the face of where $x = 0.0$, and

\begin{equation}
\begin{aligned}
\mathbf{u}=&(0.0,\\
&\mu_2(0.5+(y-0.5) \cos (\pi / 3)-(z-0.5) \sin (\pi / 3)-y) / 2, \\
&\mu_2(0.5+(y-0.5) \sin (\pi / 3)+(z-0.5) \cos (\pi / 3)-x)) / 2)
\end{aligned}
\end{equation}

\noindent
at the boundary condition on the face of where $x = 1.0$. Other faces have a roller boundary condition (i.e., no normal displacement). Our parameter space is similar to the 2-Dimensional case as we have two parameters, $\bm{\mu} = (\mu_1, \mu_2)$, and $\mu_1= [0.1, 0.9]$ and $\mu_2 = [0.1, 0.9]$.




\section{A collections of reduced order model used in this manuscript} \label{sec:rom}

\subsection{Reduced order modeling with Barlow Twins self-supervised learning}\label{sec:si_btrom}

\cite{kadeethum2022reduced} has proposed a unified data-driven ROM, BT-ROM, that (1) bridges the performance gap between the linear and nonlinear manifold approaches and (2) can operate on unstructured meshes, which provides flexibility in its application to standard numerical solvers, on-site measurements, or experimental data. Even though it has been extended to boosting BT-ROM (BBT-ROM) and BT-ROM with uncertainty quantification (UQ-BT-ROM) \cite{kadeethum2022fomassistrom, kadeethum2022uqbtrom}, throughout this study, we will focus only on the BT-ROM variation.  

The summary of BT-ROM is shown in Supplementary fig. \ref{fig:bbt_explain}. Our BT-ROM starts with an initialization of the training set $\bm{\mu}$, our parameter space. Subsequently, we query a FOM (see \ref{sec:physics_problems}) for each parameter $\bm{\mu}$ in the training set. We note that the same procedures apply for validation and testing sets, but we will only discuss the training set for brevity. The third step entails a data compression stage through training BT-AE developed in \cite{kadeethum2022reduced}. 

\begin{figure}[!ht]
  \centering
    \includegraphics[width=14.5cm,keepaspectratio]{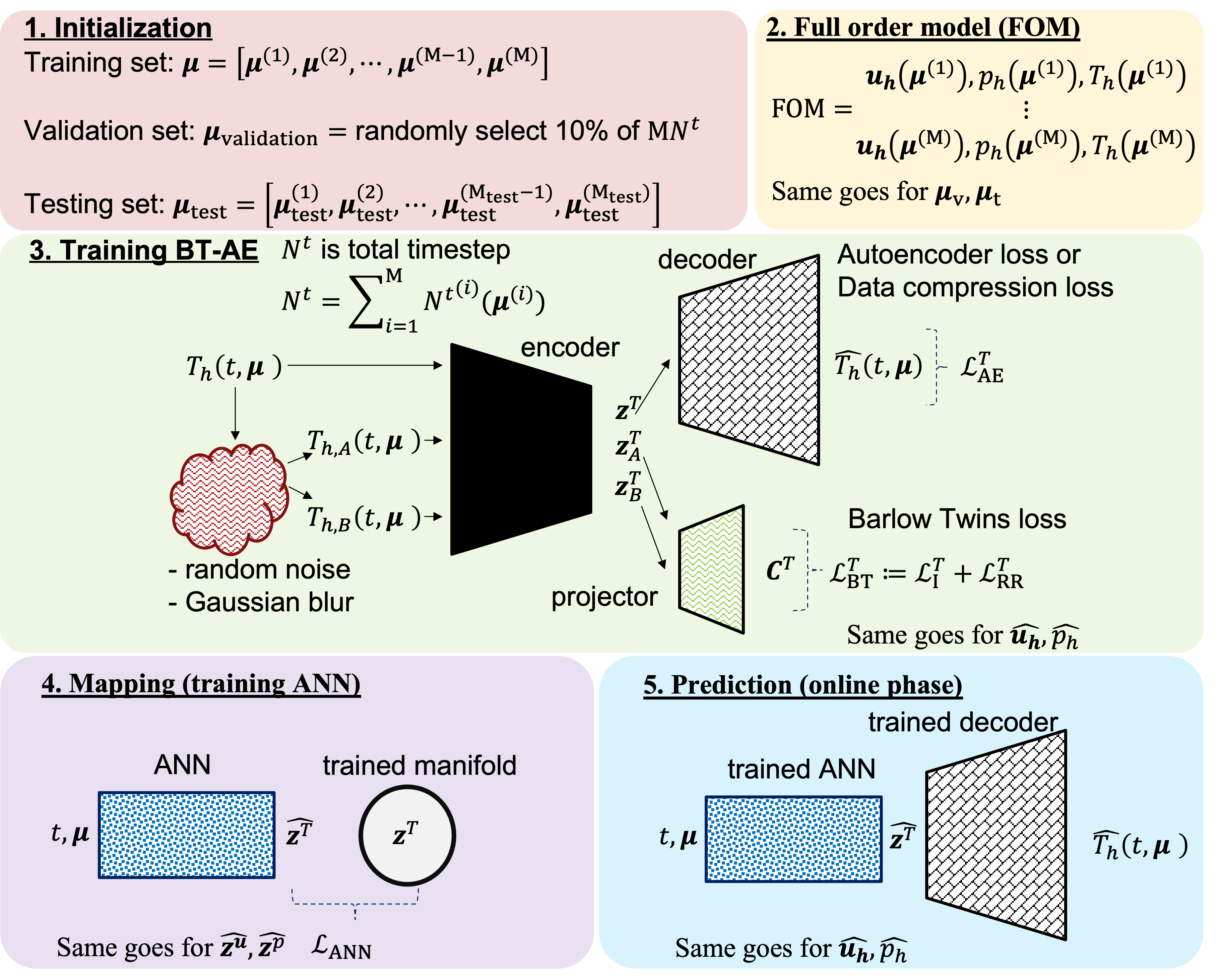}
  \caption{The summary of procedures taken to establish the proposed BT-AE}
  \label{fig:bbt_explain}
\end{figure}

We illustrate our algorithm in Al. \ref{al:ae-bt}, but for further details and implementation. Here, we only summarize the main components; please refer to \cite{kadeethum2022reduced, kadeethum2022fomassistrom}. The training involves two tasks; the first is the training of BT (encoder and projector) - the outer loop with $\mathbf{B}_{\mathrm{outer}}$. The second sub-task is the training of AE (encoder and decoder) - the inner loop with $\mathbf{B}_{\mathrm{inner}}$. First, we alter our training set by creating $\bm{X}_{h,A}\left(t, \bm{\mu}\right)$ and $\bm{X}_{h,B}\left(t, \bm{\mu}\right)$ from $\bm{X}\left(t, \bm{\mu}\right)$) through the addition of random noise

\begin{equation}\label{eq:add_noise}
\widetilde{\bm{X}_{h,A}}\left(t, \bm{\mu}\right), \widetilde{\bm{X}_{h,B}}\left(t, \bm{\mu}\right) = \bm{X}\left(t, \bm{\mu}\right) + \epsilon \operatorname{SD}\left(\bm{X}\left(t, \bm{\mu}\right)\right) \mathcal{G}\left(0,1\right)
\end{equation}

\noindent
where $\widetilde{\bm{X}_{h,A}}\left(t, \bm{\mu}\right), \widetilde{\bm{X}_{h,B}}\left(t, \bm{\mu}\right)$ are altered $\bm{X}\left(t, \bm{\mu}\right)$ by \eqref{eq:add_noise}. The constant $\epsilon$, which is set to $0.1$, determines the noise level, and $\mathcal{G}\left(0,1\right)$ is a random value that is sampled from the standard normal distribution with mean and standard deviation of zero and one, respectively. Subsequently, we pass $\widetilde{\bm{X}_{h,A}}\left(t, \bm{\mu}\right), \widetilde{\bm{X}_{h,B}}\left(t, \bm{\mu}\right)$ through Gaussian blur operation

\begin{equation}\label{eq:gb}
{\bm{X}_{h,A}}\left(t, \bm{\mu}\right), {\bm{X}_{h,B}}\left(t, \bm{\mu}\right)=\frac{1}{\sqrt{2 \pi \operatorname{SD}\left(\widetilde{\bm{X}_{h,A}}\left(t, \bm{\mu}\right), \widetilde{\bm{X}_{h,B}}\left(t, \bm{\mu}\right)\right)^{2}}} \mathrm{exp}\left({-\frac{{\widetilde{\bm{X}_{h,A}}\left(t, \bm{\mu}\right), \widetilde{\bm{X}_{h,B}}\left(t, \bm{\mu}\right)}^{2}}{2 \operatorname{SD}\left(\widetilde{\bm{X}_{h,A}}\left(t, \bm{\mu}\right), \widetilde{\bm{X}_{h,B}}\left(t, \bm{\mu}\right)\right)^{2}}}\right)
\end{equation}

\noindent
to obtain $\bm{X}_{h,A}\left(t, \bm{\mu}\right)$ and $\bm{X}_{h,B}\left(t, \bm{\mu}\right)$. Next, we pass $\bm{X}_{h,A}\left(t, \bm{\mu}\right)$ and $\bm{X}_{h,B}\left(t, \bm{\mu}\right)$ to the encoder (it is noted we have only one encoder) resulting in $\bm{z}_A^{\bm{X}}\left(t, \bm{\mu}\right)$ and $\bm{z}_B^{\bm{X}}\left(t, \bm{\mu}\right)$. We then use $\bm{z}_A^{\bm{X}}\left(t, \bm{\mu}\right)$ and $\bm{z}_B^{\bm{X}}\left(t, \bm{\mu}\right)$ as an input to the projector resulting in the cross-correlation matrix $\mathbf{C}^{\bm{X}}\left(t, \bm{\mu}\right)$. $\mathbf{C}^{\bm{X}}\left(t, \bm{\mu}\right)$ is a square matrix with the dimensionality of the projector's output. The Barlow Twins loss $\mathcal{L}_{\mathrm{BT}}^{\bm{X}}$, BT loss, is then calculated using

\begin{equation} \label{eq:loss_bt}
\mathcal{L}_{\mathrm{BT}}^{\bm{X}} := \mathcal{L}_{\mathrm{I}}^{\bm{X}} + \mathcal{L}_{\mathrm{RR}}^{\bm{X}}
\end{equation}

\noindent
where
\begin{equation} \label{eq:loss_I}
\mathcal{L}_{\mathrm{I}}^{\bm{X}} := \sum_{i}\left(1-\mathbf{C}_{i i}^{\bm{X}}\left(t, \bm{\mu}\right)\right)^{2},
\end{equation}

\noindent
and
\begin{equation} \label{eq:loss_RR}
\mathcal{L}_{\mathrm{RR}}^{\bm{X}} := \lambda \sum_{i} \sum_{j \neq i} {\mathbf{C}_{i j}^{\bm{X}}\left(t, \bm{\mu}\right)}^{2}.
\end{equation}

\noindent
Here, $\mathbf{C}_{i i}^{\bm{X}}\left(t, \bm{\mu}\right)$ denotes the $i$-th diagonal entry of $\mathbf{C}^{\bm{X}}\left(t, \bm{\mu}\right)$, $\lambda$ is set to $5 \times 10^{-3}$, and $\mathbf{C}_{i j}^{\bm{X}}$ are off-diagonal entries of $\mathbf{C}^{\bm{X}}$.

Subsequently, we obtain $\bm{z}^{\bm{X}}\left(t, \bm{\mu}\right)$ by passing $\bm{X}_h\left(t, \bm{\mu}\right)$ to the encoder. The $\bm{z}^{\bm{X}}\left(t, \bm{\mu}\right)$ is used to reconstruct $\widehat{\bm{X}}_h\left(t, \bm{\mu}\right)$ through the decoder, and we calculate our data compression loss or AE loss ($\mathcal{L}_\mathrm{AE}^{\bm{X}}$) using 

\begin{equation} \label{eq:loss_ae}
{\mathcal{L}_\mathrm{AE}^{\bm{X}}} := 
    {\mathrm{MSE}^{\bm{X}}} = 
    \frac{1}{\mathrm{M} N^t} \sum_{i=1}^{\mathrm{M}}\sum_{k=0}^{N^t}\left|\widehat{\bm{X}}_h\left(t^k, \bm{\mu}^{(i)}\right)-\bm{X}_h\left(t^k, \bm{\mu}^{(i)}\right)\right|^{2}.
\end{equation}

We use the adaptive moment estimation (ADAM) algorithm \cite{kingma2014adam} to train the framework. The learning rate ($\eta$) is calculated as \cite{loshchilov2016sgdr}

\begin{equation}\label{eq:learning_rate}
\eta_{c}=\eta_{\min }+\frac{1}{2}\left(\eta_{\max }-\eta_{\min }\right)\left(1+\cos \left(\frac{\mathrm{step_c}}{\mathrm{step_f}} \pi\right)\right)
\end{equation}

\noindent
where $\eta_{c}$ is a learning rate at step $\mathrm{step_c}$, $\eta_{\min }$ is the minimum learning rate, which is set as $1 \times 10^{-16}$, $\eta_{\max }$ is the initial learning rate, which is selected as $1 \times 10^{-5}$, $\mathrm{step_c}$ is the current step, and $\mathrm{step_f}$ is the final step. To prevent our networks from overfitting behavior, we follow early stopping and generalized cross-validation criteria \cite{prechelt1998early,prechelt1998automatic}. Note that instead of literally stopping our training cycle, we only save the set of trained weights and biases from being used in the online phase when the current validation loss is lower than the lowest validation from all the previous training cycles.  \par

\begin{algorithm}[!ht]
\caption{Training autoencoder (AE) with Barlow Twins (BT) self-supervised learning (BT-ROM)}\label{al:ae-bt}
\# Training data $\bm{X}_h$ and distorted data $\bm{X}_{h, A}$, $\bm{X}_{h, B}$ are input of {encoder}  \newline
\# latent spaces $\bm{z}^{\bm{X}}$, $\bm{z}_A^{\bm{X}}$, and $\bm{z}_B^{\bm{X}}$ are output {encoder}  \newline
\# latent space $\bm{z}^{\bm{X}}$ is output of {decoder}  \newline
\# Approximation of $\bm{X}_h$, i.e., $\widehat{\bm{X}_h}$ is output of {decoder}  \newline
\# latent spaces $\bm{z}_A^{\bm{X}}$ and $\bm{z}_B^{\bm{X}}$ are input of
{projector} \newline
\# cross-correlation matrix $\mathbf{C}^{\bm{X}}$ is output of {projector} \newline
\begin{algorithmic}[1]
\State Initialize (or load pre-trained models) {encoder}, {decoder}, and {projector} \Comment{size of latent space $\mathrm{Q}$ has to be specified.}
\State Initialize (or load pre-trained optimizers) three optimizers for each of {encoder}, {decoder}, and {projector}
\State Load training set $\bm{\mu}$ \Comment{the total training data is $\mathrm{M} N^t$}
\State Randomly select 5\% of $\mathrm{M} N^t$ for validation set $\bm{\mu}_{\mathrm{validation}}$ \Comment{the total training data becomes 95\% of $\mathrm{M} N^t$}
\State Add random noise \Comment{see \eqref{eq:add_noise}}
\State Add Gaussian blur \Comment{see \eqref{eq:gb}}
\State From step 5 and 6, we obtain $\bm{X}_{h,A}\left(t, \bm{\mu}\right)$ and $\bm{X}_{h,B}\left(t, \bm{\mu}\right)$ from $\bm{X}\left(t, \bm{\mu}\right)$
\ForEach {epoch}
\State \emph{Outer loop: training BT} \Comment{Batch size $\mathbf{B}_{\mathrm{outer}}$}
\ForEach {$\mathbf{B}_{\mathrm{outer}}$}
\State $\bm{z}^{\bm{X}}_{A}\left(t, \bm{\mu}\right) = \mathrm{encoder}\left(  {\bm{X}}_{h,A}\left(t, \bm{\mu}\right)  \right)$
\State $\bm{z}^{\bm{X}}_{B}\left(t, \bm{\mu}\right) = \mathrm{encoder}\left(  {\bm{X}}_{h,B}\left(t, \bm{\mu}\right)  \right)$
\State $\mathbf{C}^{\bm{X}}\left(t, \bm{\mu}\right) = \mathrm{projector} \left( \bm{z}^{\bm{X}}_{A}\left(t, \bm{\mu}\right), \bm{z}^{\bm{X}}_{B}\left(t, \bm{\mu}\right)  \right)$ 
\State Calculate BT loss $\mathcal{L}_{\mathrm{BT}}^{\bm{X}}$ \Comment{see \eqref{eq:loss_bt}}
\State Back-propagation of BT loss w.r.t. each $\mathrm{encoder}\left( {\mathbf{W}}, {\mathbf{b}}\right)$ and $\mathrm{projector}\left( {\mathbf{W}}, {\mathbf{b}}\right)$
\State Update $\mathrm{encoder}\left( {\mathbf{W}}, {\mathbf{b}}\right)$ and $\mathrm{projector}\left( {\mathbf{W}}, {\mathbf{b}}\right)$ using BT optimizer
\State Update learning rate $\eta_{c}$ of BT optimizer \Comment{see \eqref{eq:learning_rate}}
\State \emph{Inner loop: training AE} \Comment{Batch size $\mathbf{B}_{\mathrm{inner}}$}
\ForEach {$\mathbf{B}_{\mathrm{inner}}$}
\State $\bm{z}^{\bm{X}}\left(t, \bm{\mu}\right) = \mathrm{encoder}\left(  {\bm{X}}_h\left(t, \bm{\mu}\right)  \right)$
\State $\widehat{{\bm{X}}_h}\left(t, \bm{\mu}\right) = \mathrm{decoder}\left(  \bm{z}^{\bm{X}}\left(t, \bm{\mu}\right)  \right)$
\State Calculate AE loss ${\mathcal{L}_\mathrm{AE}^{\bm{X}}}$ (data compression loss)  \Comment{see \eqref{eq:loss_ae}}
\State Back-propagation of AE loss w.r.t. each $\mathrm{encoder}\left( {\mathbf{W}}, {\mathbf{b}}\right)$ and $\mathrm{decoder}\left( {\mathbf{W}}, {\mathbf{b}}\right)$
\State Update $\mathrm{encoder}\left( {\mathbf{W}}, {\mathbf{b}}\right)$ and $\mathrm{decoder}\left( {\mathbf{W}}, {\mathbf{b}}\right)$ using AE optimizer
\State Update learning rate $\eta_{c}$ of AE optimizer \Comment{see \eqref{eq:learning_rate}}
\EndFor
\EndFor
\EndFor
\end{algorithmic}
Reflecting the third step in Supplementary fig. \ref{fig:bbt_explain}.
\end{algorithm}

We reiterate the autoencoder's architecture used for BT-ROM in the Supplementary tab. \ref{tab:dcov}; please refer to \cite{kadeethum2022reduced, kadeethum2022fomassistrom} for further details on architecture search and hyper-parameters tuning. To sum up, the BT-ROM model is composed of one encoder, one decoder, and one projector with their own sets of weight matrices ($\mathbf{W}$) and biases (${\mathbf{b}}$). Each linear layer is in the Supplementary tab. \ref{tab:dcov} is subjected to the LeakyReLU activation function with a negative slope of 0.2. No batch normalization or dropout layers are used in this architecture, making it preferable and lightweight. \par

\begin{table}[!ht]
\centering
\caption{Autoencoder - reflecting the third step in Supplementary fig. \ref{fig:bbt_explain}. (input and output sizes are represented by {[}$\mathrm{B}$, $\mathrm{DOF}${]}. $\mathrm{B}$ is a batch size, and $\bm{z}$ is nonlinear manifolds.)}
\begin{tabular}{|l|c|c|}
\hline
\textbf{block}                  & \multicolumn{1}{l|}{\textbf{input size}} & \multicolumn{1}{l|}{\textbf{output size}} \\ \hline
$1^{\mathrm{st}}$ linear layer   & {[}$\mathrm{B}$, $\mathrm{DOF}${]}           & {[}$\mathrm{B}$, $\mathrm{int(DOF}/2)${]}                \\ \hline
$2^{\mathrm{nd}}$ linear layer   & {[}$\mathrm{B}$, $\mathrm{int(DOF}/2)${]}             & {[}$\mathrm{B}$, $\mathrm{int(DOF}/4)${]}                                    \\ \hline
$3^{\mathrm{rd}}$ linear layer   & {[}$\mathrm{B}$, $\mathrm{int(DOF}/4)${]}            & {[}$\mathrm{B}$, $\mathrm{int(DOF}/8)${]}                                  \\ \hline
$4^{\mathrm{th}}$ linear layer   & {[}$\mathrm{B}$, $\mathrm{int(DOF}/8)${]}            & {[}$\mathrm{B}$, $\mathrm{int(DOF}/16)${]}                                   \\ \hline
$5^{\mathrm{th}}$ linear layer   & {[}$\mathrm{B}$, $\mathrm{int(DOF}/16)${]}              & {[}$\mathrm{B}$, $\mathrm{int(DOF}/32)${]}                                       \\ \hline
$1^{\mathrm{st}}$ \textbf{bottleneck}     & {reshape([}$\mathrm{B}$, $\mathrm{int(DOF}/32)${])}         & {[}$\mathrm{B}$, $\bm{z}$ {]}                                       \\ \hline
$2^{\mathrm{nd}}$ \textbf{bottleneck}     & {[}$\mathrm{B}$, $\bm{z}$ {]}         & {reshape([}$\mathrm{B}$, $\mathrm{int(DOF}/32)${])}                                   \\ \hline
$6^{\mathrm{th}}$ linear layer     & {[}$\mathrm{B}$, $\mathrm{int(DOF}/32)${]}       & {[}$\mathrm{B}$, $\mathrm{int(DOF}/16)${]}                             \\ \hline
$7^{\mathrm{th}}$ linear layer     & {[}$\mathrm{B}$, $\mathrm{int(DOF}/16)${]}         & {[}$\mathrm{B}$, $\mathrm{int(DOF}/8)${]}                              \\ \hline
$8^{\mathrm{th}}$ linear layer    & {[}$\mathrm{B}$, $\mathrm{int(DOF}/8)${]}         & {[}$\mathrm{B}$, $\mathrm{int(DOF}/4)${]}                                        \\ \hline
$9^{\mathrm{th}}$ linear layer     & {[}$\mathrm{B}$, $\mathrm{int(DOF}/4)${]}          & {[}$\mathrm{B}$, $\mathrm{int(DOF}/2)${]}                                        \\ \hline
$10^{\mathrm{th}}$ linear layer     & {[}$\mathrm{B}$, $\mathrm{int(DOF}/2)${]}      & {[}$\mathrm{B}$, $\mathrm{DOF}${]}                                     \\ \hline
\end{tabular}
\label{tab:dcov}
\end{table}
   
For the next step, we map $t$ and $\bm{\mu}$ to its representation in the linear and nonlinear manifold $\bm{z}^{\bm{X}}\left(t, \bm{\mu}\right)$. Note that if we deal with steady-state problems, we simply drop the $t$ term. We follow a procedure proposed by \cite{kadeethum2022reduced} and use linear radial basis function (RBF) interpolation \cite{wright2003radial} to map $t$ and $\bm{\mu}$ to $\bm{z}^{\bm{X}}\left(t, \bm{\mu}\right)$. During the online phase, we utilize the \emph{trained} RBF and the \emph{trained} decoder to approximate $\widehat{{\bm{X}}}_{h}\left(\cdot; t, \bm{\mu}\right)$ for each inquiry (i.e., a value of $\bm{\mu}$) through

\begin{equation}
\widehat{\bm{z}}^{\bm{X}}\left(t, \bm{\mu}\right) = \operatorname{RBF} \left( t, \bm{\mu} \right),
\label{eq:online_solution_z_nonlinear}
\end{equation}

\noindent
and, subsequently, 

\begin{equation}
\widehat{{\bm{X}}}_{h}\left( t, \bm{\mu}\right) = \operatorname{decoder} \left(\widehat{\bm{z}}^{\bm{X}}\left( t, \bm{\mu}\right) \right).
\label{eq:online_solution_nonlinear}
\end{equation}

\section{Data generation}\label{sec:data_generation}

Here, we briefly explain how we create our training, validation, and testing sets. We divide our investigation into two phases. In the first phase, we focus on different physics but similar topology (Sec. Similar topology in the main text). Here, we illustrate the effects of the number of training samples and the number of parents on the BT-ROM's and p-BT-ROM's performance. In the second phase, we use different physics and topologies (Sec. Different topologies in the main text). For this phase, we only focus on the effect of the number of parents on the BT-ROM's and p-BT-ROM's performance.  We have described each topology in detail in \ref{sec:geo}. \par

In terms of physics problems utilized throughout this study, we have four main problems. We have described each problem in detail in \ref{sec:physics_problems}. The first one (Problem \#1) is a transport problem with given different velocity fields, i.e., $\bm{q}$ is the superficial velocity vector defined by $\bm{q} = (20.0, \mu_1 \cos (\pi x))$, and our parameter $\bm{\mu} = (\mu_1 = [5.0, 25.0])$ \ref{sec:physics_problems_tv}). For this problem, we use a mesh provided in Supplementary fig. \ref{fig:sum_geo}a, and a sample of this problem is shown in Supplementary fig. \ref{fig:same_topo_samples}a. The second problem (Problem \#2) is a transport problem with given different diffusivity coefficients, i.e., $\bm{\mu} = (\bm{D} = (\mu_1, 0.0, 0.0, \mu_1))$ and $\mu_1= [0.1, 1.0]$ (\ref{sec:physics_problems_td}). For this problem, we use two meshes presented in Figs. \ref{fig:sum_geo}a, b, and their samples are illustrated in  Supplementary fig. \ref{fig:same_topo_samples}b and Supplementary fig. \ref{fig:diff_topo_samples}a, respectively. \par

The third problem (Problem \#3) is a gravity-driven flow in porous media with the Rayleigh number ($\mathrm{Ra}$) as a parameter $\bm{\mu} = (\mu_1 = \mathrm{Ra})$, and $\mathrm{Ra}= [350.0, 450.0]$ (\ref{sec:physics_problems_el}).  For this problem, we use two meshes presented in Figs. \ref{fig:sum_geo}a, c, and their samples are illustrated in  Supplementary fig. \ref{fig:same_topo_samples}c and Supplementary fig. \ref{fig:diff_topo_samples}b, respectively. The fourth problem (Problem \#4) is the finite deformation of a hyperelastic material (\ref{sec:physics_problems_hy}). For this problem, $\bm{\mu}$ is different for the problem in a 2-Dimensional domain, mesh: Supplementary fig. \ref{fig:sum_geo}a and sample: Supplementary fig. \ref{fig:same_topo_samples}d, and a 3-Dimensional domain, mesh: Supplementary fig. \ref{fig:sum_geo}d and sample: Supplementary fig. \ref{fig:diff_topo_samples}c. For the 2-Dimensional setting, we enforce $\mathbf{u} = (\mu_1, \mu_2)$ at the boundary condition on the face of where $y = 1.0$, top surface, and $\bm{\mu} = (0.05\mu_1, 0.05\mu_2)$, and $\mu_1= [-1.0, 1.0]$ and $\mu_2 = [-1.0, 1.0]$. For the 3-Dimensional setting, we have traction forces $\mathbf{T} = ({\mu_1},  0.0, 0.0)$, and we enforce

\begin{equation}
\begin{aligned}
\mathbf{u}=&(0.0,\\
&\mu_2(0.5+(y-0.5) \cos (\pi / 3)-(z-0.5) \sin (\pi / 3)-y) / 2, \\
&\mu_2(0.5+(y-0.5) \sin (\pi / 3)+(z-0.5) \cos (\pi / 3)-x)) / 2)
\end{aligned}
\end{equation}

\noindent
at the boundary condition on the face of where $x = 1.0$. $\bm{\mu} = (\mu_1, \mu_2)$, and $\mu_1= [0.1, 0.9]$ and $\mu_2 = [0.1, 0.9]$. \par

\begin{figure}[!ht]
  \centering
  \includegraphics[width=12.0cm,keepaspectratio]{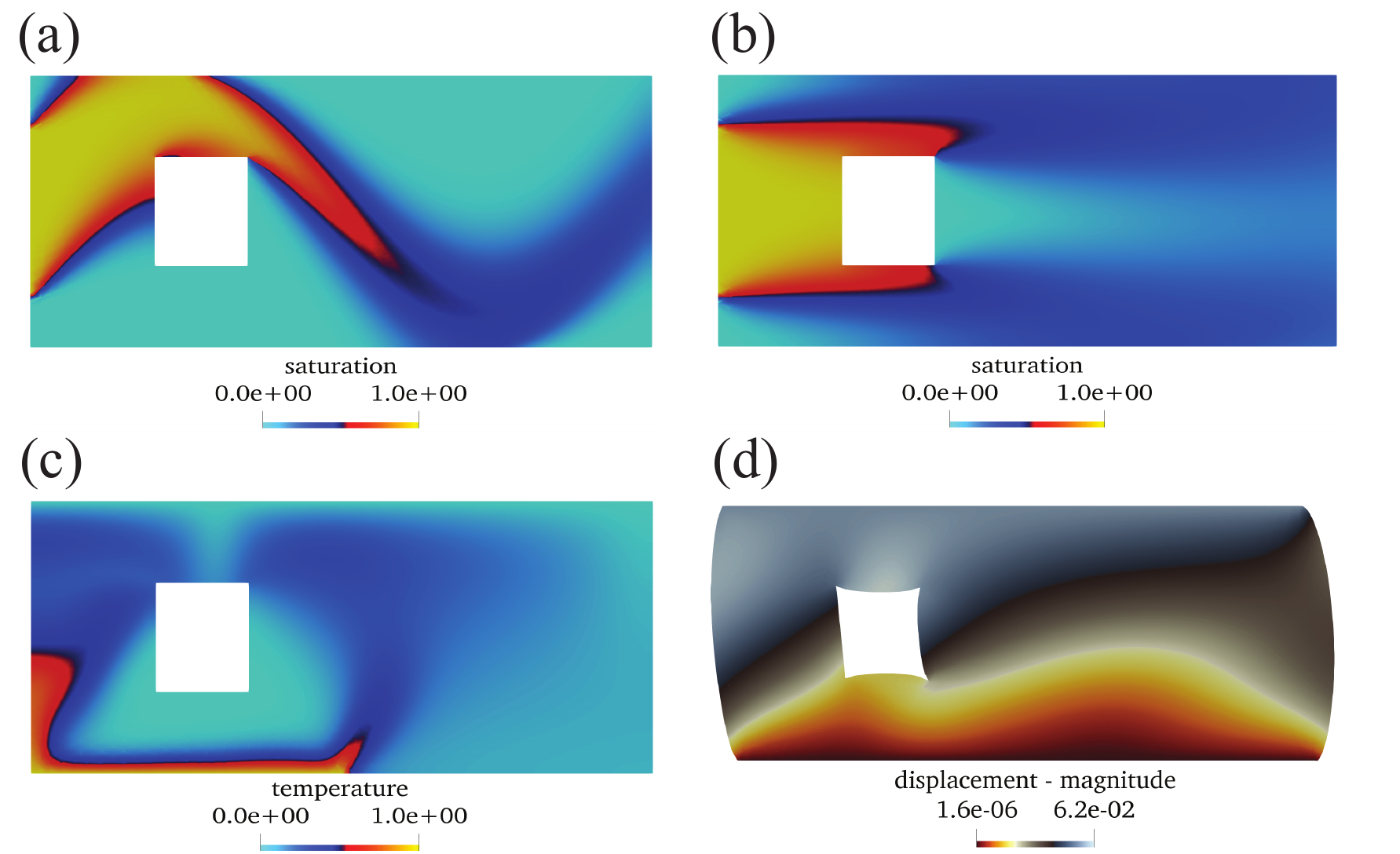}
  \caption{Samples of physics problems simulated using mesh provided in Supplementary fig. \ref{fig:sum_geo}a (similar topology): (a) Problem \#1 ( \ref{sec:physics_problems_tv}), (b) Problem \#2 (\ref{sec:physics_problems_td}), (c) Problem \#3 (\ref{sec:physics_problems_el}), and (d) Problem \#4 (\ref{sec:physics_problems_hy}). We note that problems \#1 to \#3 are time-dependent problems while Problem \#4 is at a steady-state. For (a) - (c), we show the last time-step.}
  \label{fig:same_topo_samples}
\end{figure}

\begin{figure}[!ht]
  \centering
  \includegraphics[width=12.0cm,keepaspectratio]{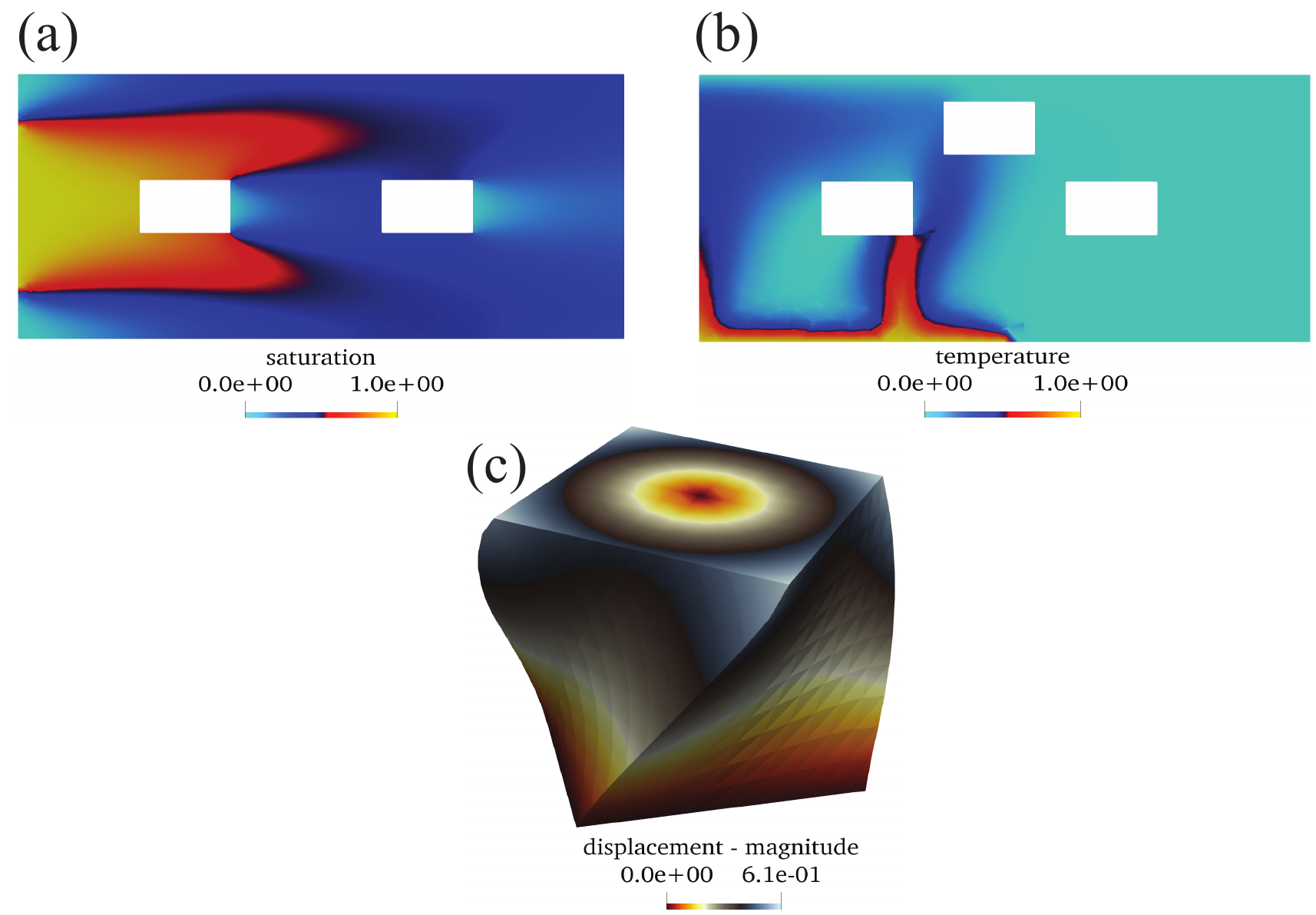}
  \caption{Samples of physics problems simulated using different meshes provided in Supplementary fig. \ref{fig:sum_geo}b-d (different topologies): (a) Problem \#2 (\ref{sec:physics_problems_td}) with mesh shown in Supplementary fig. \ref{fig:sum_geo}b, (b) Problem \#3 (\ref{sec:physics_problems_el}) with mesh shown in Supplementary fig. \ref{fig:sum_geo}c, and (c) Problem \#3 (\ref{sec:physics_problems_hy}) with mesh shown in Supplementary fig. \ref{fig:sum_geo}d. We note that problems \#2 and \#3 are time-dependent problems while Problem \#4 is at a steady-state. For (a) - (b), we show the last time-step.}
  \label{fig:diff_topo_samples}
\end{figure}

We generate training, validation, and test sets to train and evaluate our model. For the training set, we uniformly sample our parameter space $\bm{\mu}$ with $\mathrm{M}$ snapshots (i.e., assuming $\bm{\mu} = [\mu_1]$, $\mu_1 = (0.0, 1.0)$, and $\mathrm{M} = 3$, our $\mu_1 = [0.0, 0.5, 1.0]$). For Problems \#1 to \#3, our actual training set would be $\mathrm{M}N_t$ (i.e., $\mathrm{M}$ snapshots $\times$ number of time-steps for each $\mathrm{M}$). We design our framework this way to handle data provided from FOM with adaptive time-stepping, which is essential for an advection-dominated problem (i.e., to satisfy Courant–Friedrichs–Lewy (CFL) condition, Problems \#1 to \#3). The framework can also deliver the solution at any time, including times that do not exist in the training set. To elaborate, this does not imply that our model can extrapolate, but it can deliver any timestamps inside the range of the training set (see \cite{kadeethum2022reduced, kadeethum2021nonTH} for more details on this). For Problem \#4, the actual training set is $\mathrm{M}$ because it is a steady-state problem. We then randomly select 5\% of the training samples to be used as a validation set. For Problems \#1 to \#3, the final total training samples are $0.95\mathrm{M} N^t$ while, for Problem \#4, the final total training samples are $0.95\mathrm{M}$ For the testing set, we randomly sample our parameter space $\bm{\mu}$ with $\mathrm{M_{test}}$ snapshots.

\section{Complementary information on numerical experiments} \label{sec:si_results}

We present supplement information on numerical experiments throughout the following sections. These results are used to support numerical examples in the main text, Sec. Results. Throughout this section, we use The mean squared error (MSE)

\begin{equation} \label{eq:mse}
{\mathrm{MSE}} = 
    \frac{1}{\mathrm{M} N^t} \sum_{i=1}^{\mathrm{M}}\sum_{j=0}^{N^t}\left|\widehat{\bm{X}}_h\left(t^j, \bm{\mu}^{(i)}\right)-\bm{X}_h\left(t^j, \bm{\mu}^{(i)}\right)\right|^{2},
\end{equation}

\noindent
and mean absolute error (MAE) 

\begin{equation}\label{eq:mae}
\mathrm{MAE}=  \frac{1}{\mathrm{M} N^t}  \sum_{i=1}^{\mathrm{M}}\sum_{j=0}^{N^t}|\widehat{\bm{X}}_h\left(t^j, \bm{\mu}^{(i)}\right)-\bm{X}_h\left(t^j, \bm{\mu}^{(i)}\right)|.
\end{equation}

\noindent
as our evaluation matrices.

\subsection{Using trained models for an initialization} \label{sec:init_results}

First, we use trained models' weights and biases to initialize a new model. If we use more than one trained model, we simply take an arithmetic average of each layer for the new model. For illustration purposes, throughout this section, we use only one trained model - Problem \#1 (\ref{sec:physics_problems_tv}) - for the initialization of the new model. Our results are presented in Supplementary fig. \ref{fig:init_ae} for the reconstruction loss of the validation set (\eqref{eq:loss_ae}) and Supplementary fig. \ref{fig:init_bt} for the Barlow Twins loss (\eqref{eq:loss_bt}) of the validation set. To emphasize this process, we pick the weights and biases of the trained model that delivers the least validation loss, shown in a red circle in Figs. \ref{fig:init_ae}a and \ref{fig:init_bt}a, for the initialization. In short, from these figures, one can observe that using the trained model as a starting point, the BT-ROM delivers lower reconstruction validations loss for Problem \#2 to \#4 ( \ref{sec:physics_problems_tv} to \ref{sec:physics_problems_hy}). However, this trend is not always true for the Barlow Twins loss, as using the trained model as a starting point can cause an adverse effect, especially for Problems \#3 and \#4. \par

\begin{figure}[!ht]
  \centering
  \includegraphics[width=14.0cm,keepaspectratio]{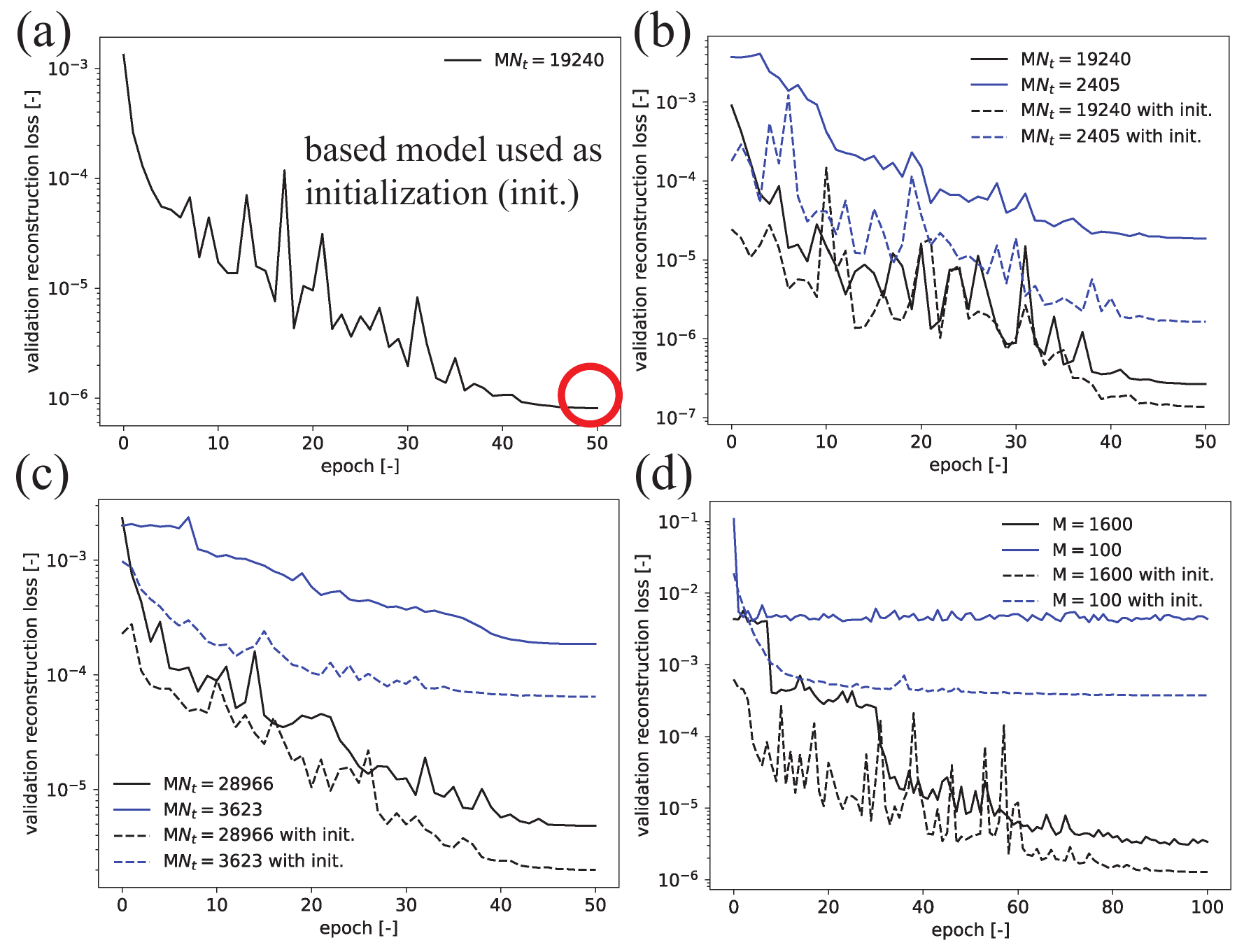}
  \caption{Using a trained model for an initialization: validation reconstruction loss (\eqref{eq:loss_ae}) results of (a) physical Problem \#1 (\ref{sec:physics_problems_tv}) used as an initialization for the following physical problems, (b) physical Problem \#2 (\ref{sec:physics_problems_td}) for both large dataset ($\mathrm{M}N_t = 19240$) and small dataset ($\mathrm{M}N_t = 2405$), (c) physical Problem \#3 (\ref{sec:physics_problems_el}) for both large dataset ($\mathrm{M}N_t = 28966$) and small dataset ($\mathrm{M}N_t = 3623$), and (d) physical Problem \#4 (\ref{sec:physics_problems_hy}) for both large dataset ($\mathrm{M} = 1600$) and small dataset ($\mathrm{M} = 100$). We note that problems \#1 to \#3 are time-dependent problems while Problem \#4 is at a steady-state. The red circle in (a) represents the epoch that has the lowest validation loss; and subsequently, be used as an initialization for the other problems.}
  \label{fig:init_ae}
\end{figure}

\begin{figure}[!ht]
  \centering
  \includegraphics[width=14.0cm,keepaspectratio]{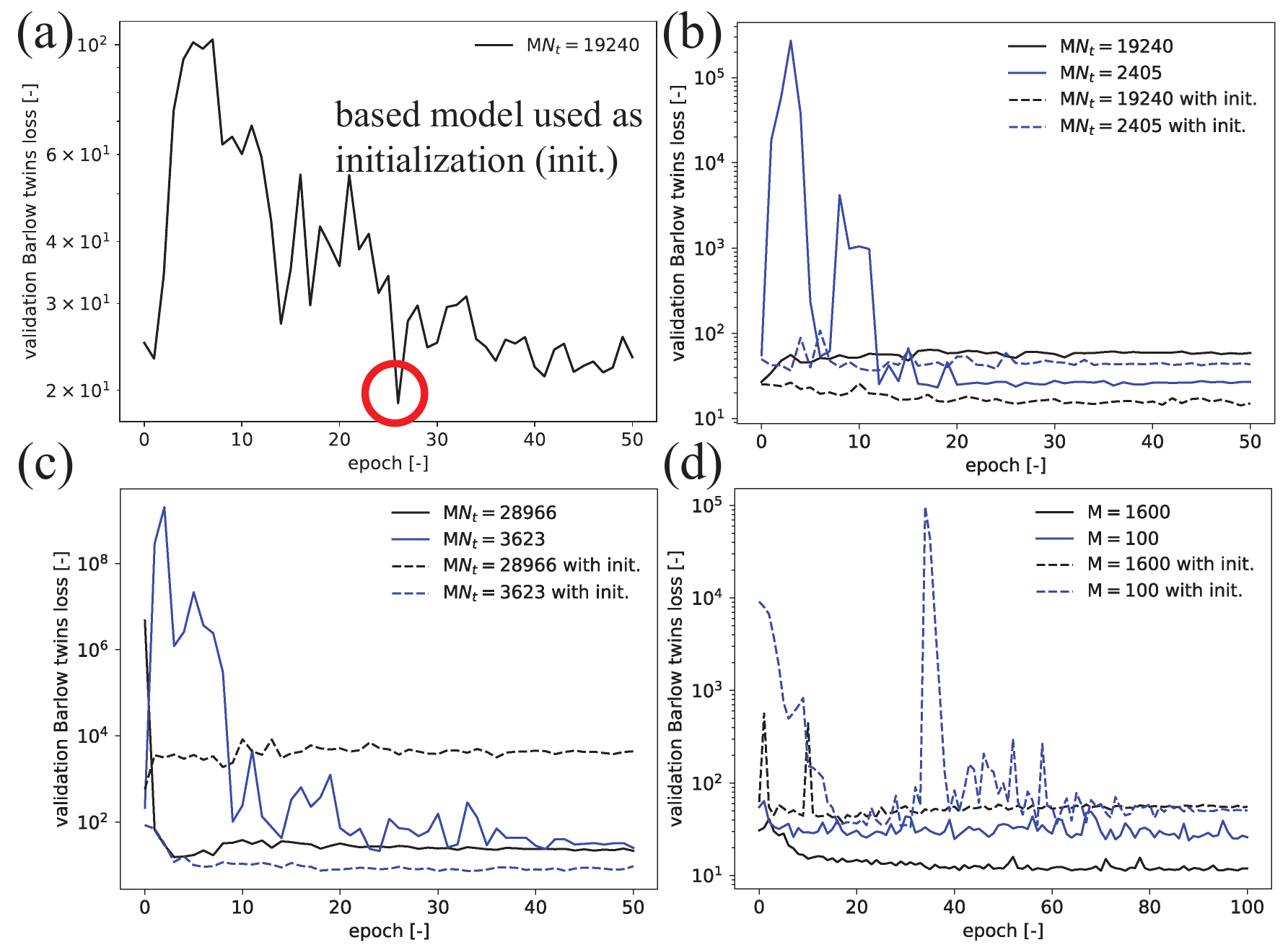}
  \caption{Using a trained model for an initialization: validation Barlow Twins loss (\eqref{eq:loss_bt}) results of (a) physical Problem \#1 (\ref{sec:physics_problems_tv}) used as an initialization for the following physical problems, (b) physical Problem \#2 (\ref{sec:physics_problems_td}) for both large dataset ($\mathrm{M}N_t = 19240$) and small dataset ($\mathrm{M}N_t = 2405$), (c) physical Problem \#3 (\ref{sec:physics_problems_el}) for both large dataset ($\mathrm{M}N_t = 28966$) and small dataset ($\mathrm{M}N_t = 3623$), and (d) physical Problem \#4 - 2-Dimensional domain (\ref{sec:physics_problems_hy}) for both large dataset ($\mathrm{M} = 1600$) and small dataset ($\mathrm{M} = 100$). We note that problems \#1 to \#3 are time-dependent problems while Problem \#4 is at a steady-state. The red circle in (a) represents the epoch that has the lowest validation loss; and subsequently, be used as an initialization for the other problems.}
  \label{fig:init_bt}
\end{figure}

\subsection{Progressive Barlow Twins reduced order modeling without initialization} \label{sec:pbt_results}

Next, we investigate the effect of using p-BT-ROM (see Sec. Methodology in the main text and \ref{sec:si_btrom}) on the validation losses: the reconstruction loss of the validation set (\eqref{eq:loss_ae}) and the Barlow Twins loss (\eqref{eq:loss_bt}) of the validation set. We emphasize that here we do not use the trained models' (or parents') weights and biases to initialize our p-BT-ROM. The results of reconstruction (Supplementary fig. \ref{fig:pnn_ae}) and Barlow Twins (Supplementary fig. \ref{fig:pnn_bt})losses are in line; so, we will only discuss the reconstruction loss results. The schematic of p-BT-ROM specifies child$\sim$parent(s) relationship is shown in Supplementary fig. \ref{fig:pnn_ae}a. To elaborate, for Supplementary fig. \ref{fig:pnn_ae}b, we test our model using Problem \#2 (\ref{sec:physics_problems_td}), and our p-BT-ROM has only one parent, Problem \#1. One can see that p-BT-ROM delivers more accurate results. \par

For Supplementary fig. \ref{fig:pnn_ae}c, we test our model using Problem \#3 (\ref{sec:physics_problems_el}), and the p-BT-ROM has two parents, Problem \#1 and Problem \#2. Again, the validation loss of p-BT-ROM is much lower than that of BT-ROM (no parent). Lastly, we use Problem \#4 (\ref{sec:physics_problems_hy}) to test our model, and the results align with previous observations. Hence, from these experiments, we observe that using p-BT-ROM can improve the model accuracy substantially. \par

\begin{figure}[!ht]
  \centering
  \includegraphics[width=14.0cm,keepaspectratio]{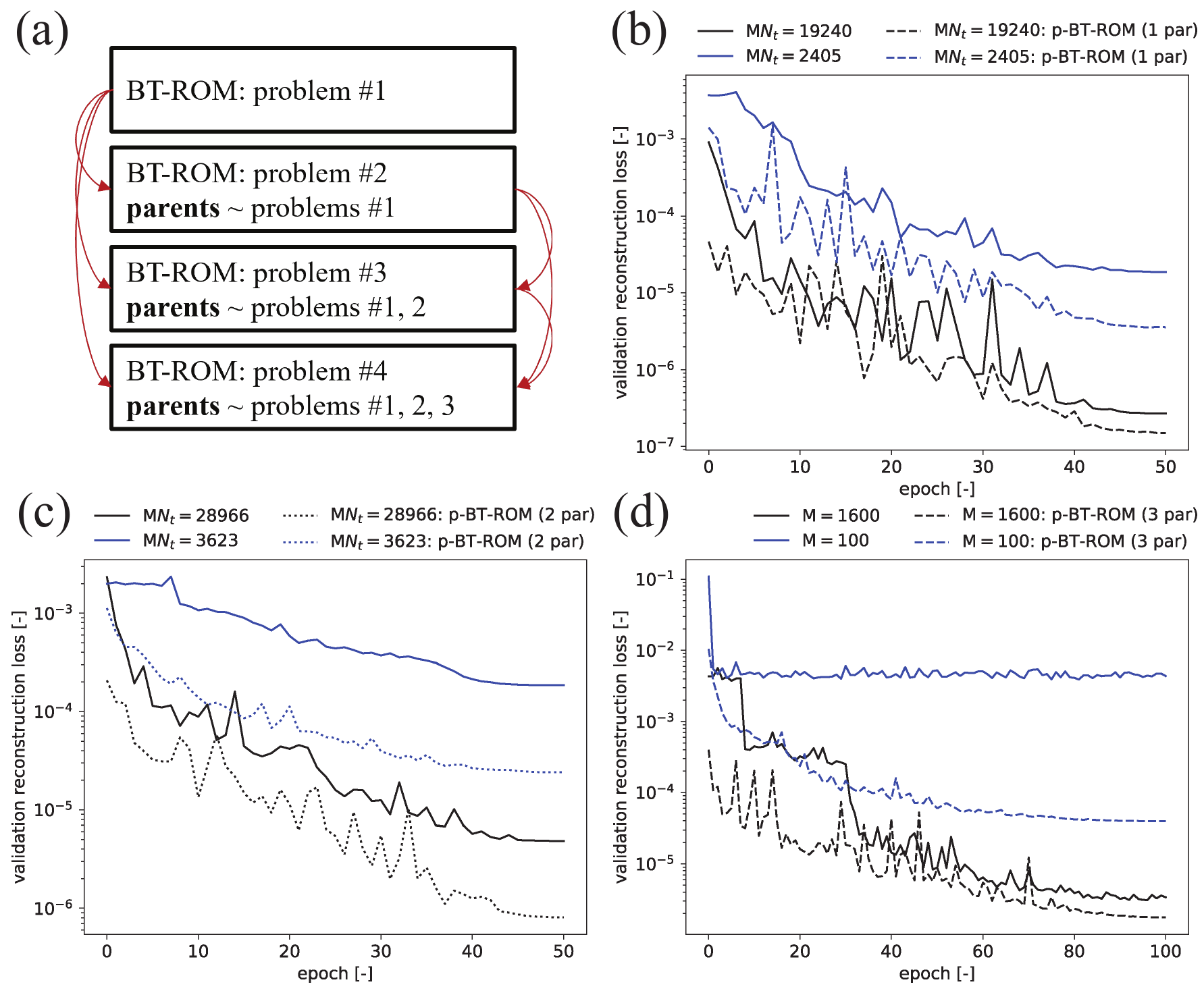}
  \caption{p-BT-ROM without initialization: (a) schematic of p-BT-ROM specifies child$\sim$parent(s) relationship, validation reconstruction loss (\eqref{eq:loss_ae}) results of (b) physical Problem \#2 (\ref{sec:physics_problems_td}) for both large dataset ($\mathrm{M}N_t = 19240$) and small dataset ($\mathrm{M}N_t = 2405$), (c) physical Problem \#3 (\ref{sec:physics_problems_el}) for both large dataset ($\mathrm{M}N_t = 28966$) and small dataset ($\mathrm{M}N_t = 3623$), and (d) physical Problem \#4 - 2-Dimensional domain (\ref{sec:physics_problems_hy}) for both large dataset ($\mathrm{M} = 1600$) and small dataset ($\mathrm{M} = 100$). We note that problems \#1 to \#3 are time-dependent problems while Problem \#4 is at a steady-state.}
  \label{fig:pnn_ae}
\end{figure}

\begin{figure}[!ht]
  \centering
  \includegraphics[width=14.0cm,keepaspectratio]{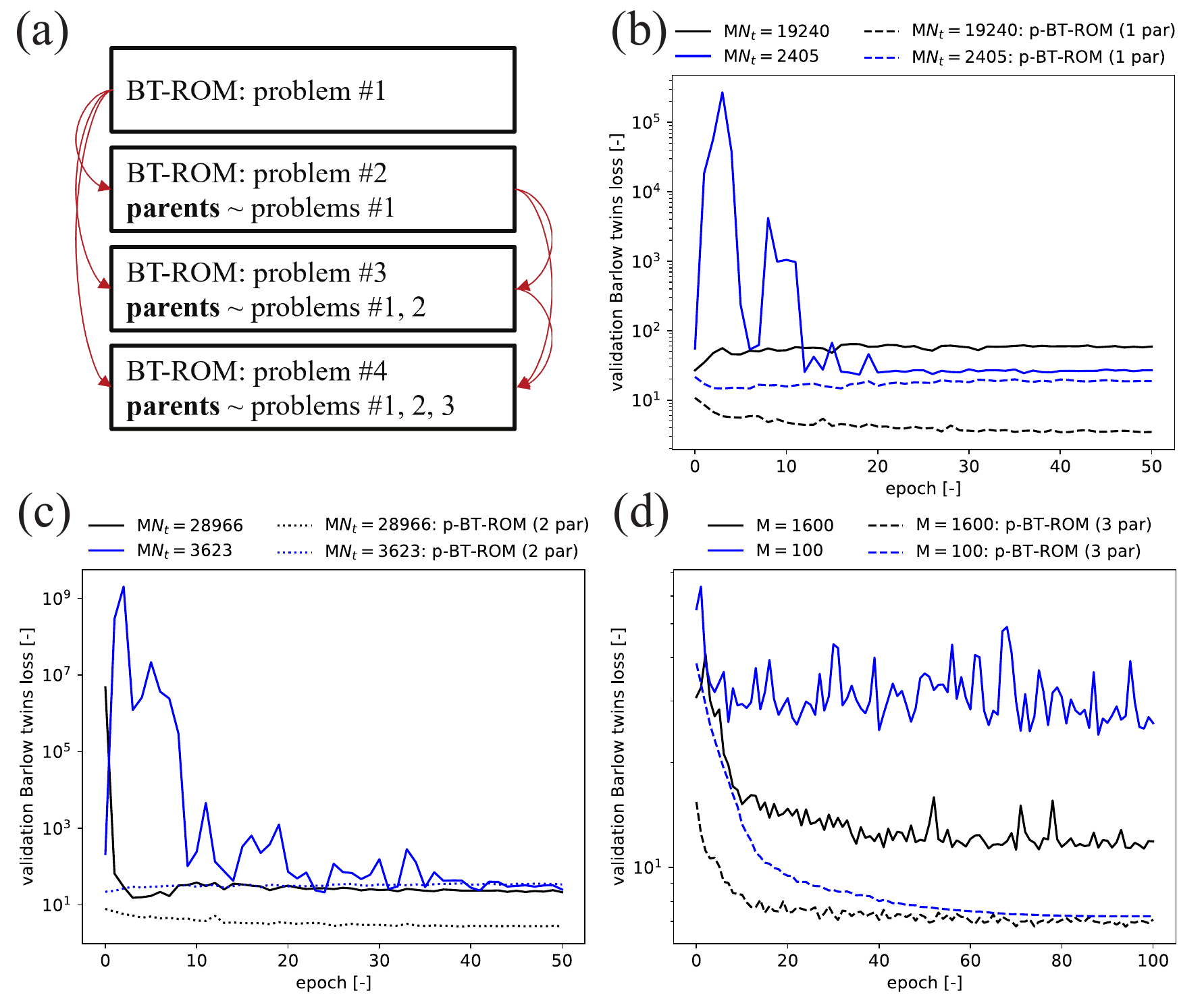}
  \caption{p-BT-ROM without initialization: (a) schematic of p-BT-ROM specifies child$\sim$parent(s) relationship, validation Barlow Twins loss (\eqref{eq:loss_bt}) results of (b) physical Problem \#2 (\ref{sec:physics_problems_td}) for both large dataset ($\mathrm{M}N_t = 19240$) and small dataset ($\mathrm{M}N_t = 2405$), (c) physical Problem \#3 (\ref{sec:physics_problems_el}) for both large dataset ($\mathrm{M}N_t = 28966$) and small dataset ($\mathrm{M}N_t = 3623$), and (d) physical Problem \#4 - 2-Dimensional domain (\ref{sec:physics_problems_hy}) for both large dataset ($\mathrm{M} = 1600$) and small dataset ($\mathrm{M} = 100$). We note that problems \#1 to \#3 are time-dependent problems while Problem \#4 is at a steady-state.}
  \label{fig:pnn_bt}
\end{figure}

\subsection{Progressive Barlow Twins reduced order modeling with initialization} \label{sec:pbt_init_results}

As previously illustrated that using a trained model as an initialization (\ref{sec:init_results}) as well as a progressive reduced order model (\ref{sec:pbt_results}) generally increases our BT-ROM accuracy. Throughout this section, we investigate the p-BT-ROM with initialization (p-BT-ROM with init.) effects on the model's accuracy (i.e., investigate the effect of using p-BT-ROM (see Sec. Methodology in the main text and \ref{sec:si_btrom}) on the validation losses: the reconstruction loss (\eqref{eq:loss_ae}) of the validation set and the Barlow Twins loss (\eqref{eq:loss_bt}) of the validation set). To reiterate, if we use more than one trained model (i.e., p-BT-ROM has more than one parent model), we take an arithmetic average of each layer for all parents for the new model. \par

We present the reconstruction loss (\eqref{eq:loss_ae}) of the validation set and the Barlow Twins loss (\eqref{eq:loss_bt}) of the validation set in Figs. \ref{fig:pnn_ae_init} and \ref{fig:pnn_bt_init}, respectively. The schematic of p-BT-ROM specifies child$\sim$parent(s) relationship is shown in Supplementary fig. \ref{fig:pnn_ae_init}a. For Figs. \ref{fig:pnn_ae_init}b and \ref{fig:pnn_bt_init}b, we focus on Problem \#2 (\ref{sec:physics_problems_td}) by using Problem \#1  (\ref{sec:physics_problems_tv}) as a parent. We observe that for the reconstruction loss using p-BT-ROM and p-BT-ROM with init. are quite similar; however, p-BT-ROM with init. delivers the best Barlow Twins loss result. \par

For Supplementary fig. \ref{fig:pnn_ae_init}c and \ref{fig:pnn_bt_init}c, we focus on Problem \#3 (\ref{sec:physics_problems_el}) with Problem \#1 and \#2 as its parents. Here, the reconstruction and Barlow Twins loss using p-BT-ROM and p-BT-ROM with init. are not much different and are best. Lastly, For Supplementary fig. \ref{fig:pnn_ae_init}d and \ref{fig:pnn_bt_init}d, we focus on Problem \#4 (\ref{sec:physics_problems_hy}) with Problem \#1, \#2, and \#3 as its parents. For this problem, the p-BT-ROM with init. delivers the best result for the reconstruction loss, while the p-BT-ROM delivers the best result for the Barlow Twins loss. \par

\begin{figure}[!ht]
  \centering
  \includegraphics[width=14.0cm,keepaspectratio]{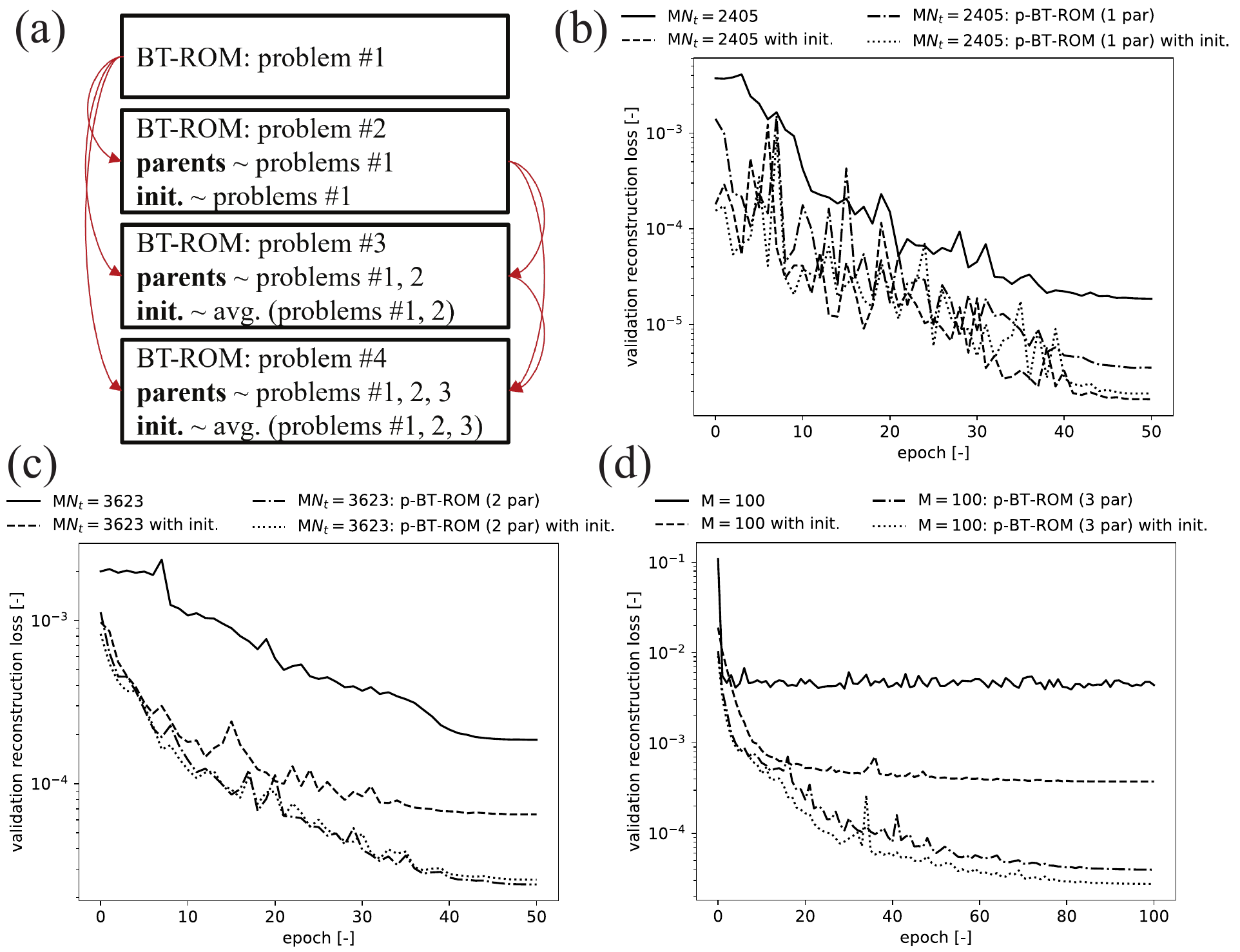}
  \caption{p-BT-ROM with initialization: (a) schematic of p-BT-ROM specifies child$\sim$parent(s) relationship and how to initialize, validation reconstruction loss (\eqref{eq:loss_ae}) results of (b) physical Problem \#2 (\ref{sec:physics_problems_td}) for both large dataset ($\mathrm{M}N_t = 19240$) and small dataset ($\mathrm{M}N_t = 2405$), (c) physical Problem \#3 (\ref{sec:physics_problems_el}) for both large dataset ($\mathrm{M}N_t = 28966$) and small dataset ($\mathrm{M}N_t = 3623$), and (d) physical Problem \#4 - 2-Dimensional domain (\ref{sec:physics_problems_hy}) for both large dataset ($\mathrm{M} = 1600$) and small dataset ($\mathrm{M} = 100$). We note that problems \#1 to \#3 are time-dependent problems while Problem \#4 is at a steady-state.}
  \label{fig:pnn_ae_init}
\end{figure}

\begin{figure}[!ht]
  \centering
  \includegraphics[width=14.0cm,keepaspectratio]{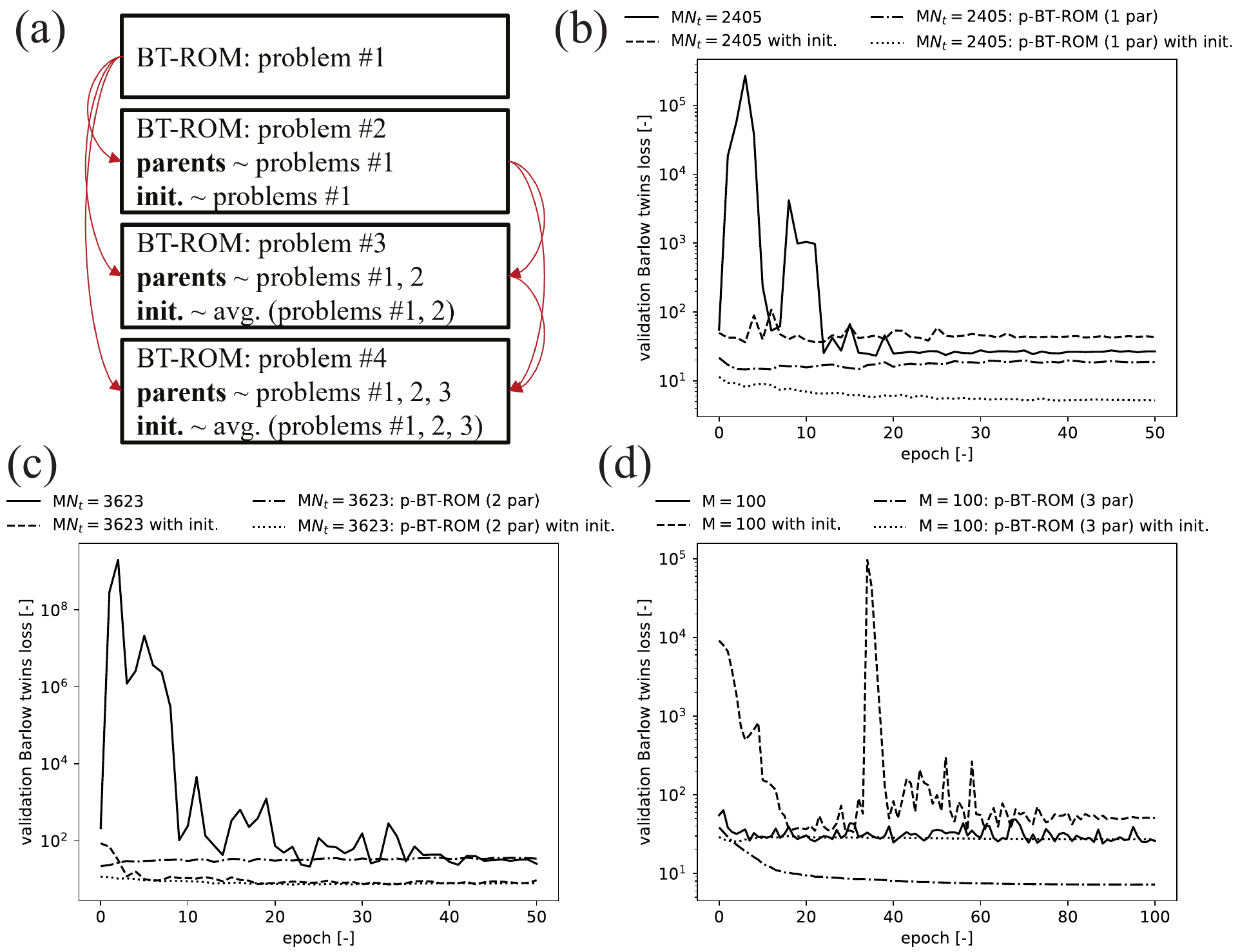}
  \caption{p-BT-ROM with initialization: (a) schematic of p-BT-ROM specifies child$\sim$parent(s) relationship and how to initialize, validation Barlow Twins loss (\eqref{eq:loss_bt}) results of (b) physical Problem \#2 (\ref{sec:physics_problems_td}) for both large dataset ($\mathrm{M}N_t = 19240$) and small dataset ($\mathrm{M}N_t = 2405$), (c) physical Problem \#3 (\ref{sec:physics_problems_el}) for both large dataset ($\mathrm{M}N_t = 28966$) and small dataset ($\mathrm{M}N_t = 3623$), and (d) physical Problem \#4 - 2-Dimensional domain (\ref{sec:physics_problems_hy}) for both large dataset ($\mathrm{M} = 1600$) and small dataset ($\mathrm{M} = 100$). We note that problems \#1 to \#3 are time-dependent problems while Problem \#4 is at a steady-state.}
  \label{fig:pnn_bt_init}
\end{figure}

\subsection{Discussion on supplement information on numerical experiments} \label{sec:si_discuss_results}

From \ref{sec:init_results}, \ref{sec:pbt_results}, and \ref{sec:pbt_init_results}, we can see that by using a trained model as an initialization (init.), p-BT-ROM, and p-BT-ROM with init. can improve our BT-ROM's accuracy. However, p-BT-ROM and p-BT-ROM with init. seem to provide a better result compared to using init. alone. However, there is no clear winner between these two approaches. Hence, for all numerical examples shown in Sec. Results in the main text, we use p-BT-ROM with init. We note that the computation cost of init. is negligible compared to the training p-BT-ROM itself. \par  

\subsection{Supplement information of similar topology section (Sec. Similar topology in the main text)}\label{sec:sup_same_topo}

Here, we investigate p-BT-ROM's performance using one mesh, see Supplementary fig. \ref{fig:sum_geo}a, for physics problems, Problems \#1 to \#4 (test model performance on different physics but similar mesh or topology).

\subsubsection{Gravity-driven in porous media - Problem \#3}\label{sec:same_topo_problem3}

Throughout this section, we focus on a gravity-driven in porous media problem - Problem \#3 (\ref{sec:physics_problems_el}). One sample of this problem is shown in Supplementary fig. \ref{fig:same_topo_samples}c. We aim to investigate the model's performance on a test set ($\mathrm{M_{test}}N_t = 7260$, $\mathrm{M_{test}} = 10$) with different training set sizes and a number of parents. This problem has three primary variables, fluid pressure, fluid velocity, and fluid temperature. Since our BT-ROM is non-intrusive, we only pick the fluid temperature as our quantity of interest. The MSE (\eqref{eq:mse}) 
and MAE (\eqref{eq:mae}) results are presented in Supplementary fig. \ref{fig:pnn_elder_s_topo_loss}. The schematic of p-BT-ROM specifies child$\sim$parent(s) relationship is shown in Supplementary fig. \ref{fig:pnn_elder_s_topo_sep}a. For the model with 1 parent, we use Problem \#1 (\ref{sec:physics_problems_tv}) as a parent with a training set of $\mathrm{M}N_t = 2405$ ($\mathrm{M} = 5$, i.e., small dataset in \ref{sec:si_results}). For the model with 2 parents, Problems \#1 and \#2 (\ref{sec:physics_problems_td}) with both have a training set of $\mathrm{M}N_t = 2405$ ($\mathrm{M} = 5$, i.e., small dataset in \ref{sec:si_results}). \par

From Supplementary fig. \ref{fig:pnn_elder_s_topo_loss}, we observe that the model with no parent and has a small training set ($\mathrm{M}N_t = 3623$, $\mathrm{M} = 5$), shown in red, performs the worst. As we add more parents, shown in blue and green, the models gain their capability and deliver better accuracy. However, even the model with two parents, shown in blue, still could not achieve the same level of accuracy as the model that trained with a large training set ($\mathrm{M}N_t = 28966$, $\mathrm{M} = 40$), but has zero parent. \par

\begin{figure}[!ht]
  \centering
  \includegraphics[width=14.0cm,keepaspectratio]{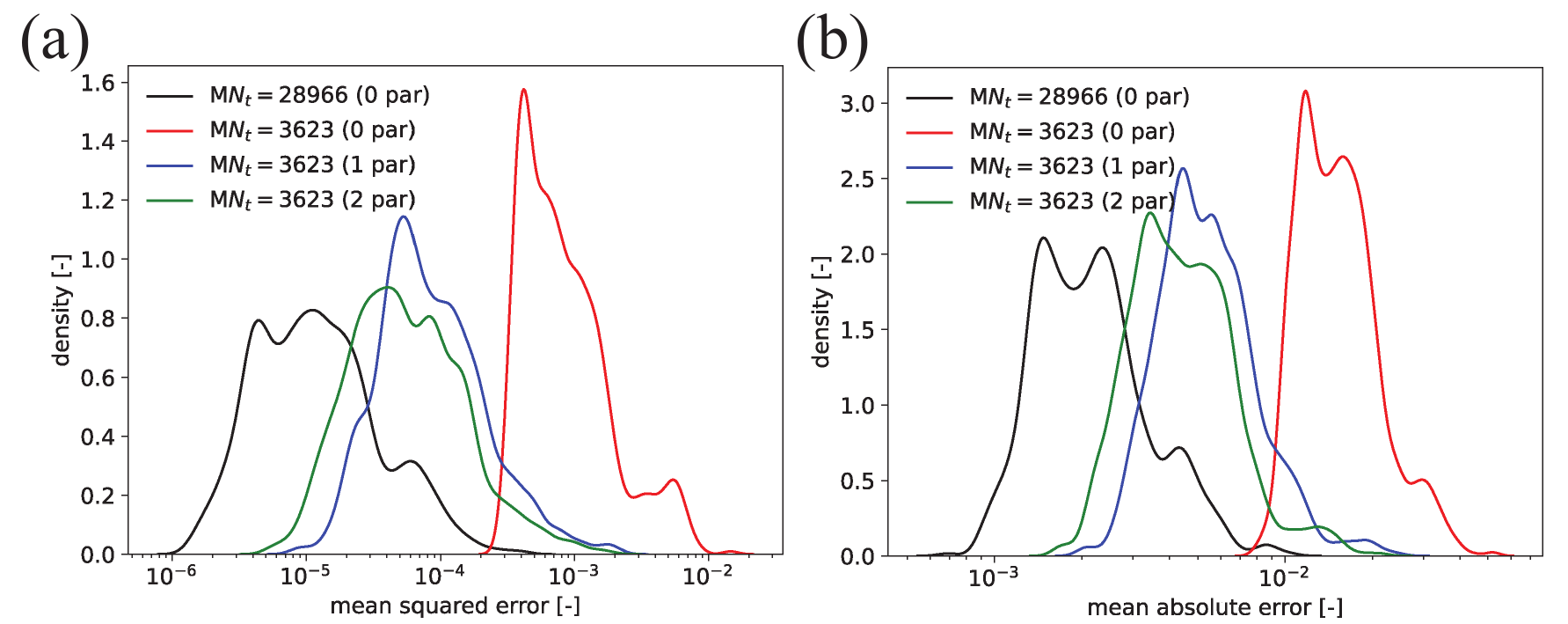}
  \caption{Similar topology - Problem \#3 (\ref{sec:physics_problems_el}): (a) mean squared error and (b) mean absolute error results for large dataset ($\mathrm{M}N_t = 28966$, $\mathrm{M} = 40$) and small dataset ($\mathrm{M}N_t = 3623, \mathrm{M} = 5$) with a different number of parents. We note that schematic of p-BT-ROM specifies child$\sim$parent(s) relationship is shown in Supplementary fig. \ref{fig:pnn_elder_s_topo_sep}a.}
  \label{fig:pnn_elder_s_topo_loss}
\end{figure}

Figs. \ref{fig:pnn_elder_s_topo_sep}b-c illustrate MAE as as a function of parameter $\bm{\mu}$ for three different cases; large dataset ($\mathrm{M}N_t = 28966$, $\mathrm{M} = 40$) with 0 parent, small dataset ($\mathrm{M}N_t = 3623$, $\mathrm{M} = 5$) with 0 parent, and small dataset ($\mathrm{M}N_t = 3623$, $\mathrm{M} = 5$) with 2 parents, respectively. One can observe that the pattern of these three figures is similar (i.e., the shape), but the magnitude of the MAE is different. This observation implies that the p-BT-ROM only alters the magnitude of the error, not its pattern.  \par

\begin{figure}[!ht]
  \centering
  \includegraphics[width=14.0cm,keepaspectratio]{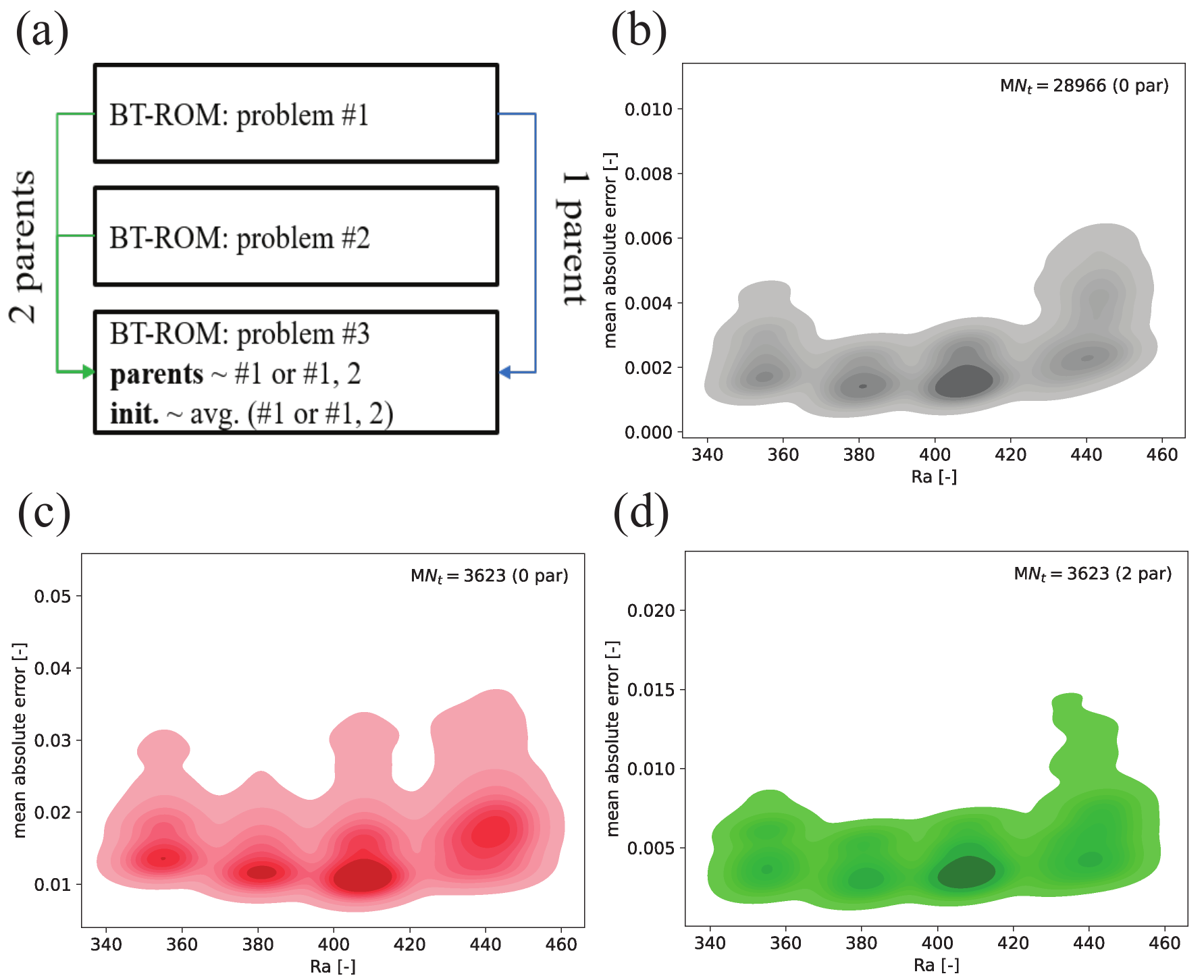}
  \caption{Similar topology - Problem \#3 (\ref{sec:physics_problems_el}): (a) schematic of p-BT-ROM specifies child$\sim$parent(s) relationship, mean absolute error as a function of parameter $\bm{\mu}$ results for (b) large dataset ($\mathrm{M}N_t = 28966, \mathrm{M} = 40$) with 0 parent, (c) small dataset ($\mathrm{M}N_t = 3623, \mathrm{M} = 5$) with 0 parent, and (d) small dataset ($\mathrm{M}N_t = 3623, \mathrm{M} = 5$) with 2 parents. We note that problems \#1 to \#3 are time-dependent problems, and their description can be found in ( \ref{sec:physics_problems_tv}, \ref{sec:physics_problems_td}, and \ref{sec:physics_problems_el}).}
  \label{fig:pnn_elder_s_topo_sep}
\end{figure}

\subsubsection{Hyperelasticity problem - Problem \#4}\label{sec:same_topo_problem4}

Next, we move to a physics problem of finite deformation of hyperelastic material - Problem \#4 (\ref{sec:physics_problems_hy}). One sample of this problem is shown in Supplementary fig. \ref{fig:same_topo_samples}d. Here, we want to investigate the model's performance by using a solid mechanics at a steady-state solution (i.e., all parents, Problems \#1 to \#3, are transient fluid mechanics problems). We use a test set of $\mathrm{M_{test}} = 100$. with different training set sizes and a number of parents. The quantity of interest, here, is a magnitude of displacement. \par

The MSE and MAE are shown in Supplementary fig. \ref{fig:pnn_hyper_s_topo_loss}, and the schematic of p-BT-ROM specifies child$\sim$parent(s) relationship is shown in Supplementary fig. \ref{fig:pnn_hyper_s_topo_sep}a. It is similar to the previous section. For the model with 1 parent, we use Problem \#1 (\ref{sec:physics_problems_tv}) as a parent with a training set of $\mathrm{M}N_t = 2405$ ($\mathrm{M} = 5$), for the model with 2 parents, Problems \#1 and \#2 (\ref{sec:physics_problems_td}) with both have a training set of $\mathrm{M}N_t = 2405$ ($\mathrm{M} = 5$), for the model with 3 parents, we add Problem \#3 (\ref{sec:physics_problems_el}) with a training set of $\mathrm{M}N_t = 3623$ ($\mathrm{M} = 5$). We note that all parents are trained using a small dataset (more details can be found in \ref{sec:si_results}). \par

From Supplementary fig. \ref{fig:pnn_hyper_s_topo_loss}, again, we observe that the model with no parent and has a small training set ($\mathrm{M} = 100$), shown in red, performs the worst, while the model with a large training set ($\mathrm{M} = 1600$) delivers the best accuracy. However, with a small training set, as we add more parents, shown in blue, green, and purple, the models gain their capability and deliver better accuracy. However, we speculate that it would be challenging to reach the same level of accuracy as that of the large dataset, even if we add even more parents. \par

\begin{figure}[!ht]
  \centering
  \includegraphics[width=14.0cm,keepaspectratio]{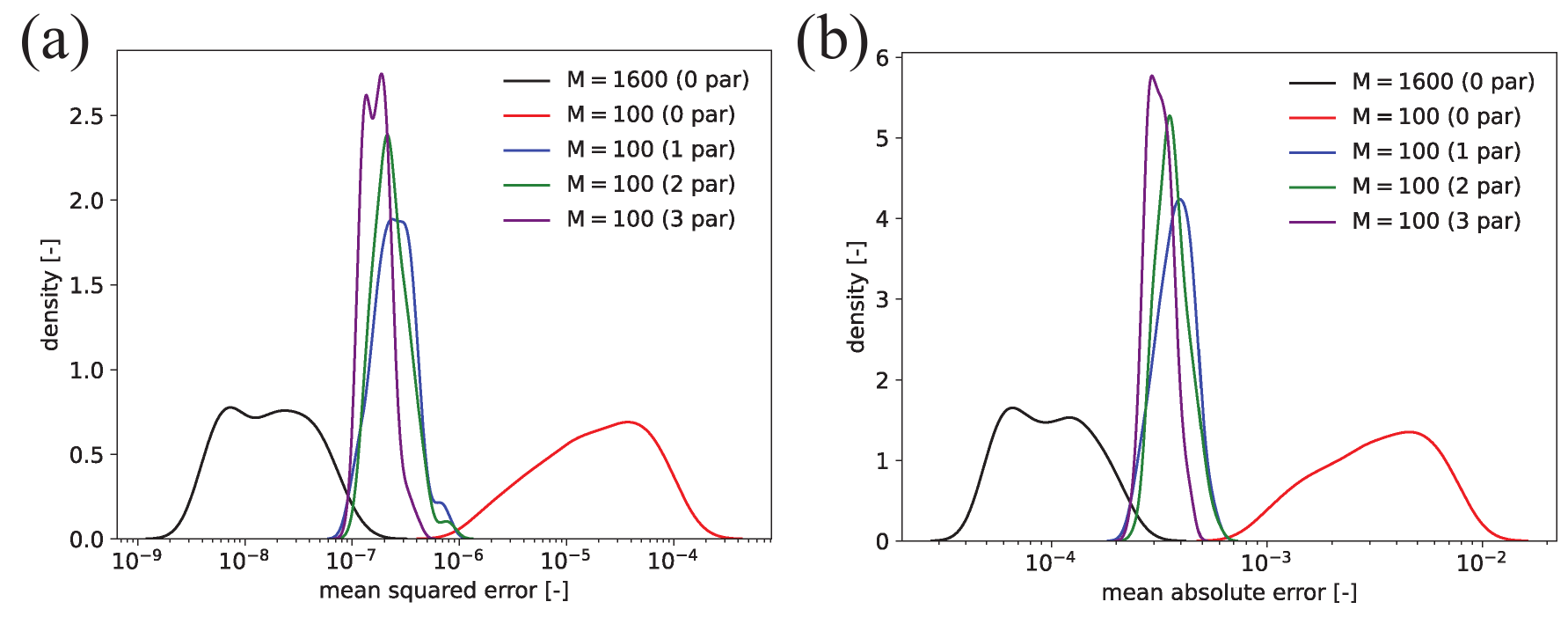}
  \caption{Similar topology - Problem \#4 (\ref{sec:physics_problems_hy}): (a) mean squared error and (b) mean absolute error results for large dataset ($\mathrm{M} = 1600$) and small dataset ($\mathrm{M} = 100$) with a different number of parents. We note that schematic of p-BT-ROM specifies child$\sim$parent(s) relationship is shown in Supplementary fig. \ref{fig:pnn_hyper_s_topo_sep}a.}
  \label{fig:pnn_hyper_s_topo_loss}
\end{figure}

Similar to the previous section as we use Figs. \ref{fig:pnn_hyper_s_topo_sep}b-c to present the MAE results as a function of parameter $\bm{\mu}$ for three different cases; large dataset ($\mathrm{M} = 1600$) with 0 parent, small dataset ($\mathrm{M} = 100$) with 0 parent, and small dataset ($\mathrm{M} = 100$) with 3 parents, respectively. Again, one can observe that the pattern of these three figures is similar (i.e., the shape), but the magnitude of the MAE is different. This observation implies that the p-BT-ROM only alters the magnitude of the error, not its pattern.  \par

\begin{figure}[!ht]
  \centering
  \includegraphics[width=14.0cm,keepaspectratio]{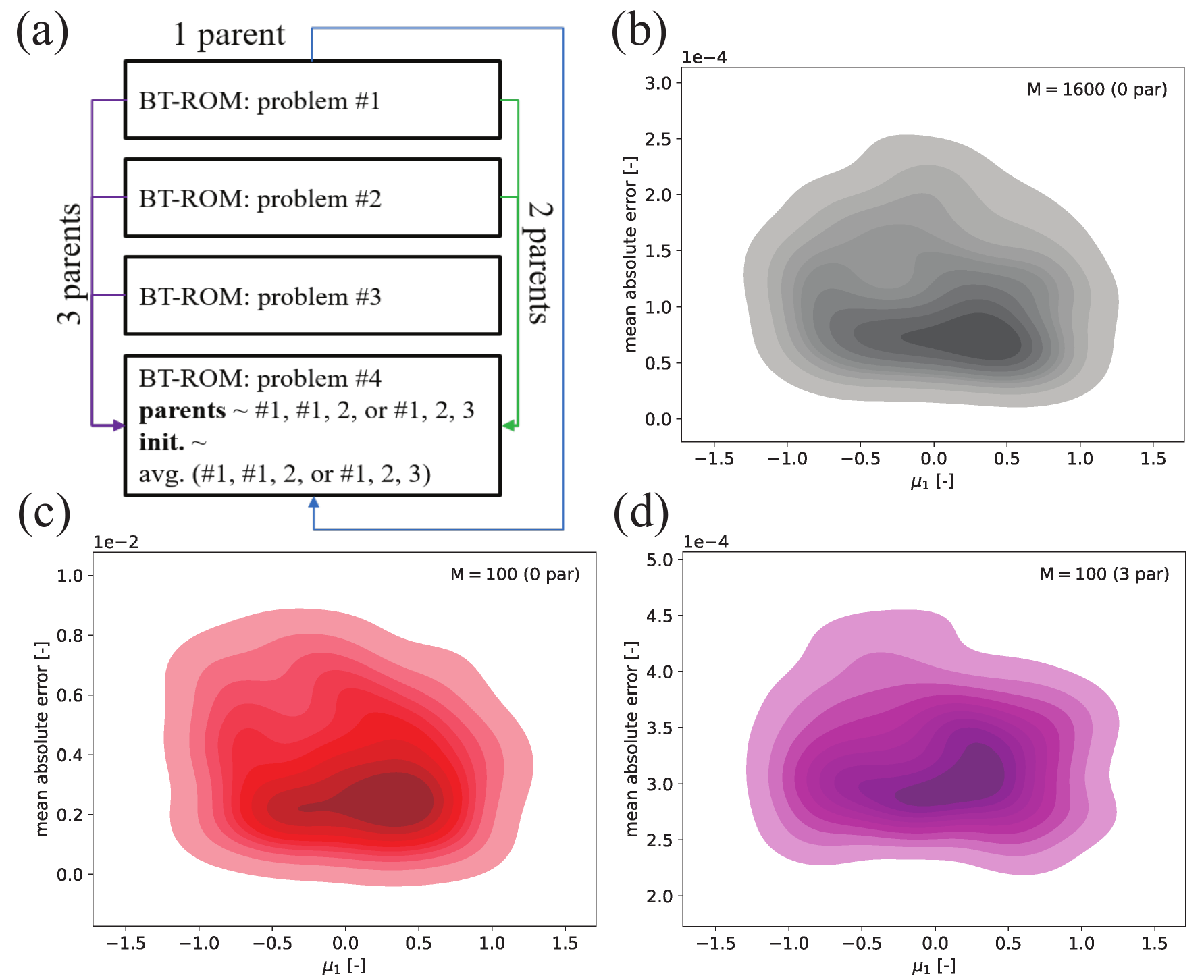}
  \caption{Similar topology - Problem \#4 (\ref{sec:physics_problems_hy}): (a) schematic of p-BT-ROM specifies child$\sim$parent(s) relationship, mean absolute error as a function of parameter $\bm{\mu}$ results for (b) large dataset ($\mathrm{M} = 1600$) with 0 parent, (c) small dataset ($\mathrm{M} = 100$) with 0 parent, and (d) small dataset ($\mathrm{M} = 100$) with 3 parents. We note that problems \#1 to \#3 are time-dependent problems while Problem \#4 is at a steady-state, and their description can be found in ( \ref{sec:physics_problems_tv}, \ref{sec:physics_problems_td}, \ref{sec:physics_problems_el}, and \ref{sec:physics_problems_hy}).}
  \label{fig:pnn_hyper_s_topo_sep}
\end{figure}

\section{Summary of number of parameters of each model} \label{sec:num_para}

We present the number of parameters of the (p-)BT-ROM with different numbers of parents in Supplementary tab. \ref{tab:count_params_s_topo} for Supplementary fig. \ref{fig:prelim_results} and Supplementary tab. \ref{tab:count_params_diff_topo} for Supplementary fig. \ref{fig:prelim_results_diff_topo}. We observe that, as expected, as we increase the number of parents, the more parameters the models contain (i.e., more gates required to control the flow of information). We also note that even though the number of parameters increases. The training and prediction costs are not much different. We train and test our model using NVIDIA Quadro RTX 8000.


\begin{table}[!ht]
\centering
\caption{Summary of the number of parameters of each model presented in Fig. \ref{fig:prelim_results} in the main text.}
\begin{tabular}{|c|c|c|c|}
\hline
parents & encoder  & decoder  & projector \\ \hline
0       & 47,744,112 & 47,752,562  & 11,168     \\ \hline
1       & 59,647,413 & 95,500,636  & 21,728     \\ \hline
2       & 71,550,714 & 143,248,710 & 32,288     \\ \hline
3       & 83,454,015 & 190,996,784 & 42,848     \\ \hline
\end{tabular}\label{tab:count_params_s_topo}
\end{table}

\begin{table}[!ht]
\centering
\caption{Summary of the number of parameters of each model presented in Fig. \ref{fig:prelim_results_diff_topo} in the main text.}
\begin{tabular}{|c|c|c|c|}
\hline
parents & encoder   & decoder   & projector \\ \hline
0       & 59,018,187  & 59,022,139  & 13,805     \\ \hline
1       & 70,920,019  & 106,770,213 & 25,834     \\ \hline
2       & 82,822,340  & 154,518,287 & 37,374     \\ \hline
3       & 94,724,940  & 202,266,361 & 48,635     \\ \hline
4       & 106,627,716 & 250,014,435 & 59,720     \\ \hline
\end{tabular}\label{tab:count_params_diff_topo}
\end{table}

\section{Complementary information on numerical results for different topologies section (Sec. Different topologies in the main text)} \label{sec:sup_result_diff_topo}

Supplementary fig. \ref{fig:mae_as_mu_diff_topo} is used to showing the pattern of MAE as a function of parameter $\bm{\mu}$. We discuss this figure in Sec. Different topologies in the main text.

\begin{figure}[!ht]
  \centering
  \includegraphics[width=14.0cm,keepaspectratio]{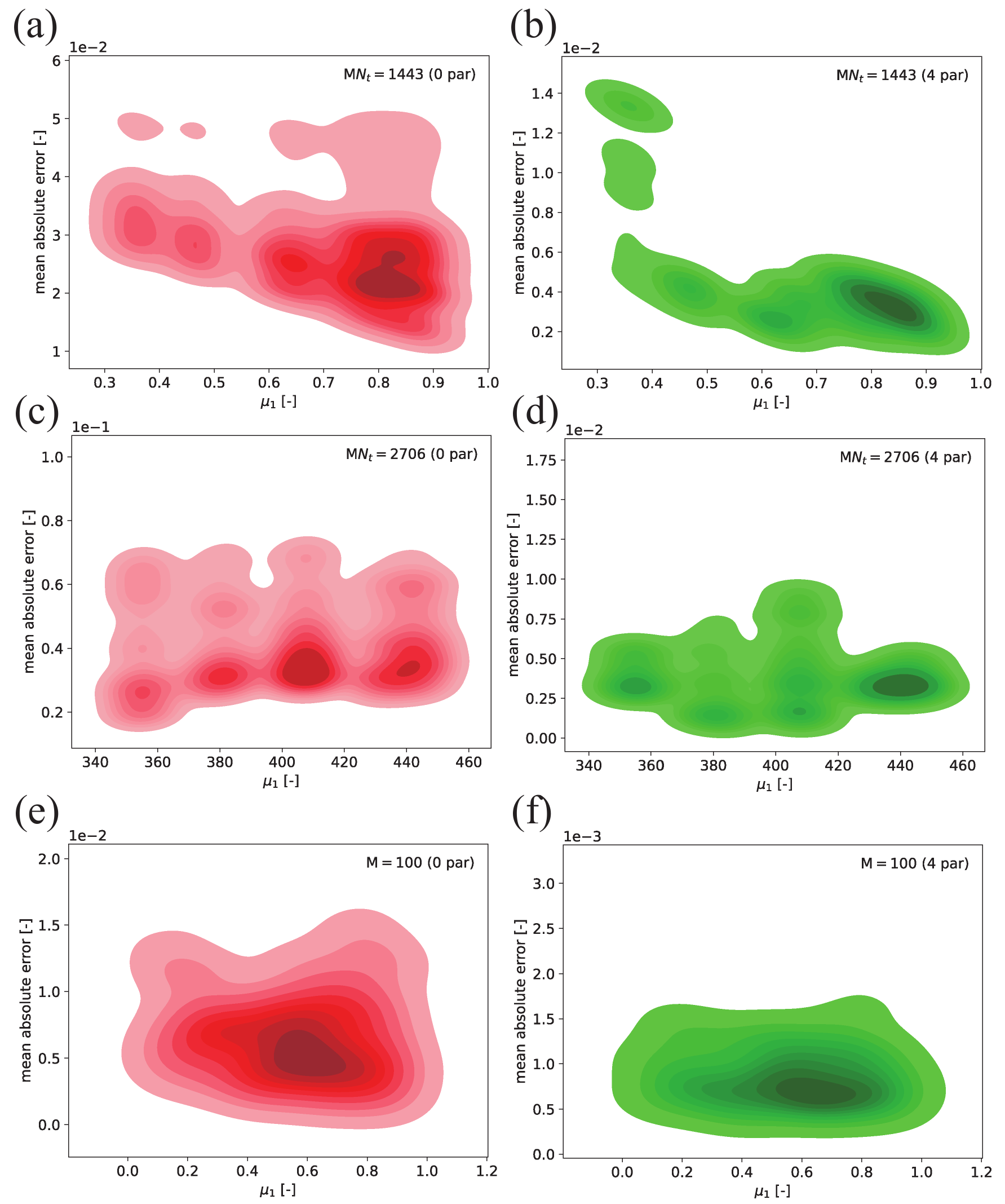}
  \caption{Different topologies - mean absolute error as a function of parameter $\bm{\mu}$ results for (a) 0 and (b) 4 parents for Problem \#2 (\ref{sec:physics_problems_td}) using the topology shown in Supplementary fig. \ref{fig:sum_geo}b, (c) 0 and (d) 4 parents for Problem \#3 (\ref{sec:physics_problems_el}) using the topology shown in Supplementary fig. \ref{fig:sum_geo}c, and (e) 0 and (f) 4 parents for Problem \#4 (\ref{sec:physics_problems_hy}) using the topology shown in Supplementary fig. \ref{fig:sum_geo}d.}
  \label{fig:mae_as_mu_diff_topo}
\end{figure}

\bibliographystyle{naturemag-doi}  
\bibliography{references}

\begin{thebibliography}{10}
\urlstyle{rm}
\expandafter\ifx\csname url\endcsname\relax
  \def\url#1{\texttt{#1}}\fi
\expandafter\ifx\csname urlprefix\endcsname\relax\def\urlprefix{URL }\fi
\expandafter\ifx\csname doiprefix\endcsname\relax\def\doiprefix{DOI: }\fi
\providecommand{\bibinfo}[2]{#2}
\providecommand{\eprint}[2][]{\url{#2}}

\bibitem{masood2022cop27}
\bibinfo{author}{Masood, E.}, \bibinfo{author}{Tollefson, J.}, \bibinfo{author}{Irwin, A.} \emph{et~al.}
\newblock \bibinfo{journal}{\bibinfo{title}{Cop27 climate talks: what succeeded, what failed and what’s next}}.
\newblock {\emph{\JournalTitle{Nature}}} \textbf{\bibinfo{volume}{612}}, \bibinfo{pages}{16--17} (\bibinfo{year}{2022}).

\bibitem{falk2022urgent}
\bibinfo{author}{Falk, J.} \emph{et~al.}
\newblock \bibinfo{journal}{\bibinfo{title}{An urgent need for cop27: confronting converging crises}}.
\newblock {\emph{\JournalTitle{Sustainability Science}}} \bibinfo{pages}{1--5} (\bibinfo{year}{2022}).

\bibitem{hu2008multiple}
\bibinfo{author}{Hu, L.} \& \bibinfo{author}{Chugunova, T.}
\newblock \bibinfo{journal}{\bibinfo{title}{Multiple-point geostatistics for modeling subsurface heterogeneity: A comprehensive review}}.
\newblock {\emph{\JournalTitle{Water Resources Research}}} \textbf{\bibinfo{volume}{44}} (\bibinfo{year}{2008}).

\bibitem{hartmann2016putting}
\bibinfo{author}{Hartmann, A.}
\newblock \bibinfo{journal}{\bibinfo{title}{Putting the cat in the box: why our models should consider subsurface heterogeneity at all scales}}.
\newblock {\emph{\JournalTitle{Wiley Interdisciplinary Reviews: Water}}} \textbf{\bibinfo{volume}{3}}, \bibinfo{pages}{478--486} (\bibinfo{year}{2016}).

\bibitem{ginn2002processes}
\bibinfo{author}{Ginn, T.} \emph{et~al.}
\newblock \bibinfo{journal}{\bibinfo{title}{Processes in microbial transport in the natural subsurface}}.
\newblock {\emph{\JournalTitle{Advances in Water Resources}}} \textbf{\bibinfo{volume}{25}}, \bibinfo{pages}{1017--1042} (\bibinfo{year}{2002}).

\bibitem{mccarthy2004colloid}
\bibinfo{author}{McCarthy, J.} \& \bibinfo{author}{McKay, L.}
\newblock \bibinfo{journal}{\bibinfo{title}{Colloid transport in the subsurface: Past, present, and future challenges}}.
\newblock {\emph{\JournalTitle{Vadose Zone Journal}}} \textbf{\bibinfo{volume}{3}}, \bibinfo{pages}{326--337} (\bibinfo{year}{2004}).

\bibitem{evans2012numerical}
\bibinfo{author}{Evans, G.}, \bibinfo{author}{Blackledge, J.} \& \bibinfo{author}{Yardley, P.}
\newblock \emph{\bibinfo{title}{Numerical methods for partial differential equations}} (\bibinfo{publisher}{Springer Science \& Business Media}, \bibinfo{year}{2012}).

\bibitem{hesthaven2016certified}
\bibinfo{author}{Hesthaven, J.}, \bibinfo{author}{Rozza, G.}, \bibinfo{author}{Stamm, B.} \emph{et~al.}
\newblock \emph{\bibinfo{title}{Certified reduced basis methods for parametrized partial differential equations}} (\bibinfo{publisher}{Springer}, \bibinfo{year}{2016}).

\bibitem{chen2023capacity}
\bibinfo{author}{Chen, F.} \emph{et~al.}
\newblock \bibinfo{journal}{\bibinfo{title}{Capacity assessment and cost analysis of geologic storage of hydrogen: A case study in intermountain-west region usa}}.
\newblock {\emph{\JournalTitle{International Journal of Hydrogen Energy}}} \textbf{\bibinfo{volume}{48}}, \bibinfo{pages}{9008--9022} (\bibinfo{year}{2023}).

\bibitem{wen2023real}
\bibinfo{author}{Wen, G.} \emph{et~al.}
\newblock \bibinfo{journal}{\bibinfo{title}{Real-time high-resolution co 2 geological storage prediction using nested fourier neural operators}}.
\newblock {\emph{\JournalTitle{Energy \& Environmental Science}}} \textbf{\bibinfo{volume}{16}}, \bibinfo{pages}{1732--1741} (\bibinfo{year}{2023}).

\bibitem{lengler2010impact}
\bibinfo{author}{Lengler, U.}, \bibinfo{author}{De~Lucia, M.} \& \bibinfo{author}{K{\"u}hn, M.}
\newblock \bibinfo{journal}{\bibinfo{title}{The impact of heterogeneity on the distribution of co2: Numerical simulation of co2 storage at ketzin}}.
\newblock {\emph{\JournalTitle{International Journal of Greenhouse Gas Control}}} \textbf{\bibinfo{volume}{4}}, \bibinfo{pages}{1016--1025} (\bibinfo{year}{2010}).

\bibitem{cho2021estimation}
\bibinfo{author}{Cho, Y.} \& \bibinfo{author}{Jun, H.}
\newblock \bibinfo{journal}{\bibinfo{title}{Estimation and uncertainty analysis of the co2 storage volume in the sleipner field via 4d reversible-jump markov-chain monte carlo}}.
\newblock {\emph{\JournalTitle{Journal of Petroleum Science and Engineering}}} \textbf{\bibinfo{volume}{200}}, \bibinfo{pages}{108333} (\bibinfo{year}{2021}).

\bibitem{choi2020sns}
\bibinfo{author}{Choi, Y.}, \bibinfo{author}{Coombs, D.} \& \bibinfo{author}{Anderson, R.}
\newblock \bibinfo{journal}{\bibinfo{title}{Sns: a solution-based nonlinear subspace method for time-dependent model order reduction}}.
\newblock {\emph{\JournalTitle{SIAM Journal on Scientific Computing}}} \textbf{\bibinfo{volume}{42}}, \bibinfo{pages}{A1116--A1146} (\bibinfo{year}{2020}).

\bibitem{kapteyn2022data}
\bibinfo{author}{Kapteyn, M.}, \bibinfo{author}{Knezevic, D.}, \bibinfo{author}{Huynh, D.}, \bibinfo{author}{Tran, M.} \& \bibinfo{author}{Willcox, K.}
\newblock \bibinfo{journal}{\bibinfo{title}{Data-driven physics-based digital twins via a library of component-based reduced-order models}}.
\newblock {\emph{\JournalTitle{International Journal for Numerical Methods in Engineering}}} \textbf{\bibinfo{volume}{123}}, \bibinfo{pages}{2986--3003} (\bibinfo{year}{2022}).

\bibitem{silva2023data}
\bibinfo{author}{Silva, V.}, \bibinfo{author}{Heaney, C.}, \bibinfo{author}{Li, Y.} \& \bibinfo{author}{Pain, C.}
\newblock \bibinfo{journal}{\bibinfo{title}{Data assimilation predictive gan (da-predgan) applied to a spatio-temporal compartmental model in epidemiology}}.
\newblock {\emph{\JournalTitle{Journal of Scientific Computing}}} \textbf{\bibinfo{volume}{94}}, \bibinfo{pages}{1--31} (\bibinfo{year}{2023}).

\bibitem{qin2021deep}
\bibinfo{author}{Qin, T.}, \bibinfo{author}{Chen, Z.}, \bibinfo{author}{Jakeman, J.} \& \bibinfo{author}{Xiu, D.}
\newblock \bibinfo{journal}{\bibinfo{title}{Deep learning of parameterized equations with applications to uncertainty quantification}}.
\newblock {\emph{\JournalTitle{International Journal for Uncertainty Quantification}}} \textbf{\bibinfo{volume}{11}} (\bibinfo{year}{2021}).

\bibitem{xu2022physics}
\bibinfo{author}{Xu, K.} \& \bibinfo{author}{Darve, E.}
\newblock \bibinfo{journal}{\bibinfo{title}{Physics constrained learning for data-driven inverse modeling from sparse observations}}.
\newblock {\emph{\JournalTitle{Journal of Computational Physics}}} \textbf{\bibinfo{volume}{453}}, \bibinfo{pages}{110938} (\bibinfo{year}{2022}).

\bibitem{pachalieva2022physics}
\bibinfo{author}{Pachalieva, A.}, \bibinfo{author}{O’Malley, D.}, \bibinfo{author}{Harp, D.} \& \bibinfo{author}{Viswanathan, H.}
\newblock \bibinfo{journal}{\bibinfo{title}{Physics-informed machine learning with differentiable programming for heterogeneous underground reservoir pressure management}}.
\newblock {\emph{\JournalTitle{Scientific Reports}}} \textbf{\bibinfo{volume}{12}}, \bibinfo{pages}{1--12} (\bibinfo{year}{2022}).

\bibitem{raissi2019physics}
\bibinfo{author}{Raissi, M.}, \bibinfo{author}{Perdikaris, P.} \& \bibinfo{author}{Karniadakis, G.}
\newblock \bibinfo{journal}{\bibinfo{title}{Physics-informed neural networks: A deep learning framework for solving forward and inverse problems involving nonlinear partial differential equations}}.
\newblock {\emph{\JournalTitle{Journal of Computational Physics}}} \textbf{\bibinfo{volume}{378}}, \bibinfo{pages}{686--707} (\bibinfo{year}{2019}).

\bibitem{weiss2016survey}
\bibinfo{author}{Weiss, K.}, \bibinfo{author}{Khoshgoftaar, T.} \& \bibinfo{author}{Wang, D.}
\newblock \bibinfo{journal}{\bibinfo{title}{A survey of transfer learning}}.
\newblock {\emph{\JournalTitle{Journal of Big data}}} \textbf{\bibinfo{volume}{3}}, \bibinfo{pages}{1--40} (\bibinfo{year}{2016}).

\bibitem{zhuang2020comprehensive}
\bibinfo{author}{Zhuang, F.} \emph{et~al.}
\newblock \bibinfo{journal}{\bibinfo{title}{A comprehensive survey on transfer learning}}.
\newblock {\emph{\JournalTitle{Proceedings of the IEEE}}} \textbf{\bibinfo{volume}{109}}, \bibinfo{pages}{43--76} (\bibinfo{year}{2020}).

\bibitem{rusu2016progressive}
\bibinfo{author}{Rusu, A.} \emph{et~al.}
\newblock \bibinfo{journal}{\bibinfo{title}{Progressive neural networks}}.
\newblock {\emph{\JournalTitle{arXiv preprint arXiv:1606.04671}}}  (\bibinfo{year}{2016}).

\bibitem{kadeethum2021framework}
\bibinfo{author}{Kadeethum, T.} \emph{et~al.}
\newblock \bibinfo{journal}{\bibinfo{title}{A framework for data-driven solution and parameter estimation of pdes using conditional generative adversarial networks}}.
\newblock {\emph{\JournalTitle{Nature Computational Science}}} \textbf{\bibinfo{volume}{1}}, \bibinfo{pages}{819–829}, \doiprefix\url{https://doi.org/10.1038/s43588-021-00171-3} (\bibinfo{year}{2021}).

\bibitem{kadeethum2022reduced}
\bibinfo{author}{Kadeethum, T.} \emph{et~al.}
\newblock \bibinfo{journal}{\bibinfo{title}{Reduced order modeling for flow and transport problems with barlow twins self-supervised learning}}.
\newblock {\emph{\JournalTitle{Scientific Reports}}} \textbf{\bibinfo{volume}{12}}, \bibinfo{pages}{1--18} (\bibinfo{year}{2022}).

\bibitem{kadeethum2022uqbtrom}
\bibinfo{author}{Kadeethum, T.}, \bibinfo{author}{Jakeman, J.}, \bibinfo{author}{Choi, Y.}, \bibinfo{author}{Bouklas, N.} \& \bibinfo{author}{Yoon, H.}
\newblock \bibinfo{journal}{\bibinfo{title}{Epistemic uncertainty-aware barlow twins reduced order modeling for nonlinear contact problems}}.
\newblock {\emph{\JournalTitle{under review}}}  (\bibinfo{year}{2022}).

\bibitem{goswami2022deep}
\bibinfo{author}{Goswami, S.}, \bibinfo{author}{Kontolati, K.}, \bibinfo{author}{Shields, M.} \& \bibinfo{author}{Karniadakis, G.}
\newblock \bibinfo{journal}{\bibinfo{title}{Deep transfer operator learning for partial differential equations under conditional shift}}.
\newblock {\emph{\JournalTitle{Nature Machine Intelligence}}} \textbf{\bibinfo{volume}{4}}, \bibinfo{pages}{1155--1164} (\bibinfo{year}{2022}).

\bibitem{huang2023gate}
\bibinfo{author}{Huang, Z.}, \bibinfo{author}{Dang, L.}, \bibinfo{author}{Xie, Y.}, \bibinfo{author}{Ma, W.} \& \bibinfo{author}{Chen, B.}
\newblock \bibinfo{journal}{\bibinfo{title}{Gate recurrent unit network based on hilbert-schmidt independence criterion for state-of-health estimation}}.
\newblock {\emph{\JournalTitle{arXiv preprint arXiv:2303.09497}}}  (\bibinfo{year}{2023}).

\bibitem{vaswani2017attention}
\bibinfo{author}{Vaswani, A.} \emph{et~al.}
\newblock \bibinfo{journal}{\bibinfo{title}{Attention is all you need}}.
\newblock {\emph{\JournalTitle{Advances in neural information processing systems}}} \textbf{\bibinfo{volume}{30}} (\bibinfo{year}{2017}).

\bibitem{alesiani2023gated}
\bibinfo{author}{Alesiani, F.}, \bibinfo{author}{Yu, S.} \& \bibinfo{author}{Yu, X.}
\newblock \bibinfo{journal}{\bibinfo{title}{Gated information bottleneck for generalization in sequential environments}}.
\newblock {\emph{\JournalTitle{Knowledge and Information Systems}}} \textbf{\bibinfo{volume}{65}}, \bibinfo{pages}{683--705} (\bibinfo{year}{2023}).

\bibitem{drori2022neural}
\bibinfo{author}{Drori, I.} \emph{et~al.}
\newblock \bibinfo{journal}{\bibinfo{title}{A neural network solves, explains, and generates university math problems by program synthesis and few-shot learning at human level}}.
\newblock {\emph{\JournalTitle{Proceedings of the National Academy of Sciences}}} \textbf{\bibinfo{volume}{119}}, \bibinfo{pages}{e2123433119} (\bibinfo{year}{2022}).

\bibitem{kadeethum2021nonTH}
\bibinfo{author}{Kadeethum, T.} \emph{et~al.}
\newblock \bibinfo{journal}{\bibinfo{title}{Non-intrusive reduced order modeling of natural convection in porous media using convolutional autoencoders: comparison with linear subspace techniques}}.
\newblock {\emph{\JournalTitle{Advances in Water Resources}}} \bibinfo{pages}{104098} (\bibinfo{year}{2022}).

\bibitem{kadeethum2022fomassistrom}
\bibinfo{author}{Kadeethum, T.} \emph{et~al.}
\newblock \bibinfo{journal}{\bibinfo{title}{Enhancing high-fidelity nonlinear solver with reduced order model}}.
\newblock {\emph{\JournalTitle{Scientific Reports}}} \textbf{\bibinfo{volume}{12}}, \bibinfo{pages}{1--15} (\bibinfo{year}{2022}).

\bibitem{lu2021learning}
\bibinfo{author}{Lu, L.}, \bibinfo{author}{Jin, P.}, \bibinfo{author}{Pang, G.}, \bibinfo{author}{Zhang, Z.} \& \bibinfo{author}{Karniadakis, G.}
\newblock \bibinfo{journal}{\bibinfo{title}{Learning nonlinear operators via deeponet based on the universal approximation theorem of operators}}.
\newblock {\emph{\JournalTitle{Nature Machine Intelligence}}} \textbf{\bibinfo{volume}{3}}, \bibinfo{pages}{218--229} (\bibinfo{year}{2021}).

\bibitem{choi2020gradient}
\bibinfo{author}{Choi, Y.}, \bibinfo{author}{Boncoraglio, G.}, \bibinfo{author}{Anderson, S.}, \bibinfo{author}{Amsallem, D.} \& \bibinfo{author}{Farhat, F.}
\newblock \bibinfo{journal}{\bibinfo{title}{Gradient-based constrained optimization using a database of linear reduced-order models}}.
\newblock {\emph{\JournalTitle{Journal of Computational Physics}}} \textbf{\bibinfo{volume}{423}} (\bibinfo{year}{2020}).

\bibitem{oommen2022learning}
\bibinfo{author}{Oommen, V.}, \bibinfo{author}{Shukla, K.}, \bibinfo{author}{Goswami, S.}, \bibinfo{author}{Dingreville, R.} \& \bibinfo{author}{Karniadakis, G.}
\newblock \bibinfo{journal}{\bibinfo{title}{Learning two-phase microstructure evolution using neural operators and autoencoder architectures}}.
\newblock {\emph{\JournalTitle{arXiv preprint arXiv:2204.07230}}}  (\bibinfo{year}{2022}).

\bibitem{gorodetsky2021mfnets}
\bibinfo{author}{Gorodetsky, A.}, \bibinfo{author}{Jakeman, J.} \& \bibinfo{author}{Geraci, G.}
\newblock \bibinfo{journal}{\bibinfo{title}{Mfnets: data efficient all-at-once learning of multifidelity surrogates as directed networks of information sources}}.
\newblock {\emph{\JournalTitle{Computational Mechanics}}} \textbf{\bibinfo{volume}{68}}, \bibinfo{pages}{741--758} (\bibinfo{year}{2021}).

\bibitem{kadeethum2021locally}
\bibinfo{author}{Kadeethum, T.}, \bibinfo{author}{Lee, S.}, \bibinfo{author}{Ballarin, F.}, \bibinfo{author}{Choo, J.} \& \bibinfo{author}{Nick, H.}
\newblock \bibinfo{journal}{\bibinfo{title}{A locally conservative mixed finite element framework for coupled hydro-mechanical-chemical processes in heterogeneous porous media}}.
\newblock {\emph{\JournalTitle{Computers \& Geosciences}}} \bibinfo{pages}{104774} (\bibinfo{year}{2021}).

\bibitem{kumar2013contact}
\bibinfo{author}{Kumar, N.} \& \bibinfo{author}{DasGupta, A.}
\newblock \bibinfo{journal}{\bibinfo{title}{On the contact problem of an inflated spherical hyperelastic membrane}}.
\newblock {\emph{\JournalTitle{International Journal of Non-Linear Mechanics}}} \textbf{\bibinfo{volume}{57}}, \bibinfo{pages}{130--139} (\bibinfo{year}{2013}).

\bibitem{luo2016topology}
\bibinfo{author}{Luo, Y.}, \bibinfo{author}{Li, M.} \& \bibinfo{author}{Kang, Z.}
\newblock \bibinfo{journal}{\bibinfo{title}{Topology optimization of hyperelastic structures with frictionless contact supports}}.
\newblock {\emph{\JournalTitle{International Journal of Solids and Structures}}} \textbf{\bibinfo{volume}{81}}, \bibinfo{pages}{373--382} (\bibinfo{year}{2016}).

\bibitem{nezamabadi2011solving}
\bibinfo{author}{Nezamabadi, S.}, \bibinfo{author}{Zahrouni, H.} \& \bibinfo{author}{Yvonnet, J.}
\newblock \bibinfo{journal}{\bibinfo{title}{Solving hyperelastic material problems by asymptotic numerical method}}.
\newblock {\emph{\JournalTitle{Computational mechanics}}} \textbf{\bibinfo{volume}{47}}, \bibinfo{pages}{77--92} (\bibinfo{year}{2011}).

\bibitem{petsc-user-ref}
\bibinfo{author}{Balay, S.} \emph{et~al.}
\newblock \bibinfo{title}{{PETSc Users Manual}}.
\newblock \bibinfo{type}{Tech. Rep.} \bibinfo{number}{ANL-95/11 - Revision 3.10}, \bibinfo{institution}{Argonne National Laboratory} (\bibinfo{year}{2018}).

\bibitem{kingma2014adam}
\bibinfo{author}{Kingma, D.} \& \bibinfo{author}{Ba, J.}
\newblock \bibinfo{journal}{\bibinfo{title}{Adam: A method for stochastic optimization}}.
\newblock {\emph{\JournalTitle{arXiv preprint arXiv:1412.6980}}}  (\bibinfo{year}{2014}).

\bibitem{loshchilov2016sgdr}
\bibinfo{author}{Loshchilov, I.} \& \bibinfo{author}{Hutter, F.}
\newblock \bibinfo{journal}{\bibinfo{title}{Sgdr: Stochastic gradient descent with warm restarts}}.
\newblock {\emph{\JournalTitle{arXiv preprint arXiv:1608.03983}}}  (\bibinfo{year}{2016}).

\bibitem{prechelt1998early}
\bibinfo{author}{Prechelt, L.}
\newblock \bibinfo{title}{Early stopping-but when?}
\newblock In \emph{\bibinfo{booktitle}{Neural Networks: Tricks of the trade}}, \bibinfo{pages}{55--69} (\bibinfo{publisher}{Springer}, \bibinfo{year}{1998}).

\bibitem{prechelt1998automatic}
\bibinfo{author}{Prechelt, L.}
\newblock \bibinfo{journal}{\bibinfo{title}{Automatic early stopping using cross validation: quantifying the criteria}}.
\newblock {\emph{\JournalTitle{Neural Networks}}} \textbf{\bibinfo{volume}{11}}, \bibinfo{pages}{761--767} (\bibinfo{year}{1998}).

\bibitem{wright2003radial}
\bibinfo{author}{Wright, G.}
\newblock \emph{\bibinfo{title}{Radial basis function interpolation: numerical and analytical developments}} (\bibinfo{publisher}{University of Colorado at Boulder}, \bibinfo{year}{2003}).

\end{thebibliography}
\end{NoHyper}

\end{document}